\documentclass{article}

\usepackage{arxiv}

\usepackage[utf8]{inputenc} 
\usepackage[T1]{fontenc}    
\usepackage{hyperref}       
\usepackage{url}            
\usepackage{booktabs}       
\usepackage{amsfonts}       
\usepackage{nicefrac}       
\usepackage{microtype}      
\usepackage{amsmath}
\usepackage{amsthm}
\usepackage{cleveref}       
\usepackage{lipsum}         
\usepackage{graphicx}
\usepackage{natbib}
\usepackage{doi}

\usepackage{color}

\usepackage{bm}
\usepackage[algoruled]{algorithm2e}
\usepackage{algorithmic}

\newcommand{\domain}{\mathop{\rm dom}}      
\newcommand{\interior}{\mathop{\rm int}}    
\newcommand{\convhull}{\mathop{\rm CH}}     
\newcommand{\hset}{\mathcal{H}}             
\newcommand{\xset}{\mathcal{X}}             
\newcommand{\cset}{\mathcal{C}}             
\newcommand{\eset}{\mathcal{E}}             
\newcommand{\psimplex}[1]{\mathcal{P}^{#1}} 
\newcommand{\secalg}{\mathcal{B}}           
\newcommand{\fwalg}{\mathcal{F}}            
\newcommand{\clip}{\mathop{\rm clip}_{[0, 1]}}

\title{Boosting as Frank-Wolfe}

\author{
    \href{https://orcid.org/0000-0003-2277-6750}{
        \includegraphics[scale=0.06]{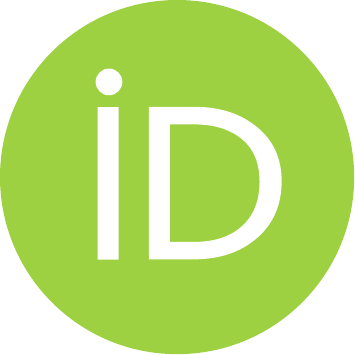}
        \hspace{1mm} Ryotaro Mitsuboshi
    } \\
    Kyushu University/RIKEN AIP \\
    \texttt{ryotaro.mitsuboshi@inf.kyushu-u.ac.jp} \\
    \And
    \href{https://orcid.org/0000-0002-1536-1269}{
        \includegraphics[scale=0.06]{orcid.pdf}
        \hspace{1mm}Kohei Hatano
    } \\
    Kyushu University/RIKEN AIP \\
    \texttt{hatano@inf.kyushu-u.ac.jp} \\
    \And
    Eiji Takimoto \\
    Kyushu University \\
    \texttt{eiji@inf.kyushu-u.ac.jp} \\
}

\renewcommand{\shorttitle}{Boosting as Frank-Wolfe}

\hypersetup{
pdftitle={A template for the arxiv style},
pdfsubject={q-bio.NC, q-bio.QM},
pdfauthor={Ryotaro Mitsuboshi, Kohei Hatano, and Eiji Takimoto},
pdfkeywords={First keyword, Second keyword, More},
}

\theoremstyle{plain}
\newtheorem{thm.}{Theorem}
\newtheorem{lem.}{Lemma}
\newtheorem{prp.}{Proposition}
\newtheorem{cor.}{Corollary}

\theoremstyle{definition}
\newtheorem{def.}{Definition}

\theoremstyle{remark}

\begin{document}
    \maketitle

    \begin{abstract}
        Some boosting algorithms, such as LPBoost, ERLPBoost, and C-ERLPBoost, 
aim to solve 
the soft margin optimization problem with the $\ell_1$-norm regularization. 
LPBoost rapidly converges to an $\epsilon$-approximate solution 
in practice, 
but it is known to take $\Omega(m)$ iterations in the worst case,
where $m$ is the sample size.
On the other hand, ERLPBoost and C-ERLPBoost are 
guaranteed to converge to an $\epsilon$-approximate solution 
in $O(\frac{1}{\epsilon^2} \ln \frac{m}{\nu})$ iterations, 
where $\nu \in [1, m]$ is a hyperparameter. 
However, the overall computations are very high compared to LPBoost. 

To address this issue,
we propose a generic boosting scheme 
that combines the Frank-Wolfe algorithm and any secondary algorithm 
and switches one to the other iteratively.
We show that the scheme retains the same convergence guarantee 
as ERLPBoost and C-ERLPBoost. 
One can incorporate any secondary algorithm to improve in practice.
This scheme comes from 
a unified view of boosting algorithms for soft margin optimization. 
More specifically, 
we show that LPBoost, ERLPBoost, and C-ERLPBoost are instances 
of the Frank-Wolfe algorithm. 
In experiments on real datasets, 
one of the instances of our scheme 
exploits the better updates of the second algorithm 
and performs comparably with LPBoost. 

    \end{abstract}


    \section{Introduction}
\label{sec:introduction}
Theory and algorithms for large-margin classifiers 
have been studied extensively 
since those classifiers guarantee low generalization errors 
when they have large margins over training examples 
(e.g.,~\cite{schapire+:as98,mohri+:mitpress18}). 
In particular, 
the $\ell_1$-norm regularized soft margin optimization problem, 
defined later, is a formulation of 
finding sparse large-margin classifiers based on the linear program (LP). 
This problem aims to optimize the $\ell_1$-margin 
by combining multiple hypotheses from some hypothesis class $\hset$. 
The resulting classifier tends to be sparse, 
so $\ell_1$-margin optimization is helpful for feature selection tasks.
Off-the-shelf LP solvers can solve the problem, 
but they are still not efficient enough for a huge class $\hset$. 

Boosting is a framework 
for solving the $\ell_1$-norm regularized margin optimization 
even though $\hset$ is infinitely large. 
Various boosting algorithms have been invented. 
LPBoost~\citep{demiriz+:ml02} is a practical algorithm 
that often works effectively. 
Although LPBoost terminates rapidly, 
It is shown that 
it takes $\Omega(m)$ iterations in the worst case, 
where $m$ is the number of training examples~\citep{warmuth+:nips07}. 
\cite{shalev-shwartz+:jml10} invented
an algorithm
called Corrective ERLPBoost 
(we call this algorithm C-ERLPBoost for shorthand) 
in the paper on ERLPBoost~\citep{warmuth+:alt08}. 
C-ERLPBooost and ERLPBoost 
find $\epsilon$-approximate solutions 
in $O(\ln(m/\nu) / \epsilon^2)$ iterations, 
where $\nu \in [1, m]$ is the soft margin parameter. 
The difference is the time complexity per iteration; 
ERLPBoost solves a convex program (CP) for each iteration, 
while C-ERLPBooost solves a sorting-like problem. 
Although ERLPBoost takes much time per iteration, 
it takes fewer iterations than C-ERLPBoost 
in practical applications. 
For this reason, 
ERLPBoost is faster than C-ERLPBoost. 
Our primary motivation is to investigate boosting algorithms 
with provable iteration bounds, which perform as fast as LPBoost.

This paper has two contributions. 
Our first contribution is to give 
a unified view of boosting for soft margin optimization. 
We show that LPBoost, ERLPBoost, and C-ERLPBoost are 
instances of the Frank-Wolfe algorithm. 

Our second contribution is to propose
a generic scheme for boosting based on the unified view.
Our scheme combines a standard Frank-Wolfe algorithm and \emph{any} algorithm 
and switches one to the other at each iteration in a non-trivial way.
We show that this scheme guarantees 
the same convergence rate, $O(\ln(m/\nu) / \epsilon^2)$,  
as ERLPBoost and C-ERLPBoost.
One can incorporate any update rule to this scheme
without losing the convergence guarantee 
so that it takes advantage of better updates 
of the second algorithm in practice.
In particular, 
we propose to choose LPBoost 
as the secondary algorithm, 
and we call the resulting algorithm 
Modified LPBoost (MLPBoost). 

In experiments on real datasets, 
MLPBoost works comparably with LPBoost, and 
MLPBoost is the fastest 
among theoretically guaranteed algorithms, as expected.

Table~\ref{table:boosting_comparison} compares 
LPBoost, ERLPBoost, C-ERLPBoost, and MLPBoost. 
\begin{table}[t]
    \centering
    \caption{%
        Comparison of the boosting algorithms. %
        C-ERLPBoost solves the problem per iteration %
        by sorting based algorithm, while our work and %
        LPBoost solves linear programming (LP). %
        ERLPBoost solves convex programming (CP) per iteration. %
        In practice, the algorithms work fast in the order %
        LPBoost, ERLPBoost, and C-ERLPBoost. %
        As we show in section~\ref{sec:experiments}, %
        our algorithm is as fast as LPBoost. %
    }
    \label{table:boosting_comparison}
    \begin{tabular}{|c|cccc|}
        \toprule
                    & LPBoost & C-ERLPBoost & ERLPBoost & One of our work \\
        \midrule
        Iter. bound 
            & $\Omega(m)$ 
            & $O\left(\frac{1}{\epsilon^2} \ln \frac{m}{\nu}\right)$ 
            & $O\left(\frac{1}{\epsilon^2} \ln \frac{m}{\nu}\right)$ 
            & $O\left(\frac{1}{\epsilon^2} \ln \frac{m}{\nu}\right)$ \\
        Problem per iter. & LP & Sorting & CP & LP \\
        \bottomrule
    \end{tabular}
\end{table}

    \section{Basic definitions}
\label{sec:preliminary}
This paper considers binary classification boosting. 
We use the same notations as in~\citep{shalev-shwartz+:jml10}. 
Let $S := ((\bm{x}_i, y_i))_{i=1}^m \in (\xset \times \{\pm 1\})^m$ be 
a sequence of $m$-examples, where $\xset$ is some set. 
Let $\hset \subset [-1, +1]^\xset$ be a set of hypotheses. 
For simplicity, we assume $|\hset| = |\{h_1, h_2, \dots, h_n\}| = n$. 
Note that our scheme can use an infinite set $\hset$. 
It is convenient to regard each $h_j \in \hset$ as 
a canonical basis vector $\bm{e}_j \in \mathbb{R}^n$. 
Let $A = (y_i h_j(\bm x_i))_{i, j} \in [-1, +1]^{m \times n}$ 
be a matrix of size $m \times n$. 
We denote $m$-dimensional capped probability simplex as 
$\psimplex{m}_{\nu} := \{ \bm{d} \in [0, 1/\nu]^m \mid \| \bm{d} \|_1 = 1\}$, 
where $\nu \in [1, m]$. 
We write $\psimplex{m} = \psimplex{m}_{1}$ for shorthand. 
For a set $\cset \subset \mathbb R^n$, 
we denote the convex hull of $\cset$ as 
$
    \convhull (\cset) 
        := \left\{
            \sum_{k} w_k \bm{s}_k
            \mid \sum_k w_k = 1, w_k \geq 0,
            \{\bm{s}_k\}_k \subset \cset
        \right\}
$.

Next, we define some properties for convex functions. 
\begin{def.}[smooth function]
    A function $f : \mathbb R^m \to \mathbb R$ is 
    said to be $\eta$-smooth 
    over a convex set $\cset \subset \mathbb R^m$ 
    w.r.t. a norm $\| \cdot \|$ if 
    \begin{align}
        \label{eq:smoothness}
        \forall \bm{x}, \bm{y} \in \cset, \quad
        f(\bm{y}) 
        \leq f(\bm{x}) + (\bm{y} - \bm{x})^\top \nabla f(\bm{x})
        + \frac{\eta}{2} \|\bm{y} - \bm{x}\|^2.
    \end{align}
\end{def.}

Similarly, we define the strongly convex function. 
\begin{def.}[strongly convex function]
    A function $f : \mathbb R^m \to \mathbb R$ is 
    said to be $\eta$-strongly convex 
    over a convex set $C \subset \mathbb R^m$ 
    w.r.t. a norm $\| \cdot \|$ if 
    \begin{align}
        \nonumber
        \forall \bm{x}, \bm{y} \in \cset, \quad
        f(\bm{y}) 
        \geq f(\bm{x}) + (\bm{y} - \bm{x})^\top \nabla f(\bm{x})
        + \frac{\eta}{2} \|\bm{y} - \bm{x}\|^2.
    \end{align}
\end{def.}

We also define Fenchel conjugate. 
\begin{def.}[Fenchel conjugate]
    The Fenchel conjugate $f^\star$ of 
    a function $f : \mathbb{R}^m \to [-\infty, +\infty]$ is defined as
    \begin{align*}
        f^\star(\bm{\theta}) = \sup_{\bm{d} \in \mathbb{R}^m} 
        \left( \bm{d}^\top \bm{\theta} - f(\bm{d}) \right).
    \end{align*}
\end{def.}
It is well known that if $f$ is a $1/\eta$-strongly convex function 
w.r.t. the norm $\|\cdot\|$ for some $\eta > 0$, 
$f^\star$ is an $\eta$-smooth function
w.r.t. the dual norm $\|\cdot\|_\star$. 
Further, if $f$ is a strongly convex function, 
the gradient vector of $f^\star$ is written as
\begin{align}
    \nonumber
    \nabla f^\star(\bm{\theta}) = 
    \arg \sup_{\bm{d} \in \mathbb{R}^m}
    \left( \bm{d}^\top \bm{\theta} - f(\bm{d}) \right).
\end{align}
One can find the proof of these properties 
here~\citep{borwein+:springer06,shalev-shwartz+:jml10}.

\begin{lem.}
    \label{lem:fenchel_conjugate_functions}
    Let $f, \tilde{f} : \mathbb{R}^m \to (-\infty, +\infty]$ be 
    functions such that 
    \begin{align*}
        \exists c > 0, \forall \bm{\theta},
        f(\bm{\theta}) \leq \tilde{f}(\bm{\theta}) \leq f(\bm{\theta}) + c.
    \end{align*}
    Then, 
    $
        f^\star(\bm{\mu}) - c
        \leq \tilde{f}^\star(\bm{\mu})
        \leq f^\star(\bm{\mu})
    $ holds for all $\bm{\mu}$.
\end{lem.}
Finally, we show the duality theorem~\citep{borwein+:springer06}.
\begin{thm.}[\cite{borwein+:springer06}]
    \label{thm:strong_duality}
    Let $f : \mathbb{R}^m \to (-\infty, +\infty]$ and 
    $g : \mathbb{R}^n \to (-\infty, +\infty]$ be convex functions, 
    and a linear map $A : \mathbb{R}^m \to \mathbb{R}^n$. 
    Define the Fenchel problems
    \begin{align}
        \label{eq:strong_duality_primal}
        \gamma & = \inf_{\bm{d}} f(\bm{d}) + g(A^\top\bm{d}), \\
        \label{eq:strong_duality_dual}
        \rho   & = \sup_{\bm{w}} -f^\star(-A \bm{w}) - g^\star(\bm{w}).
    \end{align}
    Then, $\gamma \geq \rho$ holds. 
    Further, $\gamma = \rho$ holds if~\footnote{%
        For a set $\cset \subset \mathbb{R}^n$, %
        $
            \interior(C) = \{
                \bm{w} \in C \mid
                    \forall \bm{v} \in \mathbb{R}^n,
                    \exists t > 0,
                    \forall \tau \in [0, t], \bm{w} + \tau \bm{v} \in C
            \}
        $ and \\ %
        $
            \domain g - A^\top \domain f
            = \{ \bm{w} - A^\top \bm{d} \mid
                \bm{w} \in \domain g,
                \bm{d} \in \domain f
            \}
        $. %
    }
    $\bm{0} \in \interior \left(\domain g - A^\top \domain f\right)$. 
    Furthermore, points $\bar{\bm{d}} \in \psimplex{m}_{\nu}$ and 
    $\bar{\bm{w}} \in \psimplex{n}$ are optimal solutions 
    for problems~(\ref{eq:strong_duality_primal}) 
    and~(\ref{eq:strong_duality_dual}), respectively, 
    if and only if $-A \bar{\bm{w}} \in \partial f(\bar{\bm{d}})$ 
    and $\bar{\bm{w}} \in \partial g(A^\top \bar{\bm{d}})$.
\end{thm.}

With these notations, 
we define the soft margin optimization problem 
as the dual problem of the edge minimization problem. 
The edge minimization problem is defined as 
\begin{align}
    \label{eq:edge_minimization}
    \min_{\bm{d}}
        \max_{j \in [n]} (\bm{d}^\top A)_j
        + f(\bm{d}),
    \quad
    \text{where}
    \quad
    f(\bm{d}) = 
    \begin{cases}
        0 & \bm{d} \in \psimplex{m}_{\nu} \\
        +\infty & \bm{d} \notin \psimplex{m}_{\nu}
    \end{cases}.
\end{align}
The quantity $(\bm{d}^\top A)_j = \sum_{i=1}^m d_i y_i h_j(\bm{x}_i)$ is 
often called the \emph{edge} of the hypothesis $h_j$ 
w.r.t. the distribution $\bm{d} \in \psimplex{m}_{\nu}$. 
\cite{shalev-shwartz+:jml10} showed that 
the $\ell_1$-norm regularized soft margin optimization problem is 
formulated via Fenchel duality as 
\begin{align}
    \label{eq:soft_margin_maximization}
    \max_{\bm w \in \psimplex{n}} - f^\star(- A\bm w)
    = \max_{\bm w \in \psimplex{n}} \min_{\bm{d} \in \psimplex{m}_{\nu}}
    \bm{d}^\top A \bm{w}.
\end{align}
Furthermore, the duality gap between~(\ref{eq:edge_minimization}) 
and~(\ref{eq:soft_margin_maximization}) is zero. 
The soft margin optimization aims 
to find an optimal combined hypothesis 
$H_{T} = \sum_{j=1}^n \bar{w}_j h_j$, 
where $\bar{\bm{w}} \in \psimplex{n}$ is 
an optimal solution of~(\ref{eq:soft_margin_maximization}). 
Although the edge minimization and soft margin optimization problems are 
formulated as a linear program, 
solving the problem for a huge class $\hset$ is hard. 
Boosting is a standard approach to dealing with the problem.

\subsection{Boosting}
Boosting is a protocol between two algorithms; 
the booster and the weak learner. 
For each iteration $t = 0, 1, 2, \dots, T$, 
the booster chooses a distribution $\bm{d}_t \in \psimplex{m}_{\nu}$ 
over the training examples $S$. 
Then, the weak learner returns a hypothesis $h_{j_{t+1}} \in \hset$ 
to the booster that satisfies $({\bm{d}_t}^\top A)_{j_{t+1}} \geq g$ 
for some unknown guarantee $g > 0$. 
The boosting algorithm aims to produce a convex combination 
$H_T = \sum_{t=1}^T w_{T,j_t} h_{j_t}$ 
of the hypotheses $\{h_{j_1}, h_{j_2}, \dots, h_{j_T}\} \subset \hset$ 
that satisfies 
\begin{align}
    \label{eq:boosting_goal}
    -f^\star ( - A \bm{w}_T )
    = \min_{\bm{d} \in \psimplex{m}_{\nu}} 
    \bm{d}^\top A \bm{w}_T
    = \min_{\bm{d} \in \psimplex{m}_{\nu}} 
    \sum_{i=1}^m d_i y_i H_T(\bm{x}_i)
    \geq g - \epsilon
\end{align}
for any predefined $\epsilon > 0$. 
Suppose that the weak learner always returns a max-edge hypothesis 
for any given distribution $\bm{d} \in \psimplex{m}_{\nu}$. 
In that case, 
the goal is to find an $\epsilon$-approximate solution 
of~(\ref{eq:soft_margin_maximization}). 

\subsection{The Frank-Wolfe algorithms}
We briefly introduce the standard Frank-Wolfe algorithm. 
The original Frank-Wolfe (FW) algorithm is 
a first-order iterative algorithm 
invented by~\cite{marguerite+:nrl56}. 
The FW algorithm solves the problems of the form:
$\min_{\bm{x} \in \cset} f(\bm{x})$, 
where $\cset \subset \mathbb R^m$ is a closed convex set 
and $f : \cset \to \mathbb R$ is an $\eta$-smooth and convex function. 

In each iteration $t$, 
the FW algorithm seeks an extreme point 
$
    \bm{s}_{t+1}
    \in \arg \min_{\bm{s} \in \cset} \bm{s}^\top \nabla f(\bm{x}_t)
$. 
Then, it updates the iterate as 
$\bm{x}_{t+1} = \bm{x}_t + \lambda_t (\bm{s}_{t+1} - \bm{x}_t)$ 
for some $\lambda_t \in [0, 1]$. 
Although the classical result~\citep{marguerite+:nrl56,jaggi:icml13}
suggests $\lambda_t = 2/(t+2)$, 
$\lambda_t$ has many choices. 
For example, one can choose $\lambda_t$ as
\begin{align}
    \label{eq:short_step_update}
    \lambda_t := 
        \clip
        \frac{(\bm{x}_t - \bm{s}_{t+1})^\top \nabla f(\bm{x}_t)}
             {\eta \|\bm{s}_{t+1} - \bm{x}_t\|^2},
\end{align}
where $\clip x = \max\{0, \min\{1, x\}\}$. 
This optimal solution minimizes 
the right-hand side of the inequality~(\ref{eq:smoothness}) 
and is often called the \emph{short-step} strategy.
Alternatively, one can choose 
$
    \lambda_t \in 
    \arg\min_{\lambda \in [0, 1]} 
    f(\bm{x}_t + \lambda (\bm{s}_{t+1} - \bm{x}_t))
$ by line search. 
This step size improve the objective more than the short-step strategy. 
Since the FW algorithm aims to find an optimal solution, 
one can choose $\bm{x}_{t+1}$ by solving the problem:
$
    \bm{x}_{t+1} \gets
    \arg \min_{
        \bm{x} \in \mathop{\rm CH}(\{\bm{s}_1, \dots, \bm{s}_{t+1}\})
    } f(\bm x)
$. 
This update rule is 
called the \emph{Fully Corrective} FW algorithm~(e.g., \cite{jaggi:icml13}). 
Although the fully corrective update yields $\bm{x}_{t+1}$ 
that most decreases the objective over the convex hull, 
it loses the fast computational advantage per iteration.

The FW algorithms converge to 
an $\epsilon$-approximate solution in $O(\eta/\epsilon)$ iterations 
if the objective function is $\eta$-smooth w.r.t. 
some norm over $\cset$~\citep{jaggi:icml13,marguerite+:nrl56}. 
The best advantage of the FW algorithm is the projection-free property; 
there is no projection onto $\cset$, 
so the running time per iteration is 
faster than the projected gradient methods.

    \section{Related work}
\label{sec:related_work}
FWBoost~\citep{wang+:arxiv15} seems to be related to our work.
FWBoost is a boosting algorithm designed for 
minimizing a general loss function by the Frank-Wolfe algorithm. 
Since the Frank-Wolfe algorithm works over the closed convex set, 
they introduce the $\ell_1$-norm ball constraint. 
Note that the objective function for 
the maximization problem~(\ref{eq:soft_margin_maximization}) is 
not a strongly-smooth function, so 
one cannot apply the FWBoost directly to guarantee the convergence rate. 

LPBoost~\citep{demiriz+:ml02} is a practical boosting algorithm 
for solving problem (\ref{eq:soft_margin_maximization}). 
In each iteration $t$, LPBoost updates its distribution as 
an optimal solution to problem 
\begin{align}
    \label{eq:lpb_update}
    \min_{\bm{d}}
        \max_{k \in [t]} (\bm{d}^\top A)_{j_k}
        + f(\bm d).
\end{align}
That is, LPBoost uses an optimal solution to 
the edge minimization problem over the hypothesis set 
$\{h_{j_1}, h_{j_2}, \dots, h_{j_t}\} \subset \hset$. 
LPBoost converges to an $\epsilon$-accurate solution rapidly in practice. 
However,~\cite{warmuth+:nips07} proved that LPBoost converges 
in $\Omega(m)$ iterations for the worst case. 
After that, the stabilized version of LPBoost, 
ERLPBoost, was invented by~\cite{warmuth+:alt08}. 
ERLPBoost updates the distribution as the solution of 
\begin{align}
    \label{eq:erlpb_update}
    \min_{\bm{d}}
        \max_{k \in [t]} (\bm{d}^\top A)_{j_k}
        + f(\bm d)
        + \frac 1 \eta \Delta(\bm{d}).
\end{align}
Here, $\Delta(\bm{d}) = \sum_{i=1}^m d_i \ln d_i + \ln m$ is 
the relative entropy 
from the uniform distribution $\frac 1 m \bm{1} \in \psimplex{m}_{\nu}$. 
They proved that ERLPBoost finds a solution 
that achieves~(\ref{eq:boosting_goal}) in 
$O(\ln (m/\nu) / \epsilon^2)$ iterations. 
They also demonstrate that ERLPBoost tends to terminate 
in fewer iterations than LPBoost. 
The disadvantage of ERLPBoost is its computational complexity; 
ERLPBoost solves convex programs in each iteration. 
This disadvantage leads to much more computation time than LPBoost. 
C-ERLPBoost~\citep{shalev-shwartz+:jml10} 
is the corrective version of ERLPBoost. 
This algorithm achieves the same iteration bound 
with much faster computation per iteration than LPBoost. 
C-ERLPBoost maintains 
the weight $\bm{w}_t \in \psimplex{n}$ that only has non-zero values on
$\{ w_{t, j_1}, w_{t, j_2}, \dots, w_{t, j_t}\}$ 
corresponding to the past hypotheses 
$\{h_{j_1}, h_{j_2}, \dots, h_{j_t}\} \subset \hset$.
C-ERLPBoost updates its distribution over the training instances as 
\begin{align}
    \label{eq:cerlpb_update}
    \bm{d}_t \gets
    \arg \min_{\bm{d}}
    \bm{d}^\top A \bm{w}_t + f(\bm{d}) + \frac 1 \eta \Delta(\bm d). 
\end{align}
After receiving a hypothesis $h_{j_{t+1}} \in \hset$ 
with the corresponding basis $\bm{e}_{j+1} \in \psimplex{n}$, 
C-ERLPBoost updates the weights on hypotheses as 
$\bm{w}_{t+1} = \bm{w}_t + \lambda (\bm{e}_{j_{t+1}} - \bm{w}_t)$, 
where $\lambda_t \in [0, 1]$ is some proper value. 
Although this update rule seems to be a convex program, 
\cite{shalev-shwartz+:jml10} showed an algorithm 
that solves~(\ref{eq:cerlpb_update}) in $O(m \ln m)$ time. 
This algorithm seems better than LPBoost and ERLPBoost. 
However, \cite{warmuth+:alt08} demonstrated that C-ERLPBoost takes 
much more iterations than LPBoost and ERLPBoost. 
Therefore, the overall computation time is worse than LPBoost.

    \section{Main results}
\label{sec:main_result}
We first show 
a unified view of the boosting algorithms 
via Fenchel duality. 
From this view, LPBoost, ERLPBoost, and C-ERLPBoost can be seen as 
instances of the Frank-Wolfe algorithm 
with different step sizes and objectives. 
Using this knowledge, we derive a new boosting scheme. 
\subsection{A unified view of boosting for the soft margin optimization}
This section assumes that the weak learner 
always returns a hypothesis $h \in \hset$ 
that maximizes the edge w.r.t. the given distribution. 
We start by revisiting C-ERLPBoost. 
Recall that C-ERLPBoost (and ERLPBoost) aim to solve the convex program 
\begin{align}
    \label{eq:smoothed_problem}
    \min_{\bm{d}}
    \max_{j \in [n]} (\bm{d}^\top A)_j + \tilde{f}^\star (\bm{d}),
\end{align}
where $\tilde{f} = f + \frac{1}{\eta} \Delta$. 
Since $\frac{1}{\eta} \Delta$ is 
a $\frac{1}{\eta}$-strongly convex function w.r.t. $\ell_1$-norm, 
so does $\tilde{f}$. 
By Fenchel duality, the dual problem is 
\begin{align}
    \label{eq:smoothed_softmargin}
    \max_{\bm{w} \in \psimplex{n}} - \tilde{f}^\star( -A \bm{w} )
    = - \min_{\bm{w} \in \psimplex{n}}
        \tilde{f}^\star( -A \bm{w} )
    = - \min_{\bm{\theta} \in -A\psimplex{n}}
        \tilde{f}^\star(\bm{\theta}),
\end{align}
where $-A \psimplex{n} = \{ -A \bm{d} \mid \bm{d} \in \psimplex{n}\}$, 
with zero duality gap. 
Further, $\tilde{f}^\star$ is an $\eta$-smooth function 
w.r.t. $\ell_\infty$-norm. 
Thus, the soft margin optimization problem becomes 
a minimization problem of a smooth function. 

In each iteration $t$, 
C-ERLPBoost updates the distribution $\bm{d}_t \in \psimplex{m}_{\nu}$ 
over examples as the optimal solution of~(\ref{eq:cerlpb_update}). 
This computation corresponds to the gradient computation 
$\nabla \tilde{f}^\star( \theta_t )$, where $\theta_t = - A\bm{w}_t$. 
Then, obtain a basis vector $\bm{e}_{j_{t+1}} \in \psimplex{n}$ 
corresponding to hypothesis $h_{j_{t+1}} \in \hset$ 
that maximizes the edge; 
$j_{t+1} \in \arg \max_{j \in [n]} (\bm{d}_t^\top A)_j$. 
We can write this calculation regarding the gradient of $\tilde{f}^\star$;
\begin{align*}
    \arg \max_{\bm{e}_j : j \in [n]} \bm{d}_t^\top A \bm{e}_j
    = \arg \min_{\bm{e}_j : j \in [n]}
        (-A\bm{e}_j)^\top \nabla \tilde{f}^\star (\bm{\theta}_t)
    = \arg \min_{\bm{\theta} \in -A\psimplex{n}}
        \bm{\theta}^\top \nabla \tilde{f}^\star (\bm{\theta}_t).
\end{align*}
Thus, finding a hypothesis that maximizes edge corresponds to 
solving linear programming in the Frank-Wolfe algorithm. 
Further, C-ERLPBoost updates the weights as 
$\bm{w}_{t+1} = \bm{w}_t + \lambda_t (\bm{e}_{j_{t+1}} - \bm{w}_t)$, 
where $\lambda_t$ is the short-step~\footnote{%
    They also suggests the line search update. %
    This case yields a better progress than short-step, %
    so the same iteration bound holds. %
}, as in eq.~(\ref{eq:short_step_update}).
From these observations, we can say that the C-ERLPBoost is 
an instance of the Frank-Wolfe algorithm. 
Since $\tilde{f}^\star$ is $\eta$-smooth, we can say that 
this algorithm converges in $O(\eta/\epsilon)$ iterations 
for a max-edge weak learner. 

Similarly, we can say that LPBoost and ERLPBoost are 
instances of the Frank-Wolfe algorithm. 
Let $J_t := \{j_1, j_2, \dots, j_{t} \}$ be the set of indices 
corresponding to the hypotheses $\{h_{j_1}, h_{j_2}, \dots, h_{j_{t}}\}$ 
and $\eset_t := \{ \bm{e}_{j} \mid j \in J_t \}$ be 
the corresponding basis vectors. 
LPBoost and ERLPBoost update the distribution as 
the optimal solutions 
$\bm{d}_t^{\rm{L}} \in \partial f^\star( -A \bm{w}_t^{\rm L})$ 
and $\bm{d}_t^{\rm{E}} = \nabla \tilde{f}^\star (-A \bm{w}_t^{\rm E})$, 
where 
\begin{alignat}{2}
    \label{eq:lpb_weight}
    \text{(LPBoost)} & \quad &
        \bm{w}_t^{\rm{L}} \gets \arg \max_{\bm{w} \in \convhull(\eset_t)}
        - f^\star (-A \bm{w}), \\
    \label{eq:erlpb_weight}
    \text{(ERLPBoost)} & \quad &
        \bm{w}_t^{\rm{E}} \gets \arg \max_{\bm{w} \in \convhull(\eset_t)}
        - \tilde{f}^\star (-A \bm{w}).
\end{alignat}
Therefore, we can say that LPBoost and ERLPBoost are 
instances of the fully-corrective FW algorithm 
for objectives $f^\star$ and $\tilde{f}^\star$, respectively. 
Under the max-edge weak learner assumption, 
one can derive the same iteration bound for ERLPBoost 
since $\tilde{f}^\star$ is $\eta$-smooth. 
We summarize these connections to the following theorem. 
\begin{thm.}
    LPBoost, ERLPBoost, and C-ERLPBoost are instances of the FW algorithm. 
\end{thm.}
\subsection{Generic schemes for margin-maximizing boosting}
\begin{algorithm}[t]
    \caption{A theoretically guaranteed boosting scheme}
    \begin{algorithmic}[1]
        \REQUIRE{%
            Training examples 
            $
                S = \left( (\bm{x}_i, y_i) \right)_{i=1}^m
                \in (\xset \times \{\pm 1\})^m
            $, %
            a hypothesis set $\hset \subset [-1, +1]^\xset$, 
            a FW algorithm $\fwalg$, 
            a secondary algorithm $\secalg$, %
            and parameters $\nu > 0$ and $\epsilon > 0$. %
        }
        \STATE{Set $A = \left(y_i h_j(\bm{x}_i)\right)_{i, j} \in [-1, +1]^{m \times n}$.}
        \STATE{%
            Send $\bm{d}_0 = \frac{1}{m} \bm{1}$ to the weak learner %
            and obtain a hypothesis $h_{j_1} \in \hset$. %
        }

        \STATE{Set $\bm{w}_{1} = \bm{e}_{j_1}$.}

        \FOR{$t=1, 2, \dots, T$}
            \STATE{
                Compute the distribution 
                $
                    \bm{d}_t 
                    = \nabla \tilde{f}^\star(- A \bm{w}_t)
                    = \arg \min_{\bm{d} \in \psimplex{m}_{\nu}}
                    \left[
                    \bm{d}^\top A \bm{w}_t + \frac 1 \eta \Delta (\bm{d})
                    \right]
                $.
            }

            \STATE{%
                Obtain a hypothesis $h_{j_{t+1}} \in \hset$ 
                and the corresponding basis vector 
                $\bm{e}_{j_{t+1}} \in \psimplex{n}$. 
            }

            \STATE{
                Set $
                    \epsilon_t := \min_{0 \leq \tau \leq t}
                    (\bm{d}_\tau^\top A)_{j_{\tau + 1}}
                    + \tilde{f}^\star(- A\bm{w}_t)
                $ and let
                $\eset_{t+1} := \{ \bm{e}_{j_\tau} \}_{\tau=1}^{t+1}$.
            }

            \IF{$\epsilon_t \leq \epsilon / 2$}
                \STATE{Set $T = t$, \textbf{break}.}
            \ENDIF

            \STATE{
                Compute the FW weight 
                $
                    \bm{w}_{t+1}^{(1)}
                    = \fwalg
                    (
                        A, \bm{w}_{t}, \bm{e}_{j_{t+1}},
                        \eset_{t}, \bm{d}_t
                    )
                $.
            }
            \STATE{%
                Compute the secondary weight %
                $\bm{w}_{t+1}^{(2)} = \secalg(A, \eset_{t+1})$.
            }

            \STATE{
                Update the weight 
                $
                    \bm{w}_{t+1} 
                    \gets \arg \min_{\bm{w}_{t+1}^{(k)}: k \in \{1, 2\}} 
                    \tilde{f}^\star (-A\bm{w}_{t+1}^{(k)})
                $.
            }
        \ENDFOR
        \ENSURE{Combined classifier $H_T = \sum_{t=1}^T w_{T, t} h_t$.}
    \end{algorithmic}
    \label{alg:our_scheme}
\end{algorithm}

\begin{algorithm}[th]
    \begin{algorithmic}[1]
        \REQUIRE{%
            A matrix %
            $
                A = \left( y_i h_j (\bm{x}_i) \right)_{i, j}
                \in [-1, +1]^{m \times n}
            $ %
            and a set of basis vectors $\eset_{t+1} \subset \psimplex{n}$. %
        }

        \ENSURE{%
            $
                \bm{w} \gets \arg\max_{\bm{w} \in \convhull (\eset_{t+1})} 
                    \min_{\bm d \in \psimplex{m}_{\nu}}
                    \bm{d}^\top A \bm{w}
            $.
        }
    \end{algorithmic}
    \caption{LPBoost rule $\secalg (A, \eset_{t+1})$}
    \label{alg:lpb_subroutine}
\end{algorithm}
%
%
\begin{algorithm}[th]
    \begin{algorithmic}[1]
        \ENSURE{
            $
                \bm{w}_{t+1}^{(1)} 
                = \bm{w}_t + \lambda_t (\bm{e}_{j_{t+1}} - \bm{w}_t)
            $, where 
            $
                \lambda_t = 
                \clip
                \frac
                    {\bm{d}_t^\top A (\bm{e}_{j+1} - \bm{w}_t)}
                    {\eta \| A (\bm{e}_{j+1} - \bm{w}_t) \|_{\infty}^2}
            $.
        }
    \end{algorithmic}
    \caption{Short step rule
        $
            \fwalg
            (
                A, \bm{w}_t, \bm{e}_{j_{t+1}},
                \eset_{t}, \bm{d}_t
            )
        $
    }
    \label{alg:ss_rule}
\end{algorithm}
We propose FW-like boosting schemes 
from the above observations, 
shown in Algorithm~\ref{alg:our_scheme}. 
Algorithm~\ref{alg:our_scheme} takes two update rules, 
a FW update rule $\fwalg$ and 
a secondary update rule $\secalg$. 
Both algorithms return a weight $\bm{w} \in \psimplex{n}$. 
Intuitively, the FW update rule $\bm{w}_t^{(1)}$ is 
a safety net for the convergence guarantee. 
Further, the convergence analysis only depends on 
the FW update $\bm{w}_{t+1}^{(1)}$, 
so that one can incorporate any update rule to $\secalg$. 
For example, one can use the update rule~(\ref{eq:erlpb_weight}) 
as $\secalg(A, \eset_{t+1})$. 
Algorithm~\ref{alg:our_scheme} becomes 
ERLPBoost in this case 
since $\bm{w}_{t+1} = \bm{w}_{t+1}^{(2)}$ holds for any $t$. 
Even though this setting, the convergence guarantee holds 
so that we can prove the same convergence rate for ERLPBoost 
by our general analysis. 

Recall that our primary objective is to find a weight vector $\bm{w}$ 
that optimizes the linear program~(\ref{eq:soft_margin_maximization}). 
The most practical algorithm, LPBoost, 
solves the optimization problem over past hypotheses, 
so using the solution as $\secalg$ is a natural choice. 
Algorithm~\ref{alg:lpb_subroutine} summarizes this update. 
Note that the LPBoost update 
differs from the fully-corrective FW algorithm 
since the objective function is $\tilde{f}^\star$, not $f^\star$. 

Furthermore, as described in~\citep{shalev-shwartz+:jml10}, 
one can compute the distribution 
$\bm{d}_t = \nabla \tilde{f}^\star (-A \bm{w}_t)$ 
by a sorting-based algorithm, 
which takes $O(m \ln m)$ iterations\footnote{%
They also suggest a linear time algorithm, %
see~\citep{herbster+:jmlr01}. %
}. 
Thus, the time complexity per iteration depends on 
the secondary algorithm $\secalg$.

Before getting into the convergence analysis, 
we first justify the stopping criterion 
in Algorithm~\ref{alg:lpb_subroutine}. 
This criterion is similar to the one in C-ERLPBoost 
but better than it. 
Therefore, our algorithms tend to converge in early iterations. 
\begin{lem.}
    \label{lem:stopping_criterion}
    Let 
    $
        \epsilon_t 
        := \min_{0 \leq \tau \leq t} (\bm{d}_{\tau}^\top A)_{j_{\tau + 1}}
        + \tilde{f}^\star( -A \bm{w}_t)
    $ be the optimality gap defined in Algorithm~\ref{alg:lpb_subroutine} 
    and let $\eta = 2 \ln(m / \nu) / \epsilon$. 
    Then, $\epsilon_t \leq \epsilon / 2$ implies 
    $- f^\star (- A \bm{w}_t ) \geq g - \epsilon$. 
\end{lem.}
\begin{proof}
    By the weak-learnability assumption, 
    $\epsilon_t \geq g + \tilde{f}^\star ( -A \bm{w}_t )$. 
    The statement follows from Lemma~\ref{lem:fenchel_conjugate_functions}.
\end{proof}
Now, we prove the convergence rate for our scheme. 
This theorem shows the same convergence guarantee for 
ERLPBoost and C-ERLPBoost. 
\begin{thm.}[A convergence rate for Algorithm~\ref{alg:our_scheme}]
    \label{thm:mlpboost_convergence_guarantee}
    Assume that the weak learner returns 
    a hypothesis $h_{j_{t+1}} \in \hset$ that satisfies
    $(\bm{d}_t^\top A)_{j_{t+1}} \geq g$ 
    for some unknown guarantee $g$. 
    Let $\fwalg$ be a FW update with 
    classic step $\lambda_t = \frac{2}{t+2}$, or
    short-step as in Algorithm~\ref{alg:ss_rule}. 
    Then, for any secondary algorithm $\secalg$, 
    Algorithm~\ref{alg:our_scheme} converges to 
    an $\epsilon$-accurate solution of~(\ref{eq:soft_margin_maximization}) 
    in $O\left( \frac{1}{\epsilon^2} \ln \frac{m}{\nu} \right)$ iterations. 
\end{thm.}
\begin{proof}
    First of all, we prove the bound for the classic step size. 
    We start by showing the recursion 
    \begin{align}
        \label{eq:convergence_proof_0}
        \epsilon_{t+1}
        \leq (1 - \lambda_t) \epsilon_t + 2 \eta \lambda_t^2.
    \end{align}
    By using the definition of $\bm{w}_{t+1}$ 
    and the $\eta$-smoothness of $\tilde{f}^\star$, 
    \begin{align}
        \epsilon_t - \epsilon_{t+1}
            & \geq \tilde{f}^\star ( -A \bm{w}_t )
                - \tilde{f}^\star ( -A \bm{w}_t^{(1)} ) \nonumber \\
            & = \tilde{f}^\star ( -A \bm{w}_t )
                - \tilde{f}^\star (
                    -A \bm{w}_t + \lambda_t A(\bm{w}_t - \bm{e}_{{j_t+1}})
                ) \nonumber \\
            \label{eq:convergence_proof_1}
            & \geq \lambda_t ( A (\bm{e}_{j_{t+1}} - \bm{w}_{t}) )^\top
                \nabla \tilde{f}^\star (- A \bm{w}_t) - 2\eta \lambda_t^2,
    \end{align}
    where eq.~(\ref{eq:convergence_proof_1}) holds since 
    $A \in [-1, +1]^{m \times n}$ 
    and $\bm{e}_{j_{t+1}}, \bm{w}_t \in \psimplex{n}$. 
    By the non-negativity of the entropy function 
    and the definition of $\bm{d}_t$, we get
    \begin{align}
            \lambda_t ( A (\bm{e}_{j_{t+1}} - \bm{w}_{t}) )^\top
                \nabla \tilde{f}^\star (- A \bm{w}_t)
            & = \lambda_t \bm{d}_t^\top A (\bm{e}_{j_{t+1}} - \bm{w}_{t})
                \nonumber \\
            & \geq \lambda_t
                \left[
                    \min_{0 \leq \tau \leq t} 
                    (\bm{d}_\tau^\top A)_{j_{\tau + 1}} 
                    - \bm{d}_{t}^\top A \bm{w}_t 
                    - \frac{1}{\eta} \Delta (\bm{d}_t)
                \right]
                \nonumber \\
            \label{eq:convergence_proof_2}
            & = \lambda_t
                \left[
                    \min_{0 \leq \tau \leq t} 
                    (\bm{d}_\tau^\top A)_{j_{\tau + 1}} 
                    + \tilde{f}^\star( - A\bm{w}_t )
                \right]
            = \lambda_t \epsilon_t.
    \end{align}
    Combining eq.~(\ref{eq:convergence_proof_1}) 
    and~(\ref{eq:convergence_proof_2}), 
    we obtain~(\ref{eq:convergence_proof_0}). 

    Now, we prove the following inequality by induction on $t$.
    \begin{align}
        \label{eq:convergence_proof_3}
        \epsilon_t \leq \frac{8 \eta}{t + 2}, 
        \quad \forall t = 1, 2, \dots
    \end{align}
    For the base case $t = 1$, 
    the inequality~(\ref{eq:convergence_proof_3}) holds; 
    $
        \epsilon_1
        \leq (1 - \lambda_0) \epsilon_0 + 2 \eta \lambda_0^2
        = 2 \eta
        \leq \frac{8\eta}{1 + 2}
    $. Assume that~(\ref{eq:convergence_proof_3}) holds for $t \geq 1$. 
    By the inductive assumption,
    \begin{align*}
        \epsilon_{t+1}
            \leq (1 - \lambda_t) \epsilon_t + 2 \eta \lambda_t^2 
            \leq \frac{t}{t + 2} \frac{8 \eta}{t + 2} 
                + 2 \eta \left( \frac{2}{t + 2} \right)^2 
            = 8 \eta \frac{t}{t + 2} \frac{t + 1}{t + 2}
            \leq \frac{8 \eta}{t + 3}.
    \end{align*}
    Therefore,~(\ref{eq:convergence_proof_3}) holds 
    for all $t \geq 1$.

    By the definition of $\eta$, 
    $\epsilon_T \leq \frac{\epsilon}{2}$ holds 
    after $T \geq \frac{32}{\epsilon^2} \ln \frac{m}{\nu} - 2$ 
    iterations. Lemma~\ref{lem:stopping_criterion} yields 
    the convergence rate. 

    For the short-step case, 
    that is, the case where we employ Algorithm~\ref{alg:ss_rule} as $\fwalg$, 
    we get a similar recursion:
    \begin{align}
        \epsilon_t - \epsilon_{t+1}
            & \geq \tilde{f}^\star ( -A \bm{w}_t )
                - \tilde{f}^\star ( -A \bm{w}_t^{(1)} ) \nonumber \\
            \label{eq:convergence_proof_4}
            & \geq \lambda_t ( A (\bm{e}_{j_{t+1}} - \bm{w}_{t}) )^\top
                \nabla \tilde{f}^\star (- A \bm{w}_t)
                - \frac{\eta}{2} \lambda_t^2
                \| A(\bm{w}_t - \bm{e}_{j_{t+1}}) \|_\infty^2 \\
            & \geq \lambda ( A (\bm{e}_{j_{t+1}} - \bm{w}_{t}) )^\top
                \nabla \tilde{f}^\star (- A \bm{w}_t)
                - 2\eta \lambda^2,
                \quad \quad
                \forall \lambda \in [0, 1].
                \nonumber
    \end{align}
    Optimizing $\lambda$ in RHS 
    and applying the inequality~(\ref{eq:convergence_proof_2}), 
    we get $\epsilon_t - \epsilon_{t+1} \geq \epsilon_t^2 / 8\eta$. 
    With this inequality, one can easily verify that 
    the same iteration bound~(\ref{eq:convergence_proof_3}) holds 
    for this case. 
    See the appendix for the rest proof. 
\end{proof}
Theorem~\ref{thm:mlpboost_convergence_guarantee} shows 
a convergence guarantee 
for the \emph{classic step} and the \emph{short-step}. 
The line search step 
$
    \lambda_{t} \gets \arg \min_{\lambda \in [0, 1]}
    \tilde{f}^\star \left(
        -A (\bm{w}_t + \lambda (\bm{e}_{j_{t+1}} - \bm{w}_t))
    \right)
$ always yields better progress than the \emph{short-step}, 
so the same iteration bound holds.

\paragraph{Other variants of the boosting scheme.}
%
%
\begin{algorithm}[th]
    \begin{algorithmic}[1]
        \STATE{
            Let 
            $\bm{w}_t = \sum_{\bm{e} \in E_t} \alpha_{t, \bm{e}} \bm{e}$ 
            be the current representation of $\bm{w}_t$ w.r.t. 
            the basis vectors $E_t \subset \eset_t$ 
            with positive coefficients 
            $\{ \alpha_{t, \bm{e}} \}_{\bm{e} \in E_t}$.
        }

        \STATE{
            Compute an \emph{away} basis 
            $
                \bm{e}^{\rm Away}
                \in \arg \min_{\bm{e} \in E_t} \bm{d}_t^\top A \bm{e}
            $ and set 
            $\lambda_{t, \max} = \alpha_{t, \bm{e}^{\rm Away}}$.
        }

        \STATE{
            Compute the step size 
            $
                \lambda_{t}
                \gets \arg \min_{\lambda \in [0, \lambda_{t, \max}]}
                    \tilde{f}^\star (
                        -A (
                            \bm{w}_t
                            + \lambda (\bm{e}_{t+1} - \bm{e}^{\rm Away})
                        )
                    )
            $.
        }

        \ENSURE{
            $
                \bm{w}_{t+1}^{(1)}
                = \bm{w}_t
                + \lambda_t (\bm{e}_{j_{t+1}} - \bm{e}^{\rm Away})
            $.
        }
    \end{algorithmic}
    \caption{%
        Pairwise rule %
        $
            \fwalg
            (
                A, \bm{w}_t, \bm{e}_{j_{t+1}},
                \eset_{t}, \bm{d}_t
            )
        $
    }
    \label{alg:pairwise_rule}
\end{algorithm}
The FW update rule $\bm{w}_{t+1}^{(1)}$ 
of Algorithm~\ref{alg:ss_rule} 
comes from the FW algorithm with short-step sizes. 
One can apply other updates rules as $\fwalg$. 
Pairwise Frank-Wolfe (PFW) is 
the one of a state-of-the-art Frank-Wolfe 
algorithm~\citep{lacoste-julien+:nips15}. 
The basic idea of PFW is to move the weight from 
the most worthless hypothesis to the newly attained one. 
Algorithm~\ref{alg:pairwise_rule} is the scheme 
that applies the PFW. 
By a similar argument, one can prove the convergence rate for 
Algorithm~\ref{alg:pairwise_rule}. 
\begin{cor.}[A convergence rate for Algorithm~\ref{alg:pairwise_rule}]
    Let 
    $
        k(t) = |\{
            \tau \in [t] \mid \lambda_{\tau} < \lambda_{\tau, \max}
        \}|
    $ be the number of good steps by iteration $t$. 
    Then, Algorithm~\ref{alg:pairwise_rule} converges with rate 
    $O(\eta / k(t))$. 
\end{cor.}
Note that PFW guarantees the convergence rate for a finite class $\hset$, 
while the short-step FW guarantees for all $\hset$, 
including infinite classes.

    \section{Experiments}
\label{sec:experiments}
\begin{figure}[t]
    \centering
    \begin{tabular}{cc}
        \begin{minipage}[t]{0.45\hsize}
            \centering
            \includegraphics[keepaspectratio, scale=0.5]{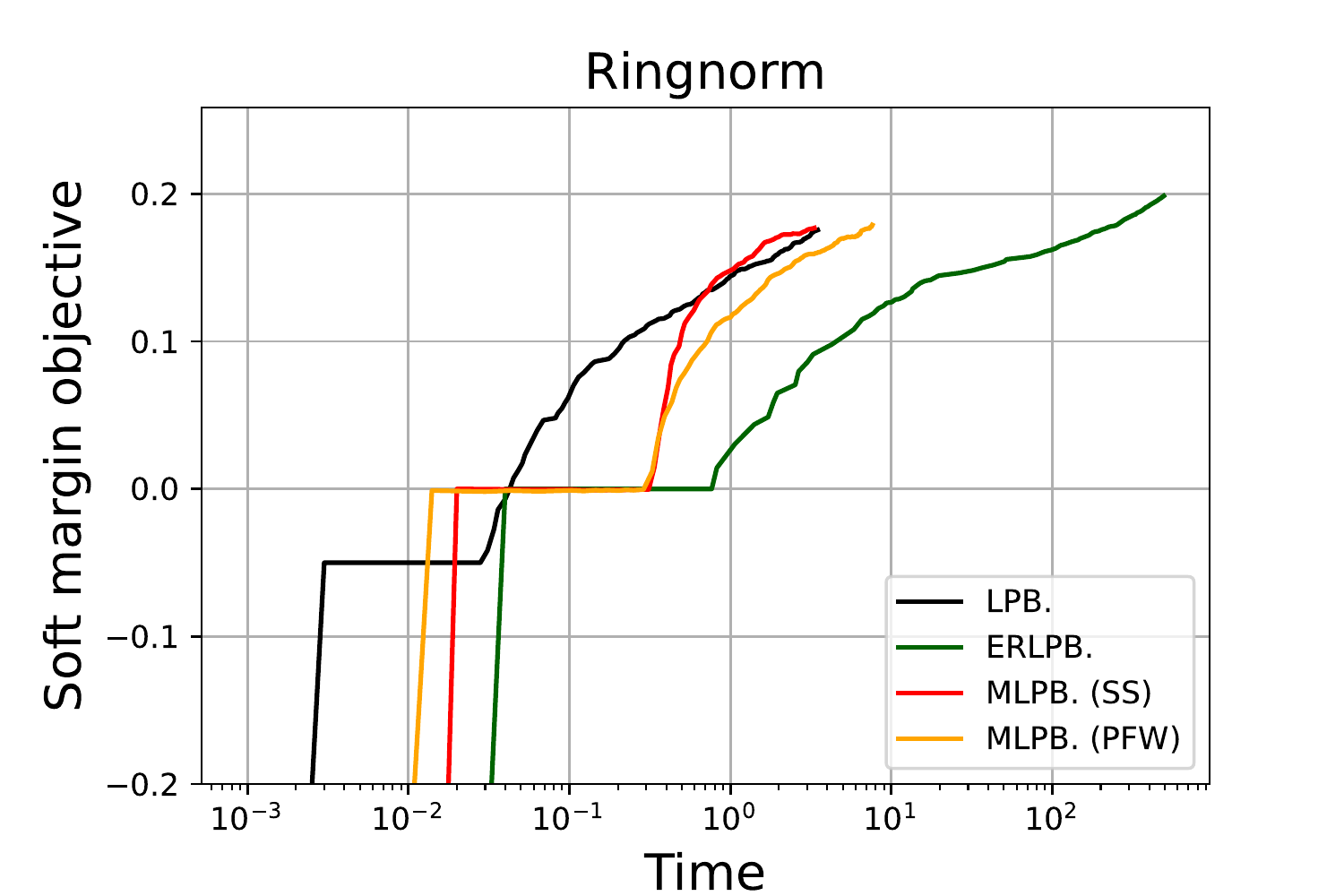}
        \end{minipage}
        &
        \begin{minipage}[t]{0.45\hsize}
            \centering
            \includegraphics[keepaspectratio, scale=0.5]{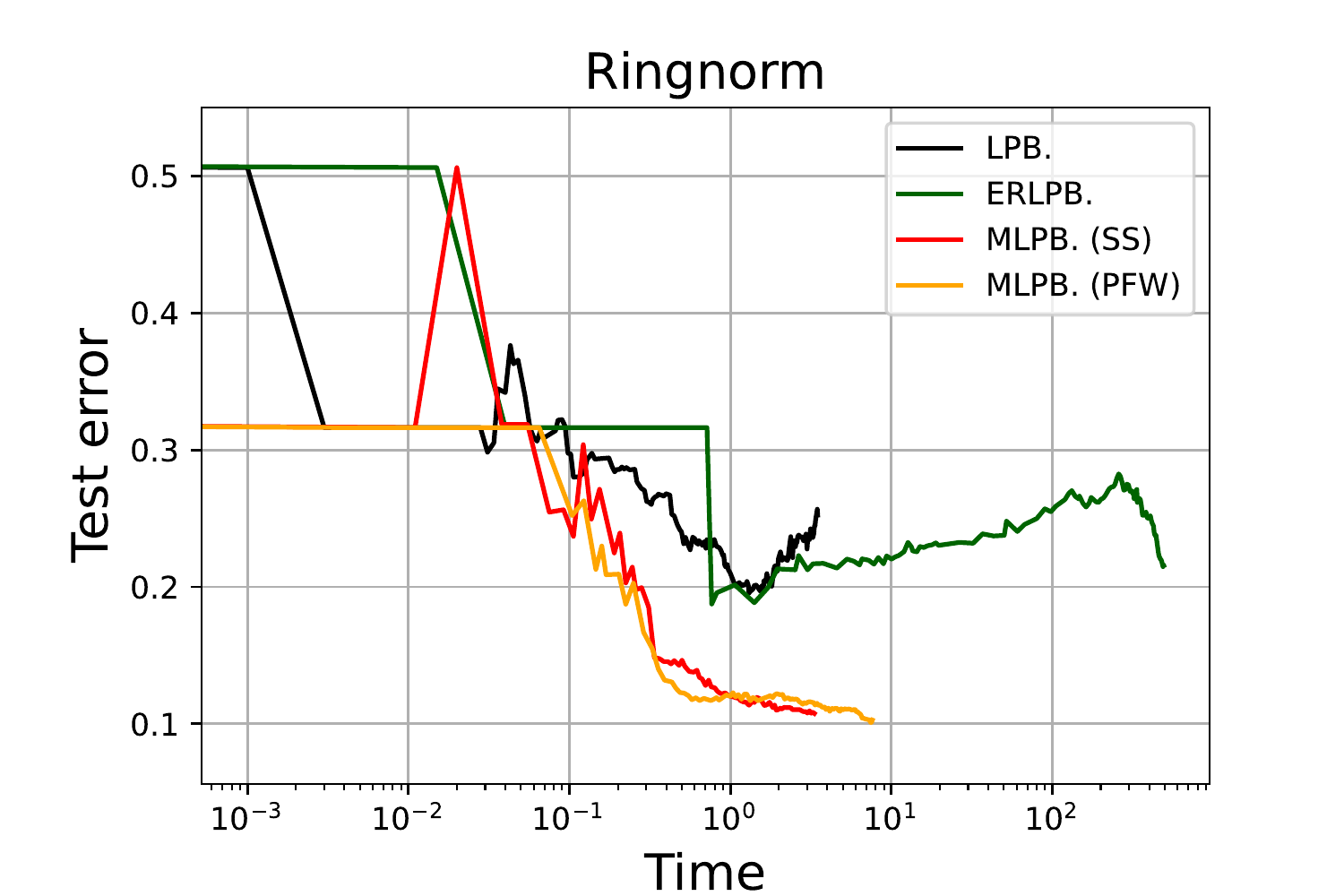}
        \end{minipage}
    \end{tabular}
    \caption{%
        Comparison of the algorithm for the 0-th fold of %
        Ringnorm dataset with parameters $\epsilon = 0.01$ %
        and $\nu = 0.1m$. %
        \emph{Left:} %
        soft margin objective value vs. computation time (seconds). %
        \emph{Right:} %
        test error vs. computation time (seconds). %
    }
\end{figure}
We compare LPBoost, ERLPBoost, and our scheme 
on Gunnar R{\"{a}}tsch's benchmark dataset~\footnote{%
    Datasets are obtained from %
    \url{http://theoval.cmp.uea.ac.uk/~gcc/matlab/default.html\#benchmarks}. %
    \label{fnote:benchmark}%
}.
We use a server with 
Intel Xeon Gold 6124 CPU 2.60GHz processors. 
We call our scheme with secondary Algorithm~\ref{alg:lpb_subroutine} 
as MLPBoost. MLPB.~(SS) and MLPB.~(PFW) are MLPBoosts 
with FW algorithms~\ref{alg:ss_rule} and~\ref{alg:pairwise_rule}, 
respectively. 
The gradient boosting algorithms, 
like XGBoost~\citep{tianqi+:kdd16} or LightGBM~\citep{ke+:nips17}, 
solve different problems, 
so we do not compare our work to them. 
Note that the FW column corresponds to C-ERLPBoost. 

In order to solve the sub-problems of 
LPBoost, ERLPBoost, and MLPBoost, 
we use the Gurobi optimizer 9.0.1~\footnote{%
    We use the Gurobi optimizer. See %
    \url{https://www.gurobi.com/}. %
}. 
\paragraph{Settings.} 
We set the capping parameters 
$\nu \in N := \{pm \mid p = 0.1, 0.2, \dots, 0.5\}$ 
and the tolerance parameter $\epsilon = 0.01$, 
where $m$ is the number of training instances. 
We use the weak learner that 
returns the best decision tree of depth 2. 
\paragraph{Computation time.} 
We measure the CPU time and the System time using 
\texttt{/usr/bin/time -v} command. 
Some algorithms do not converge in a few days, 
so we abort the experiment by \texttt{timeout 20000s} command. 
We measure the running time with capping parameters 
over $N$ for each dataset and took their average. 
Table~\ref{table:time} shows the results.
The FW column is the FW algorithm with short-steps, 
and PFW is the Pairwise FW algorithm. 
As the table shows, MLPB.~(SS) and MLPB.~(PFW) terminates 
much faster than FW and PFW, respectively. 
These results indicate that 
LPBoost rule $\secalg$, shown in Algorithm~\ref{alg:lpb_subroutine}, 
significantly improves the objective. 
See the appendix for further comparisons. 
\begin{table}[h]
    \caption{ %
        Comparison of the computation time (seconds). %
        Each cell is the average computation time over %
        the capping parameters over $N$. %
        Some algorithm does not terminate in a few hours %
        so we abort them within some appropriate time. %
    }
    \label{table:time}
    \centering
    \begin{tabular}{lrrrrrrr}
    \toprule
            & \begin{tabular}{c} Shape \end{tabular} & \begin{tabular}{c} LPB. \end{tabular} & \begin{tabular}{c} ERLPB. \end{tabular} & \begin{tabular}{c} MLPB. \\ (SS only) \end{tabular} & \begin{tabular}{c} MLPB. \\ (PFW only) \end{tabular} & \begin{tabular}{c} MLPB. \\ (SS) \end{tabular} & \begin{tabular}{c} MLPB. \\ (PFW) \end{tabular}
 \\ \midrule \addlinespace[0.5em]
    Banana  &  $(5300, 3)$ &         $168.26$ &        $3434.75$ &          $>10^4$ &          $>10^4$ &        $1418.41$ &        $1398.68$
        \\ \addlinespace[0.5em]
    B.Cancer&  $(263, 10)$ &           $3.61$ &          $73.45$ &         $180.16$ &         $270.50$ &          $23.43$ &          $19.81$
        \\ \addlinespace[0.5em]
    Diabetes&   $(768, 9)$ &          $47.53$ &        $1478.77$ &          $>10^4$ &        $3471.77$ &         $201.46$ &         $270.51$
        \\ \addlinespace[0.5em]
    F.Solar &  $(144, 10)$ &           $2.30$ &           $2.46$ &          $13.34$ &          $80.73$ &          $31.64$ &          $46.45$
        \\ \addlinespace[0.5em]
    German  & $(1000, 21)$ &          $77.56$ &        $1391.91$ &          $>10^4$ &        $5692.32$ &         $181.43$ &         $201.88$
        \\ \addlinespace[0.5em]
    Heart   &  $(270, 14)$ &          $10.03$ &         $193.58$ &          $>10^3$ &         $183.09$ &          $44.11$ &          $24.26$
        \\ \addlinespace[0.5em]
    Image   & $(2086, 19)$ &           $8.25$ &         $107.52$ &          $>10^3$ &         $502.83$ &          $32.01$ &          $10.51$
        \\ \addlinespace[0.5em]
    R.norm  & $(7400, 21)$ &          $22.09$ &        $1148.16$ &          $>10^4$ &        $3350.87$ &          $26.76$ &          $36.73$
        \\ \addlinespace[0.5em]
    Splice  & $(2991, 61)$ &          $19.35$ &         $490.92$ &          $>10^4$ &         $943.98$ &         $122.08$ &          $37.88$
        \\ \addlinespace[0.5em]
    Thyroid &   $(215, 6)$ &           $0.70$ &           $0.66$ &         $367.51$ &           $0.35$ &           $2.71$ &           $0.61$
        \\ \addlinespace[0.5em]
    Titanic &    $(24, 4)$ &           $0.25$ &           $0.13$ &           $0.58$ &           $0.10$ &           $1.96$ &           $0.12$
        \\ \addlinespace[0.5em]
    Twonorm & $(7400, 21)$ &         $105.40$ &       $13031.38$ &          $>10^4$ &         $989.54$ &         $478.22$ &         $397.91$
        \\ \addlinespace[0.5em]
    Waveform& $(5000, 22)$ &         $437.29$ &        $9018.54$ &          $>10^4$ &          $>10^4$ &        $2243.07$ &        $1619.56$
        \\ \addlinespace[0.5em]
    \bottomrule
\end{tabular}
\end{table}
\paragraph{The worst case for LPBoost.} 
Although LPBoost outperforms the running time 
in Table~\ref{table:time}, 
it takes $m/2$ iterations for the worst case~\citep{warmuth+:nips07}. 
Even in this case, MLPBoost and ERLPBoost terminate in 2 iterations. 
\paragraph{Test errors.} 
For each dataset in the benchmark datasets, 
we first split them into train/test sets. 
Then, we perform the $5$-fold cross-validation over the training set, 
varying the capping parameter $\nu \in N$ to find the best one. 
Finally, train the algorithm using the whole training set 
with the best parameter and measure the test error with the test set. 
Table~\ref{table:cv5fold} summarizes the result. 
Since all the variants of MLPBoost solve the same problem, 
we only show MLPB.~(SS) for comparison. 
As the table shows, MLPBoosts achieve small test errors 
for most datasets. 
\begin{table}[h]
    \caption{%
        Test errors for $5$-fold cross validation %
        for the best parameters. 
    }
    \label{table:cv5fold}
    \bigskip
    \centering
    \begin{tabular}{lrrr}
    \toprule
	         &       LPB. &     ERLPB. & MLPB. (SS) \\ \midrule \addlinespace[0.5em]
	Banana   &       0.28 &       0.37 &       0.10 \\          \addlinespace[0.5em]
	B.Cancer &       0.40 &       0.49 &       0.28 \\          \addlinespace[0.5em]
	Diabetes &       0.26 &       0.26 &       0.24 \\          \addlinespace[0.5em]
	F.Solar  &       0.38 &       0.52 &       0.69 \\          \addlinespace[0.5em]
	German   &       0.28 &       0.35 &       0.27 \\          \addlinespace[0.5em]
	Heart    &       0.24 &       0.29 &       0.17 \\          \addlinespace[0.5em]
	Image    &       0.10 &       0.20 &       0.02 \\          \addlinespace[0.5em]
	Ringnorm &       0.18 &       0.18 &       0.03 \\          \addlinespace[0.5em]
	Splice   &       0.11 &       0.10 &       0.05 \\          \addlinespace[0.5em]
	Thyroid  &       0.09 &       0.05 &       0.05 \\          \addlinespace[0.5em]
	Titanic  &       0.60 &       0.60 &       0.60 \\          \addlinespace[0.5em]
	Twonorm  &       0.03 &       0.04 &       0.03 \\          \addlinespace[0.5em]
    \bottomrule
\end{tabular}
\end{table}

    \section{Conclusion}
\label{sec:conclusion}
We explored a relationship between 
the boosting algorithms for soft margin optimization 
and Frank-Wolfe algorithms via Fenchel duality. 
Using this unified view, 
we derived a scheme 
that can incorporate any secondary algorithm 
without losing the convergence guarantee. 
Even though our work is the fastest 
in the theoretically guaranteed boosting algorithms, 
LPBoost is still the fastest. 
We left the problem of inventing a faster boosting algorithm 
with a theoretical guarantee as future work.

    \bibliographystyle{unsrtnat}
    \bibliography{mitsuboshi}

    \appendix
    \section{Technical lemmas and proofs}
\label{appendix:technical_lem.}
The following lemma shows the maximum value of the relative entropy 
from the uniform distribution 
over the capped probability simplex $\psimplex{m}_{\nu}$.
\begin{lem.}
    \label{lem:relative_entropy_bound}
    $
        \max_{\bm{d} \in \psimplex{m}_{\nu}}
        \Delta(\bm{d}) \leq \ln \frac{m}{\nu}
    $.
\end{lem.}
\begin{proof}
    Since the relative entropy from the uniform distribution 
    achieves its maximal value 
    at the extreme points of $\psimplex{m}_{\nu}$, 
    a maximizer has the form 
    \begin{align}
        \nonumber
        \bm{d} = (
            \underbrace{1/\nu, 1/\nu, \dots, 1/\nu}_{k \text{ elements}},
            s,
            0, 0, \dots, 0
        ),
        \quad s = 1 - \frac{k}{\nu} \leq \frac 1 \nu
    \end{align}
    for some $k \in [m]$. 
    Plugging the maximizer into $\Delta(\bm{d})$, 
    we can write the objective function as 
    $\Delta(\bm{d}) = (k/\nu) \ln (m/\nu) + s \ln (sm)$. 
    If $s = 0$, $\Delta(\bm{d}) = \ln (m/\nu)$ holds since $k = \nu$. 
    If $s > 0$, $k / \nu < 1$ so that 
    \begin{align*}
        \Delta(\bm{d}) \leq \frac k \nu \ln \frac m \nu
            + \left(1 - \frac k \nu\right) \ln \frac m \nu
            = \ln \frac m \nu.
    \end{align*}
\end{proof}
Therefore, by setting $\eta = \frac{2}{\epsilon} \ln \frac{m}{\nu}$, 
the entropy term does not exceed $\frac{\epsilon}{2}$. 

The following lemma shows the dual problem of the edge minimization. 
\begin{lem.}
    \label{lem:dual_problem}
    Let $f : \mathbb R^m \to \{0, +\infty\}$, 
    $g : \mathbb R^n \to \mathbb R$ be 
    functions defined as 
    \begin{align}
        \nonumber
        f(\bm{d}) =
        \begin{cases}
                  0 & \bm{d} \in    \psimplex{m}_{\nu} \\
            +\infty & \bm{d} \notin \psimplex{m}_{\nu}
        \end{cases},
        \qquad \qquad \qquad
        g(\bm{\theta}) = \max_{j \in [n]} \theta_j.
    \end{align}
    Then, the dual problem of edge minimization 
    \begin{align}
        \label{eq:appendix_edge_minimization}
        \min_{\bm{d}} f(\bm{d}) + g(A^\top \bm{d})
    \end{align}
    is the soft margin maximization 
    \begin{align}
        \label{eq:appendix_smm}
        \max_{\bm{w} \in \psimplex{n}} - f^\star( -A \bm{w} ).
    \end{align}
    Further, the strong duality holds. 
\end{lem.}
\begin{proof}
    We can use Theorem~\ref{thm:strong_duality} to derive the dual problem. 
    Since 
    \begin{align*}
        \bm{g}^\star(\bm{w}) = 
        \begin{cases}
                  0 & \bm{w} \in    \psimplex{n} \\
            +\infty & \bm{w} \notin \psimplex{n}
        \end{cases},
    \end{align*}
    one can verify the dual form is given as~(\ref{eq:appendix_smm}). 
    To prove the strong duality, 
    it is enough to prove 
    $
        \bm{0} \in \interior\left(
            \domain g - A^\top \domain f
        \right)
    $. 
    By definition, $\domain g = \mathbb R^n$ and 
    $\domain f = \psimplex{m}_{\nu}$ and hence 
    \begin{align*}
        \domain g - A^\top \domain f
        = \left\{
            \bm{w} - A^\top \bm{d} \mid
            \bm{w} \in \mathbb{R}^n, \bm{d} \in \psimplex{m}_{\nu}
        \right\}.
    \end{align*}
    Obviously, $\bm{0} \in \interior (\domain g - A^\top \domain f)$ 
    and thus the strong duality holds. 
\end{proof}
Since 
\begin{align*}
    f^\star(-A\bm{w}) 
    = \sup_{\bm{d}} \left[ - \bm{d}^\top A \bm{w} - f(\bm{d}) \right]
    = \max_{\bm{d} \in \psimplex{m}_{\nu}} - \bm{d}^\top A \bm{w}
    = \min_{\bm{d} \in \psimplex{m}_{\nu}} \bm{d}^\top A \bm{w},
\end{align*}
we can write the dual problem~(\ref{eq:appendix_smm}) explicitly:
\begin{align*}
    \max_{\bm{w} \in \psimplex{n}} - f^\star ( -A \bm{w} )
        = - \min_{\bm{w} \in \psimplex{n}}
            \max_{\bm{d} \in \psimplex{m}_{\nu}}
        \bm{d}^\top A \bm{w}.
\end{align*}
We get the dual problem for the regularized edge minimization problem 
by a similar derivation. 
\begin{cor.}
    \label{lem:smoothed_dual_problem}
    Let $f, g$ be the functions defined in Lemma~\ref{lem:dual_problem} and 
    let $\Delta(\bm{d}) = \sum_{i=1}^m d_i \ln d_i + \ln(m)$ be 
    the relative entropy function from the uniform distribution. 
    Define $\tilde{f} = f + (1/\eta) \Delta$ for some $\eta > 0$. 
    Then, the dual problem of 
    \begin{align}
        \nonumber
        \min_{\bm{d}} \tilde{f}(\bm{d}) + g(A^\top \bm{d})
        \qquad \text{ is } \qquad
        \max_{\bm{w} \in \psimplex{n}} - \tilde{f}^\star(-A \bm{w}).
    \end{align}
\end{cor.}

\subsection{Proof of Lemma~\ref{lem:fenchel_conjugate_functions}}
\begin{proof}
    By the definition of Fenchel conjugate, 
    \begin{align*}
        \tilde{f}^\star(\bm{\theta})
        & = \sup_{\bm{d}} \left\{
            \bm{d}^\top \bm{\theta} - \tilde{f}(\bm{d})
        \right\}
        \geq \sup_{\bm{d}} \left\{
            \bm{d}^\top \bm{\theta} - f(\bm{d}) - c
        \right\}
        = f^\star(\bm{\theta}) - c, \\
        \tilde{f}^\star(\bm{\theta})
        & = \sup_{\bm{d}} \left\{
            \bm{d}^\top \bm{\theta} - \tilde{f}(\bm{d})
        \right\}
        \leq \sup_{\bm{d}} \left\{
            \bm{d}^\top \bm{\theta} - f(\bm{d})
        \right\}
        = f^\star(\bm{\theta}).
    \end{align*}
\end{proof}
By Lemma~\ref{lem:relative_entropy_bound} 
and~\ref{lem:fenchel_conjugate_functions}, 
we get 
$
    \tilde{f}^\star (-A\bm{w}) - \frac{\epsilon}{2}
    \leq f(-A\bm{w}) \leq \tilde{f}^\star(-A\bm{w})
$ for all $-A \bm{w}$ 
if $\eta \geq \frac{2}{\epsilon} \ln \frac{m}{\nu}$.

\subsection{
    Proof of Theorem~\ref{thm:mlpboost_convergence_guarantee}
    for the short-step FW rule~\ref{alg:ss_rule}
}
Recall that in the proof of Theorem~\ref{thm:mlpboost_convergence_guarantee}, 
we showed the inequality 
$\epsilon_t - \epsilon_{t+1} \geq \frac{1}{8 \eta} \epsilon_t^2$ 
for the short-step case. 
We prove $\epsilon_t \leq \frac{8\eta}{t + 2}$ by induction on $t$. 
For the base case, $t = 1$, by Lemma~\ref{lem:fenchel_conjugate_functions}, 
\begin{align*}
    \epsilon_1
    = \min_{\tau \in \{0, 1\}} (\bm{d}_\tau^\top A)_{j_{\tau + 1}}
        + \tilde{f}^\star (- A\bm{w}_1)
    \leq 1 + f^\star (- A\bm{w}_1)
    \leq 2
    \leq \frac{8\eta}{1 + 2}.
\end{align*}
For the inductive case, assume that $\epsilon_t \leq \frac{8\eta}{t+2}$ 
for $t \geq 1$. 
By the inequality
$\epsilon_t - \epsilon_{t+1} \geq \frac{1}{8 \eta} \epsilon_t^2$, 
we have
\begin{align}
    \label{eq:appendix_shortstep_01}
    \epsilon_{t+1} 
    \leq \left(1 - \frac{1}{8\eta} \epsilon_t\right) \epsilon_t.
\end{align}
By simple calculation, 
one can see that the maximizer $\epsilon$ of the RHS over $\mathbb{R}$ is 
$\epsilon = 4\eta$. 
By the inductive assumption, $\epsilon_t \leq \frac{8 \eta}{t + 2}$. 
Since $\frac{8\eta}{t+2}$ is 
the maximizer of~(\ref{eq:appendix_shortstep_01}) 
over $[0, \frac{8\eta}{t+2}]$, 
we can plug this value into~(\ref{eq:appendix_shortstep_01}). 
\begin{align*}
    \epsilon_{t+1} 
    \leq \left(1 - \frac{1}{8\eta} \frac{8\eta}{t+2}\right)
    \frac{8\eta}{t+2}
    = \frac{t+1}{t+2} \frac{8\eta}{t+2}
    \leq \frac{8\eta}{t+3}
\end{align*}
Therefore, $\epsilon_{t} \leq \frac{8\eta}{t+2}$ holds for all $t \geq 1$. 
Thus, we obtain the desired result. 

    \section{Additional experiments}
\label{sec:appendix_experiemnt}
\begin{figure}[p]
    \centering
    \begin{tabular}{ccc}
        \begin{minipage}[t]{0.31\hsize}
            \centering
            \includegraphics[keepaspectratio, scale=0.30]
            {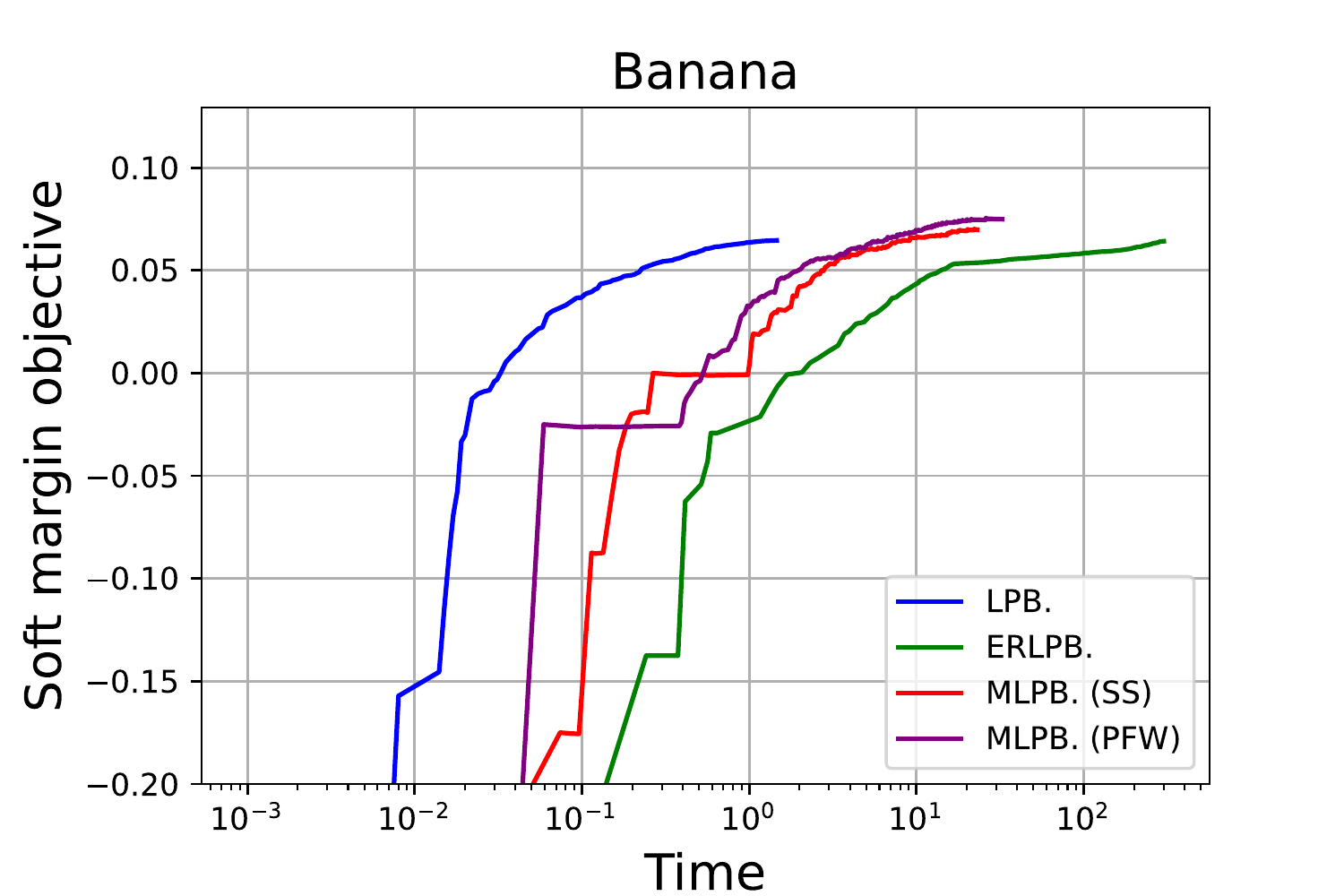}
        \end{minipage}
        &
        \begin{minipage}[t]{0.31\hsize}
            \centering
            \includegraphics[keepaspectratio, scale=0.30]
            {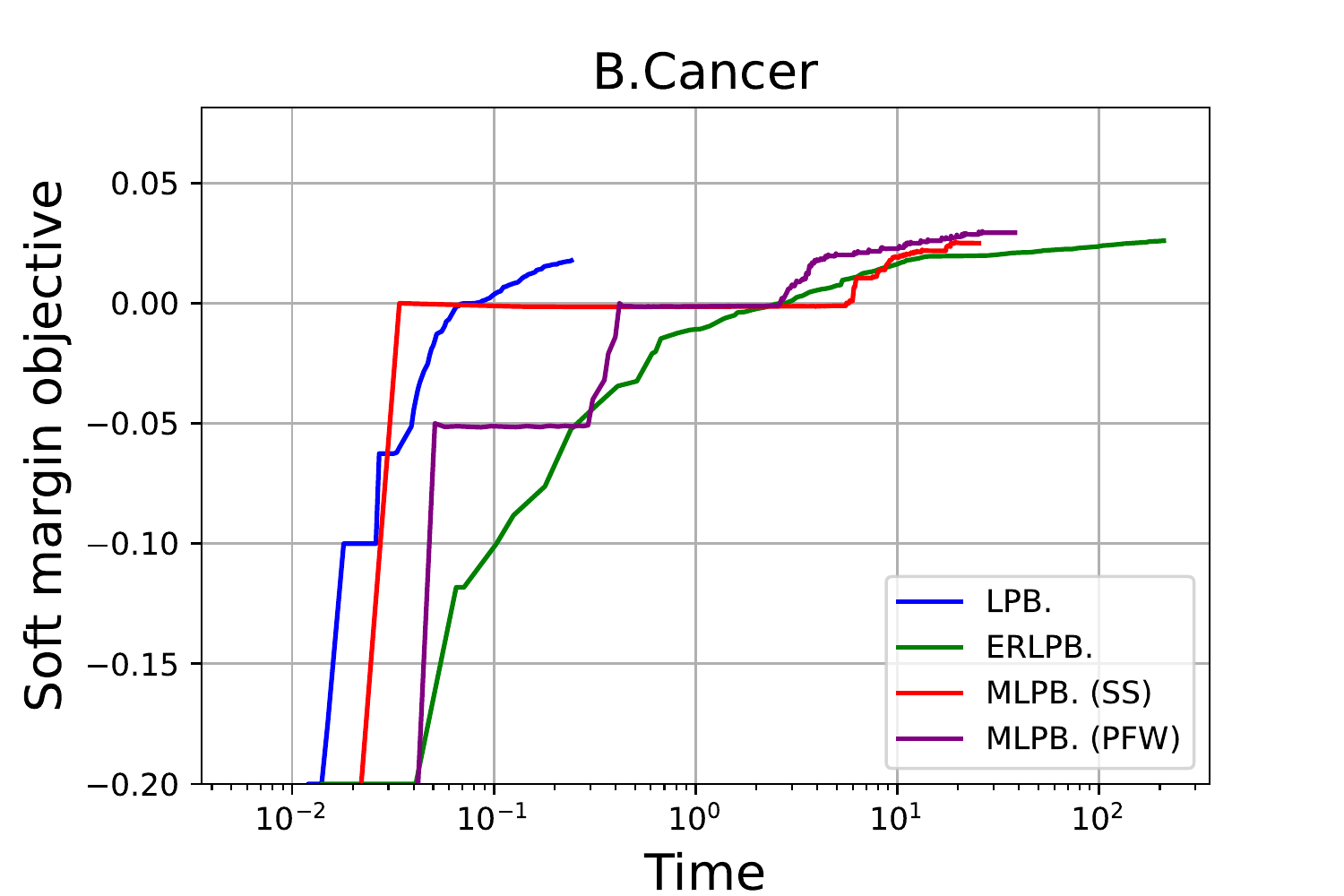}
        \end{minipage}
        &
        \begin{minipage}[t]{0.31\hsize}
            \centering
            \includegraphics[keepaspectratio, scale=0.30]
            {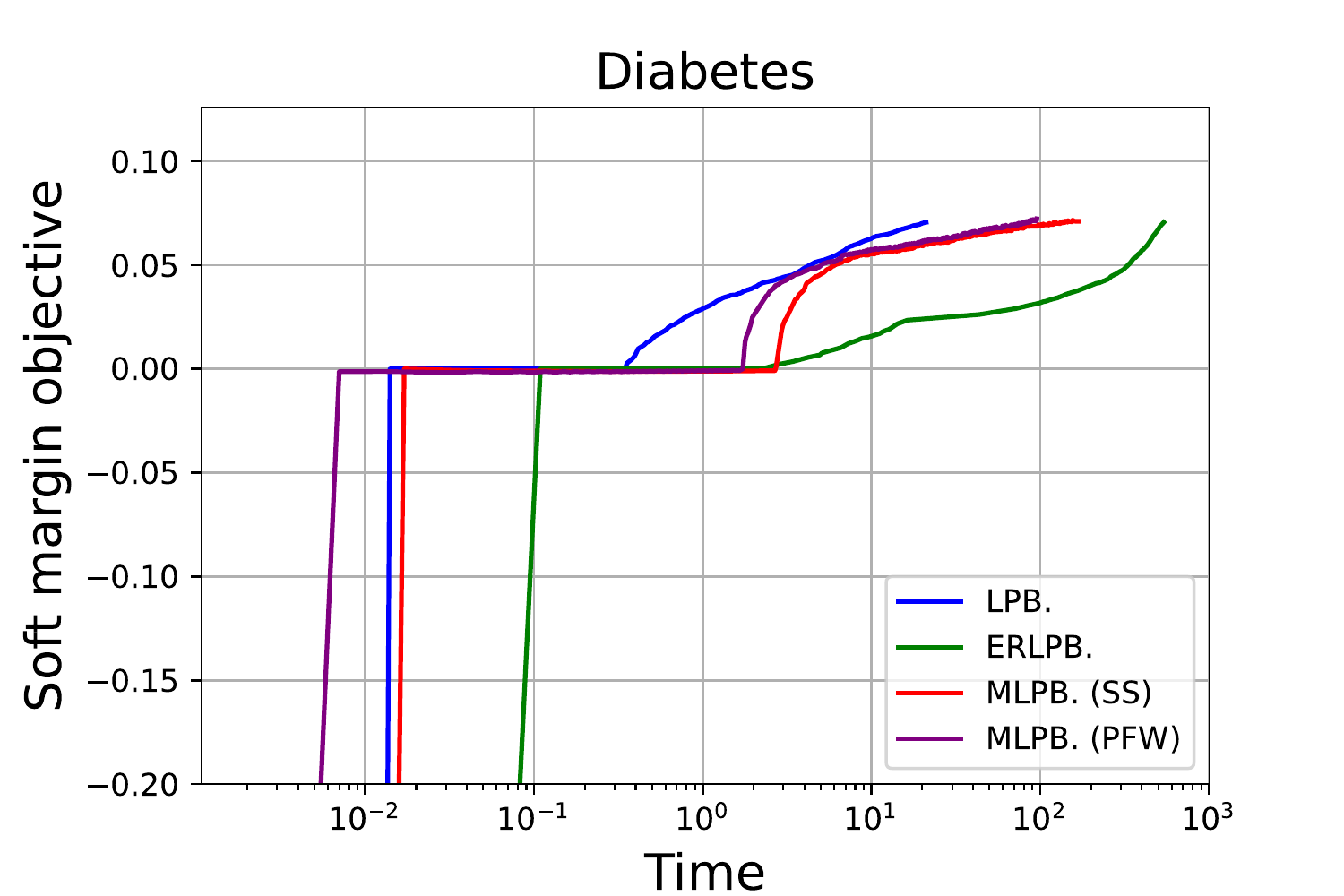}
        \end{minipage}
        \\
        \begin{minipage}[t]{0.31\hsize}
            \centering
            \includegraphics[keepaspectratio, scale=0.30]
            {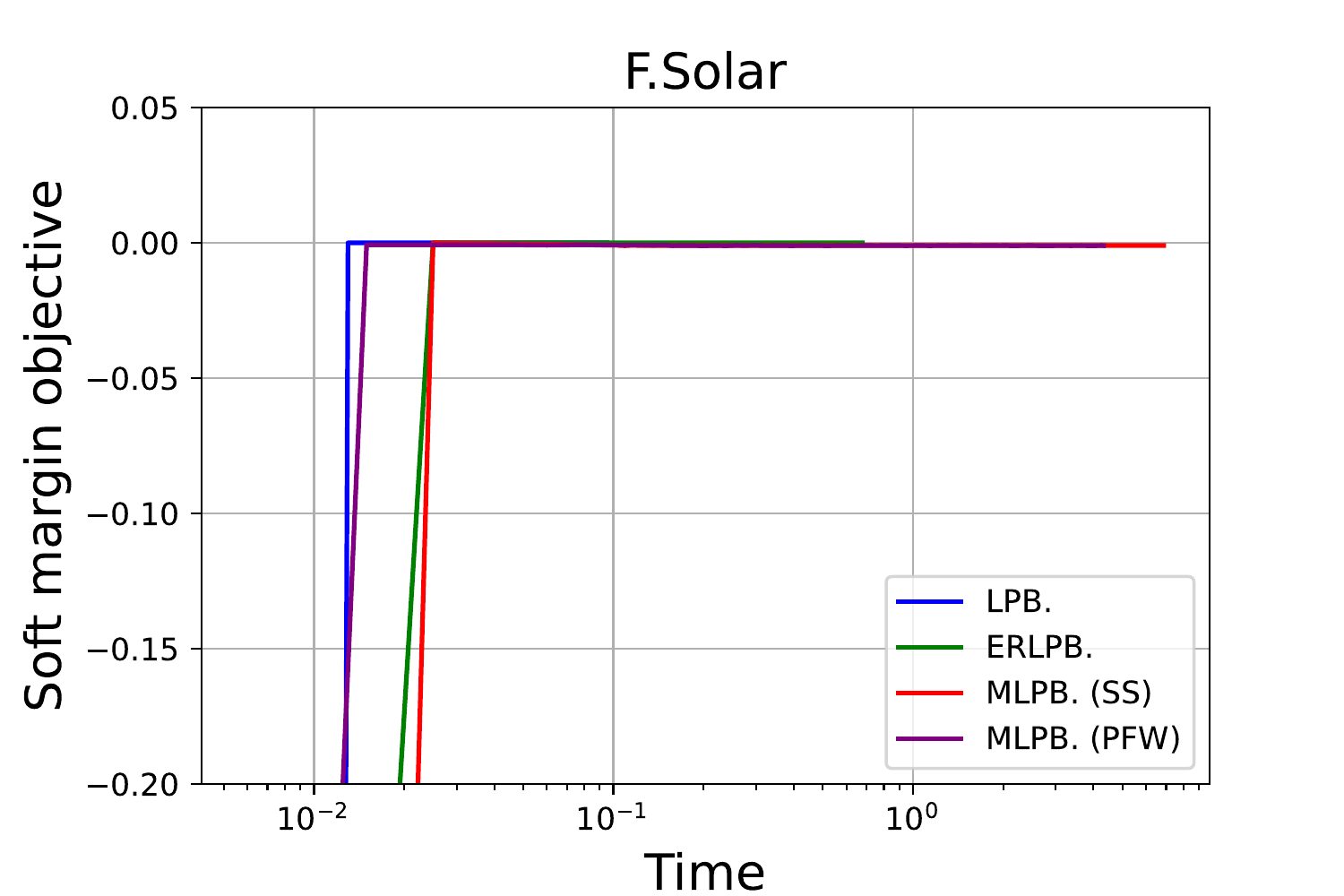}
        \end{minipage}
        &
        \begin{minipage}[t]{0.31\hsize}
            \centering
            \includegraphics[keepaspectratio, scale=0.30]
            {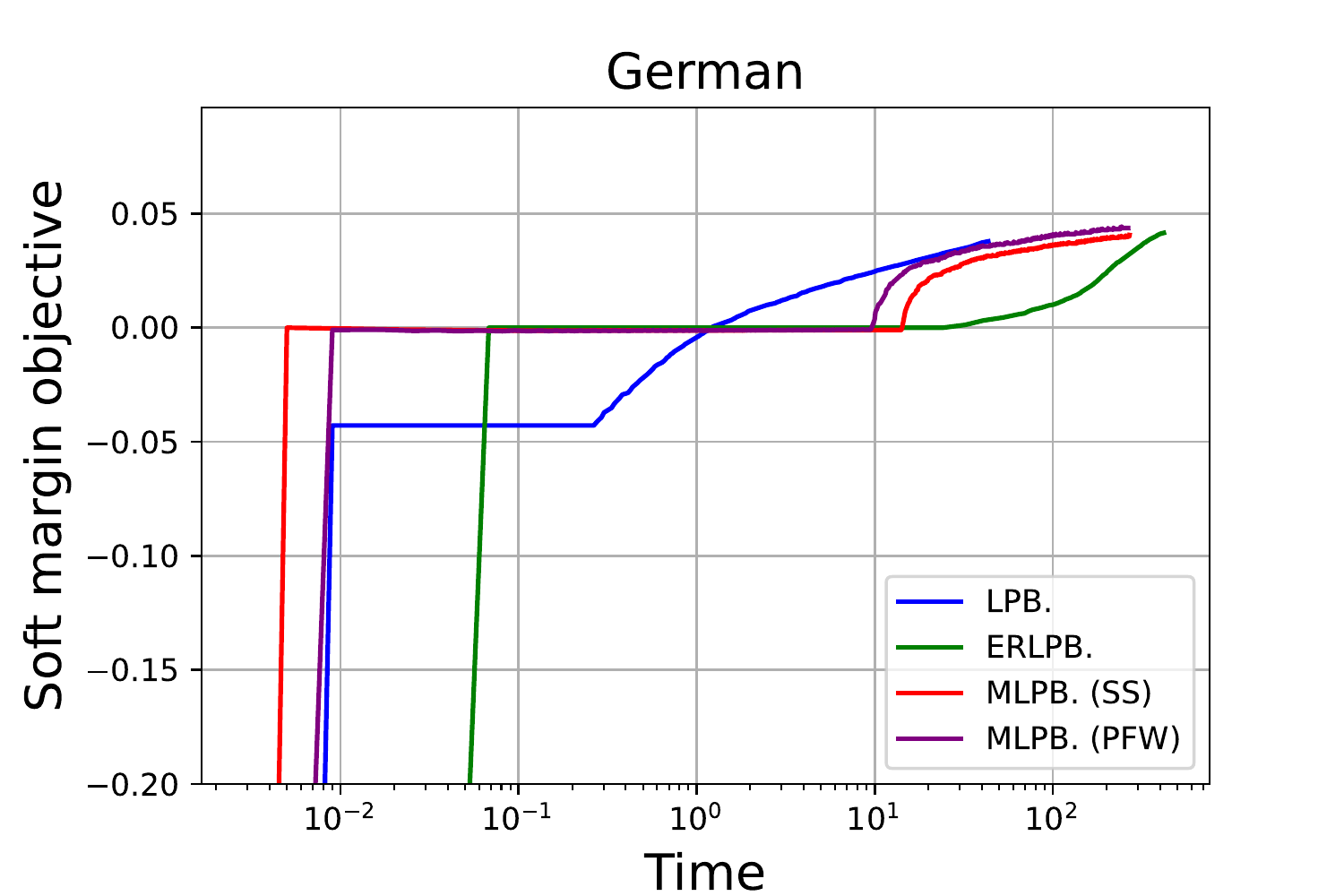}
        \end{minipage}
        &
        \begin{minipage}[t]{0.31\hsize}
            \centering
            \includegraphics[keepaspectratio, scale=0.30]
            {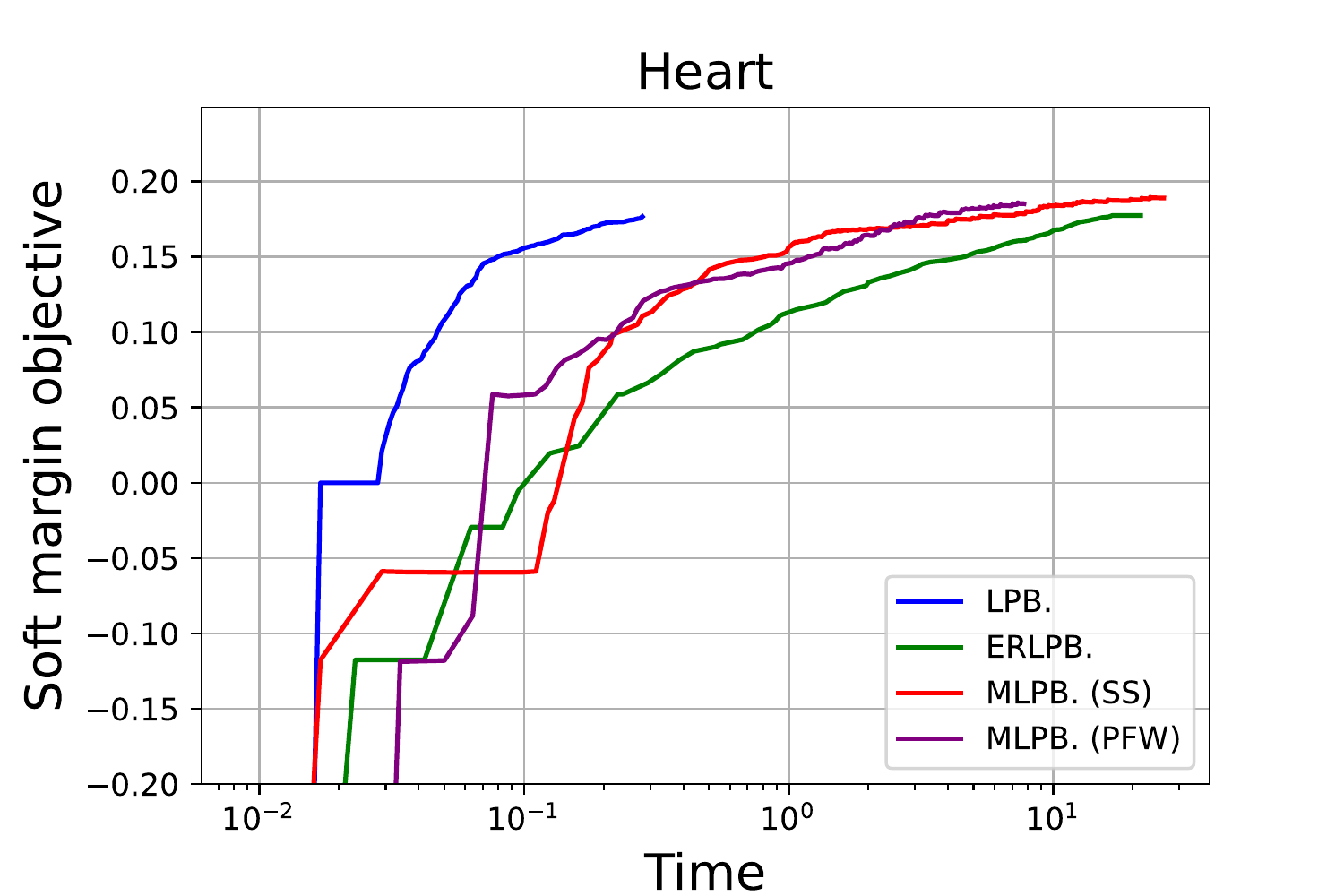}
        \end{minipage}
        \\
        \begin{minipage}[t]{0.31\hsize}
            \centering
            \includegraphics[keepaspectratio, scale=0.30]
            {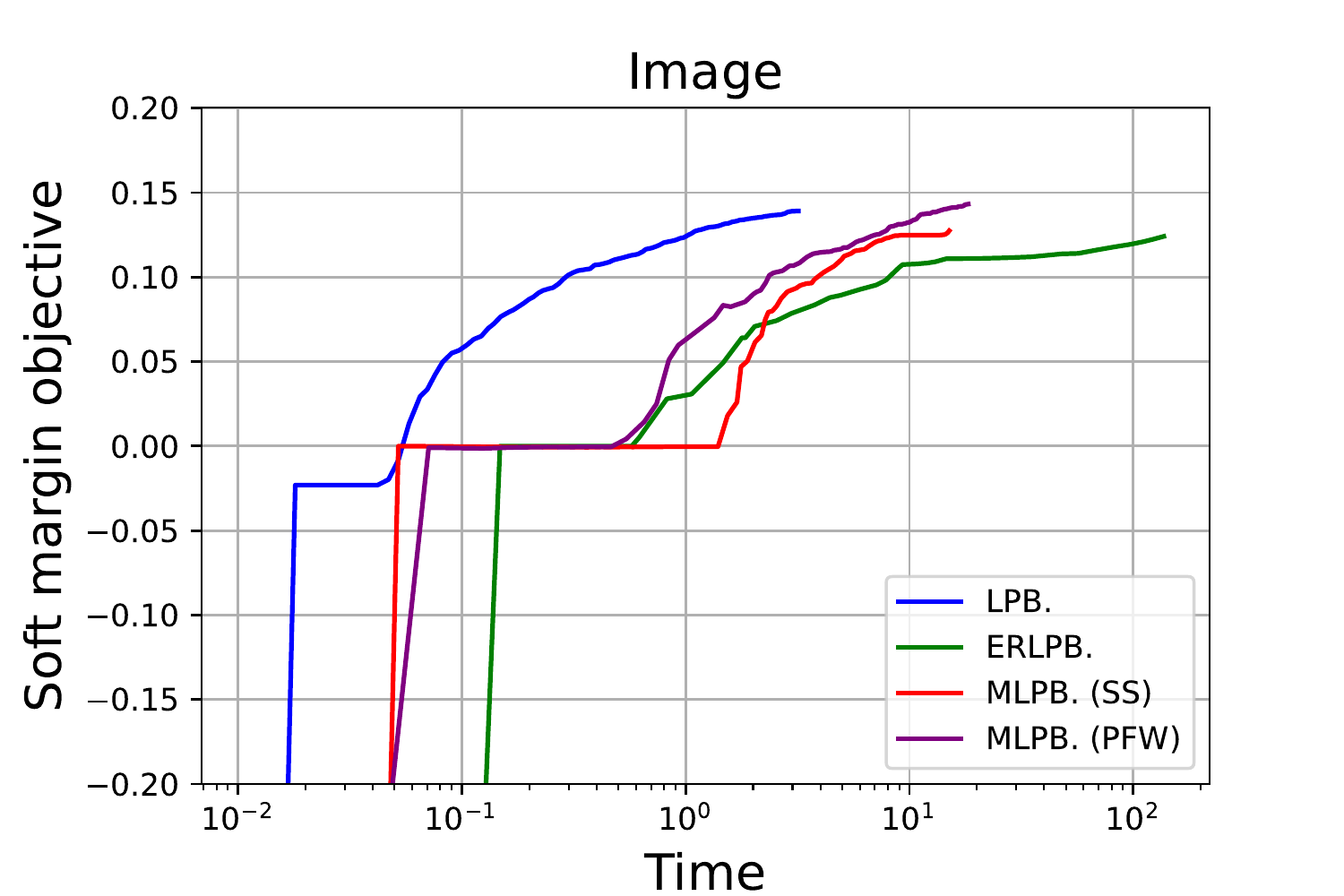}
        \end{minipage}
        &
        \begin{minipage}[t]{0.31\hsize}
            \centering
            \includegraphics[keepaspectratio, scale=0.30]
            {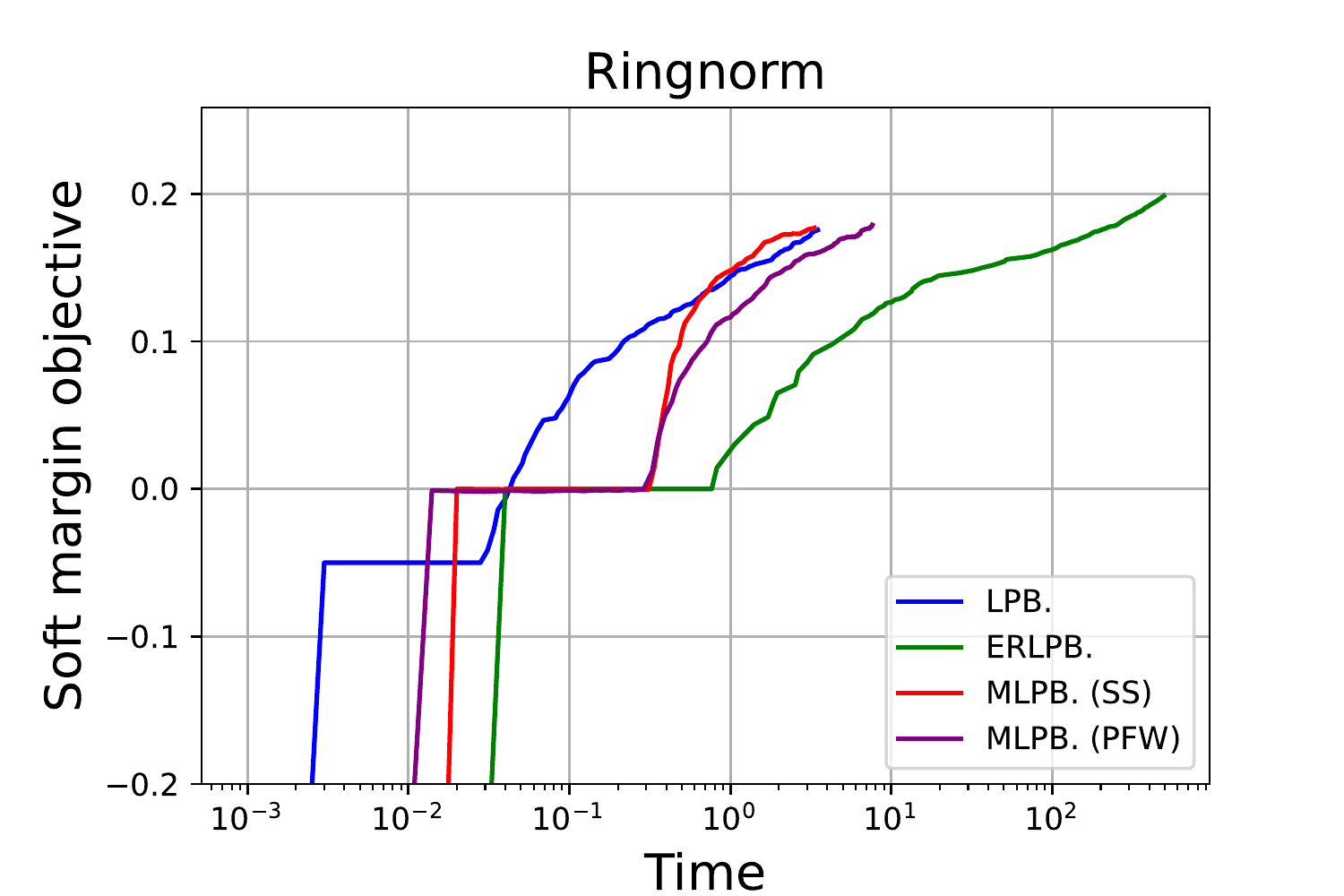}
        \end{minipage}
        &
        \begin{minipage}[t]{0.31\hsize}
            \centering
            \includegraphics[keepaspectratio, scale=0.30]
            {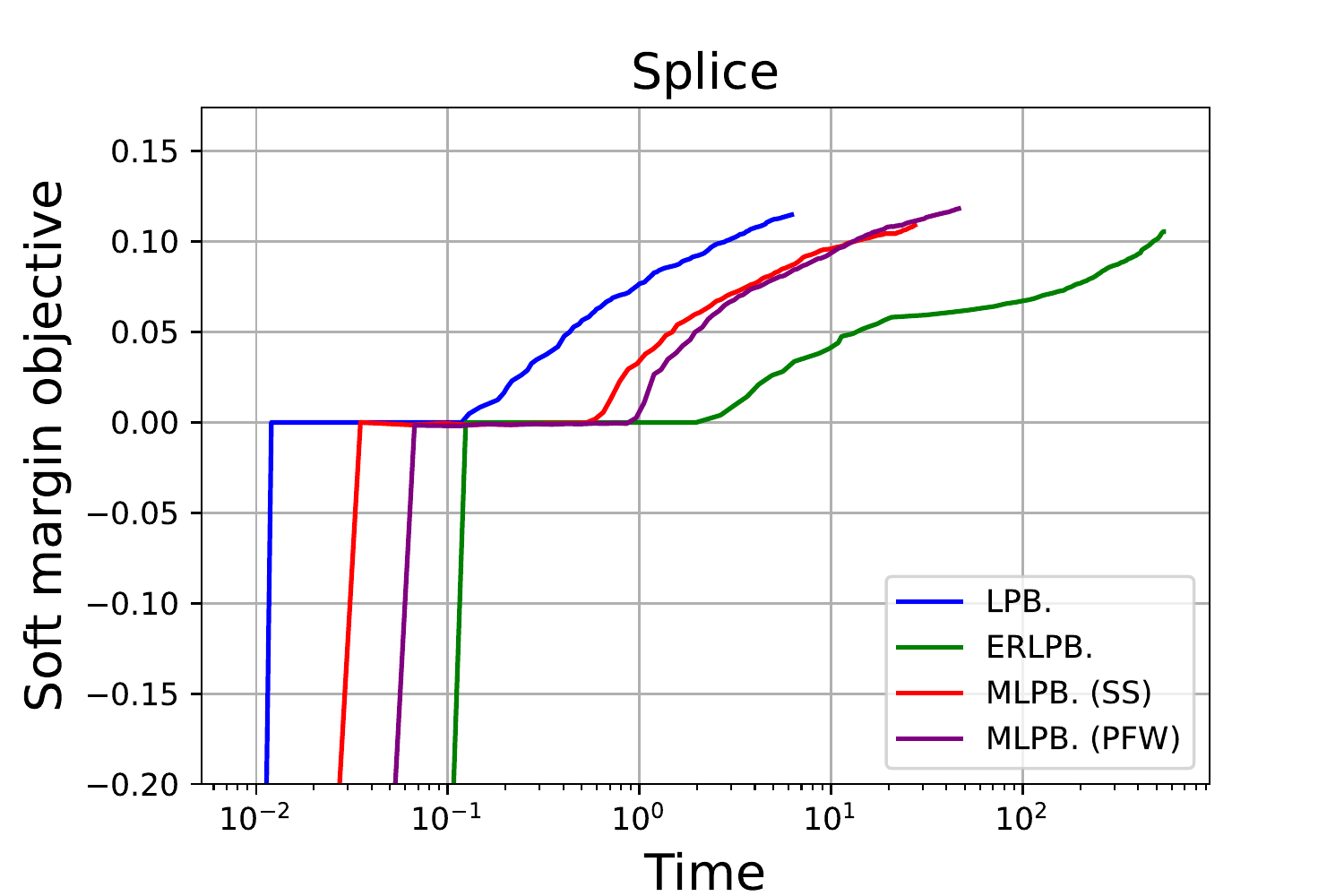}
        \end{minipage}
        \\
        \begin{minipage}[t]{0.31\hsize}
            \centering
            \includegraphics[keepaspectratio, scale=0.30]
            {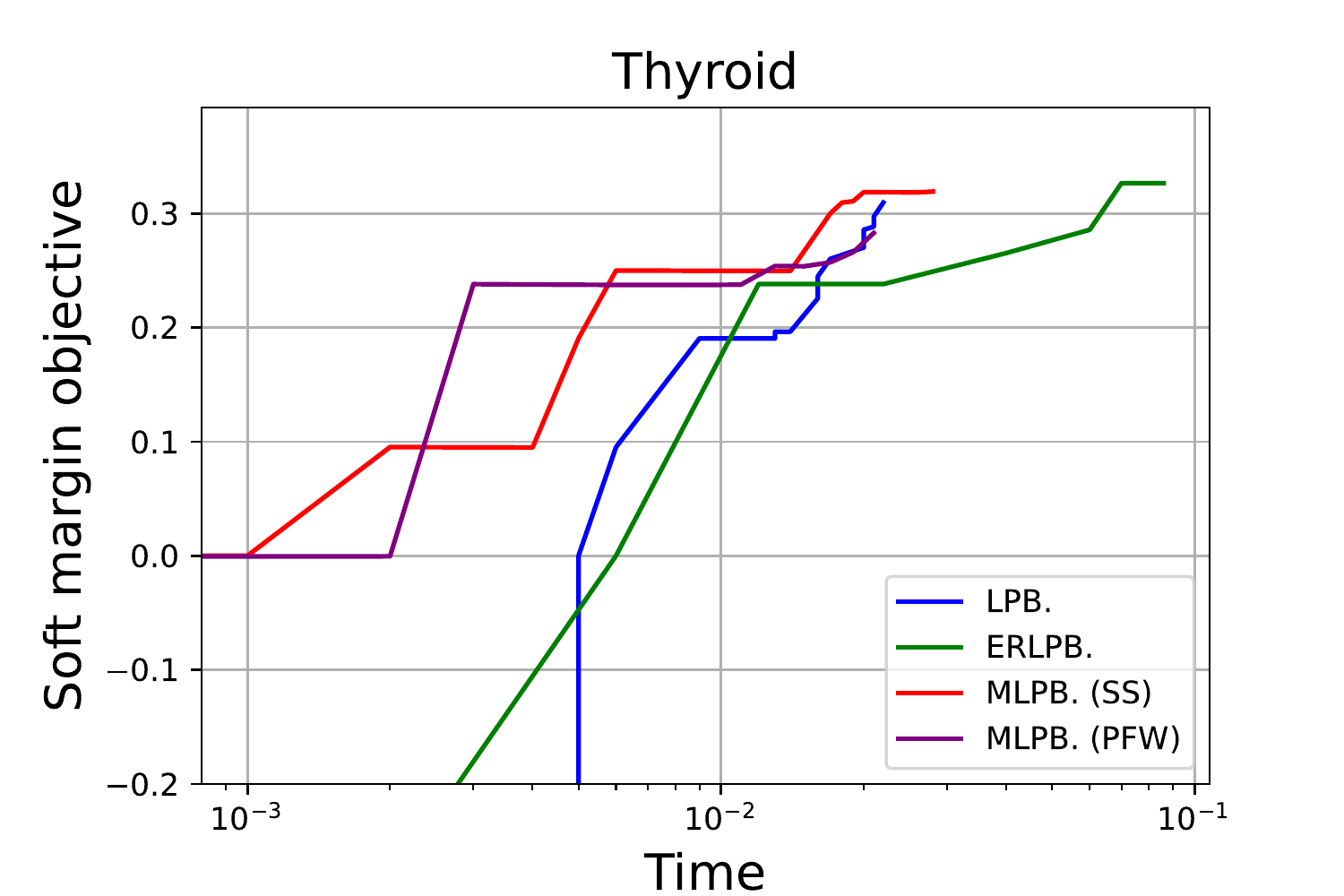}
        \end{minipage}
        &
        \begin{minipage}[t]{0.31\hsize}
            \centering
            \includegraphics[keepaspectratio, scale=0.30]
            {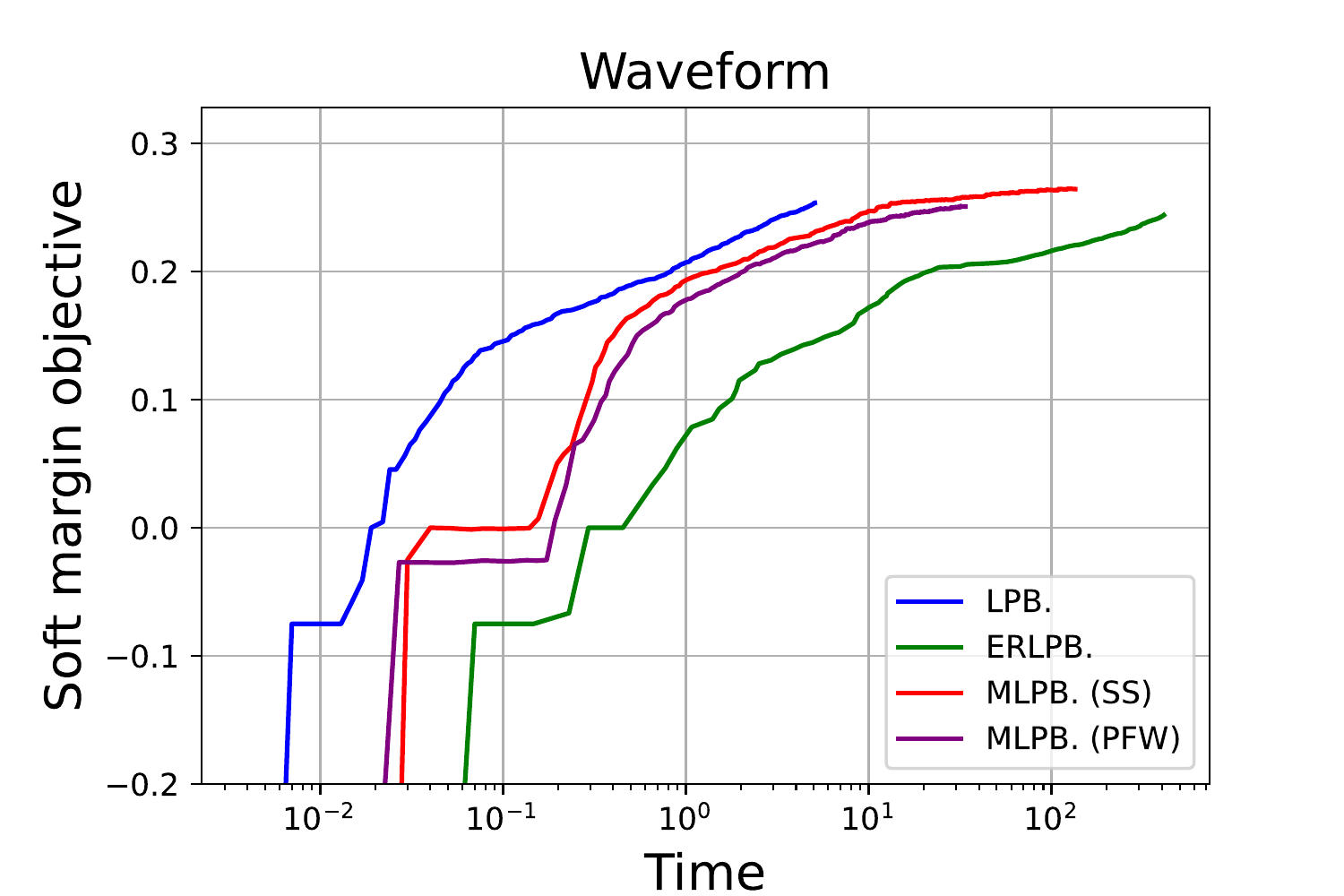}
        \end{minipage}
        &
        \begin{minipage}[t]{0.31\hsize}
            \centering
            \includegraphics[keepaspectratio, scale=0.30]
            {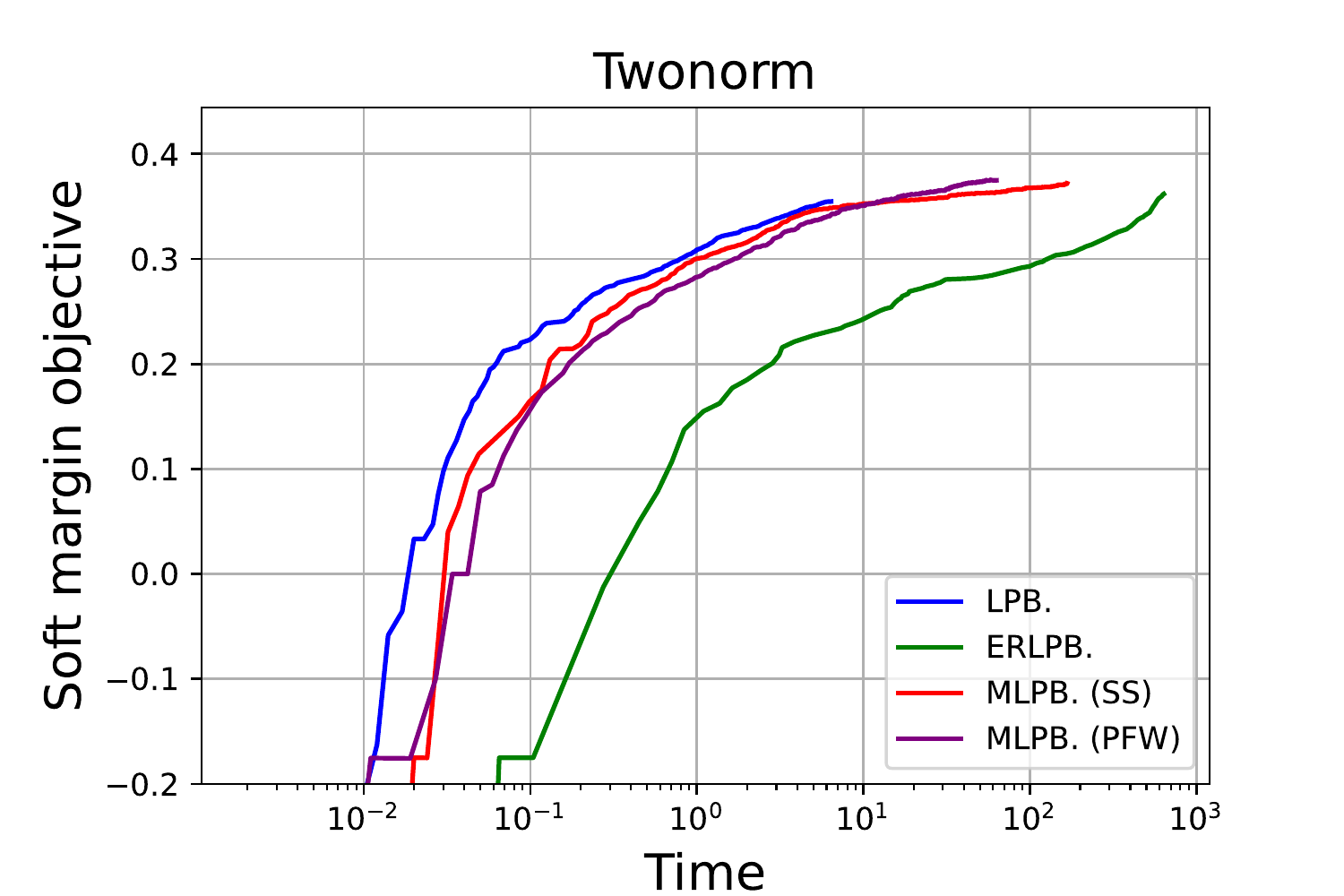}
        \end{minipage}
    \end{tabular}
    \caption{%
        Time vs. soft margin objective %
        with parameters $\nu = 0.1m$ and $\epsilon = 0.01$. %
        Note that the time axis is log-scale. %
        For many datasets, MLPBoosts tend to achieve %
        a large margin rapidly. %
    }
    \label{fig:appendix_margin_objectives}
\end{figure}

This section includes experiments, not in the main paper. 
We first show the comparison of boosting algorithms, 
LPBoost, ERLPBoost, C-ERLPBoost, and our scheme. 
Since C-ERLPBoost is an instance of the short-step FW algorithm, 
we call it FW. 
We call our scheme with secondary algorithm~\ref{alg:lpb_subroutine} 
as MLPBoost. 
Figure~\ref{fig:appendix_margin_objectives} shows the convergence curve. 
MLPB.~(SS) is MLPBoost with FW algorithm~\ref{alg:ss_rule}, 
and MLPB.~(PFW) is MLPBoost 
with Pairwise FW algorithm~\ref{alg:pairwise_rule}. 
As expected, our algorithm converges faster than ERLPBoost 
and is competitive with LPBoost. 

\begin{figure}[p]
    \centering
    \begin{tabular}{ccc}
        \begin{minipage}[t]{0.31\hsize}
            \centering
            \includegraphics[keepaspectratio, scale=0.30]
            {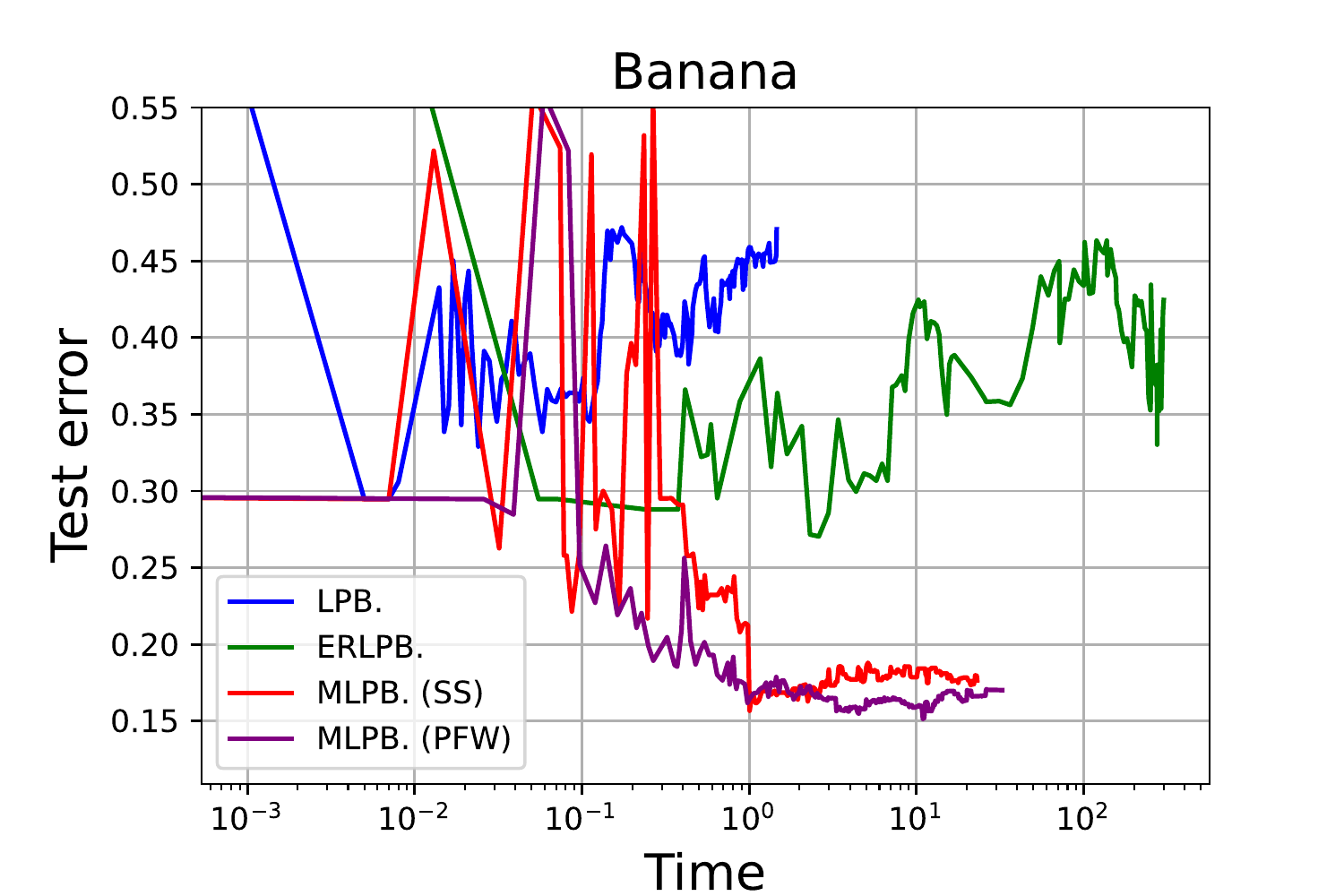}
        \end{minipage}
        &
        \begin{minipage}[t]{0.31\hsize}
            \centering
            \includegraphics[keepaspectratio, scale=0.30]
            {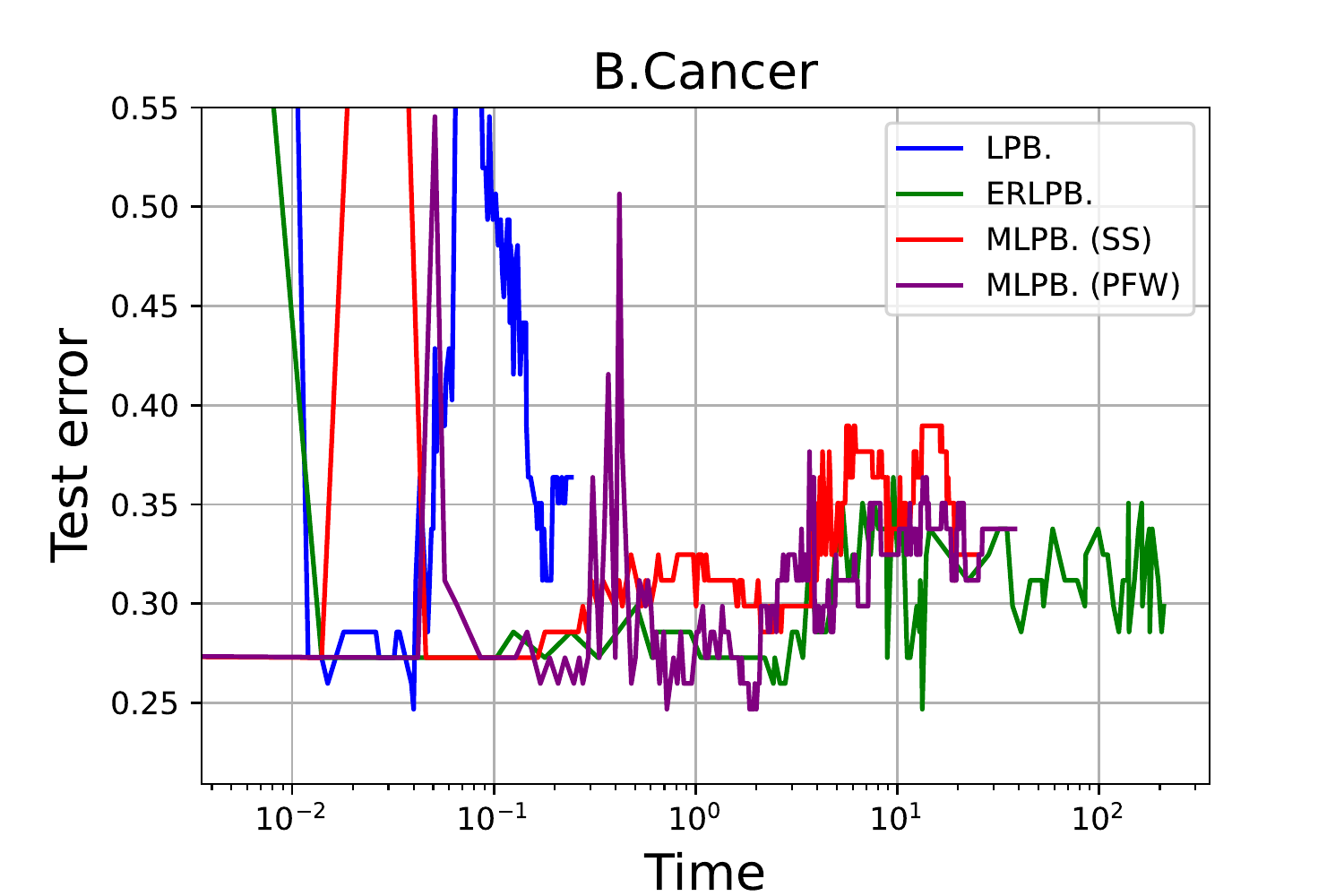}
        \end{minipage}
        &
        \begin{minipage}[t]{0.31\hsize}
            \centering
            \includegraphics[keepaspectratio, scale=0.30]
            {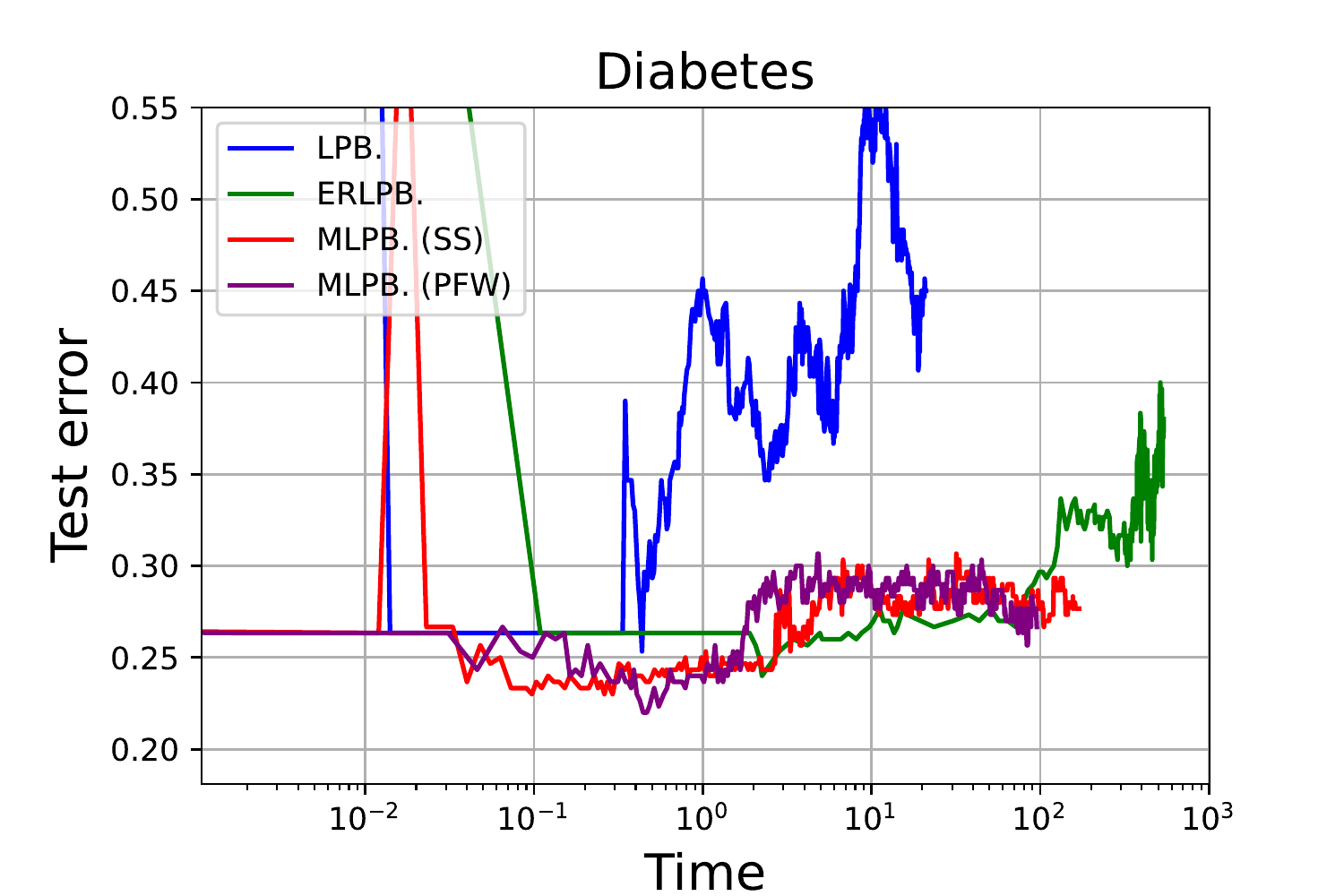}
        \end{minipage}
        \\
        \begin{minipage}[t]{0.31\hsize}
            \centering
            \includegraphics[keepaspectratio, scale=0.30]
            {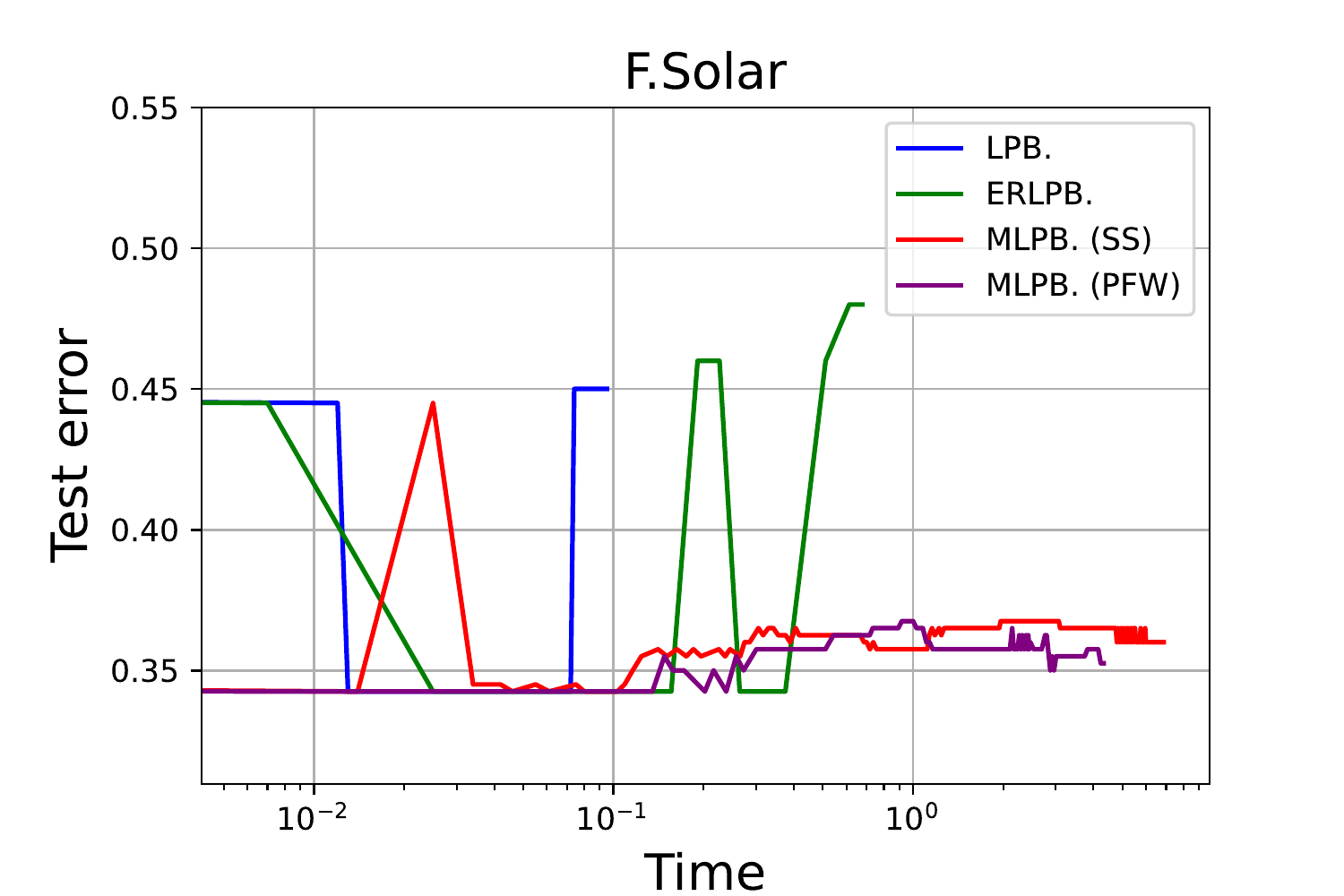}
        \end{minipage}
        &
        \begin{minipage}[t]{0.31\hsize}
            \centering
            \includegraphics[keepaspectratio, scale=0.30]
            {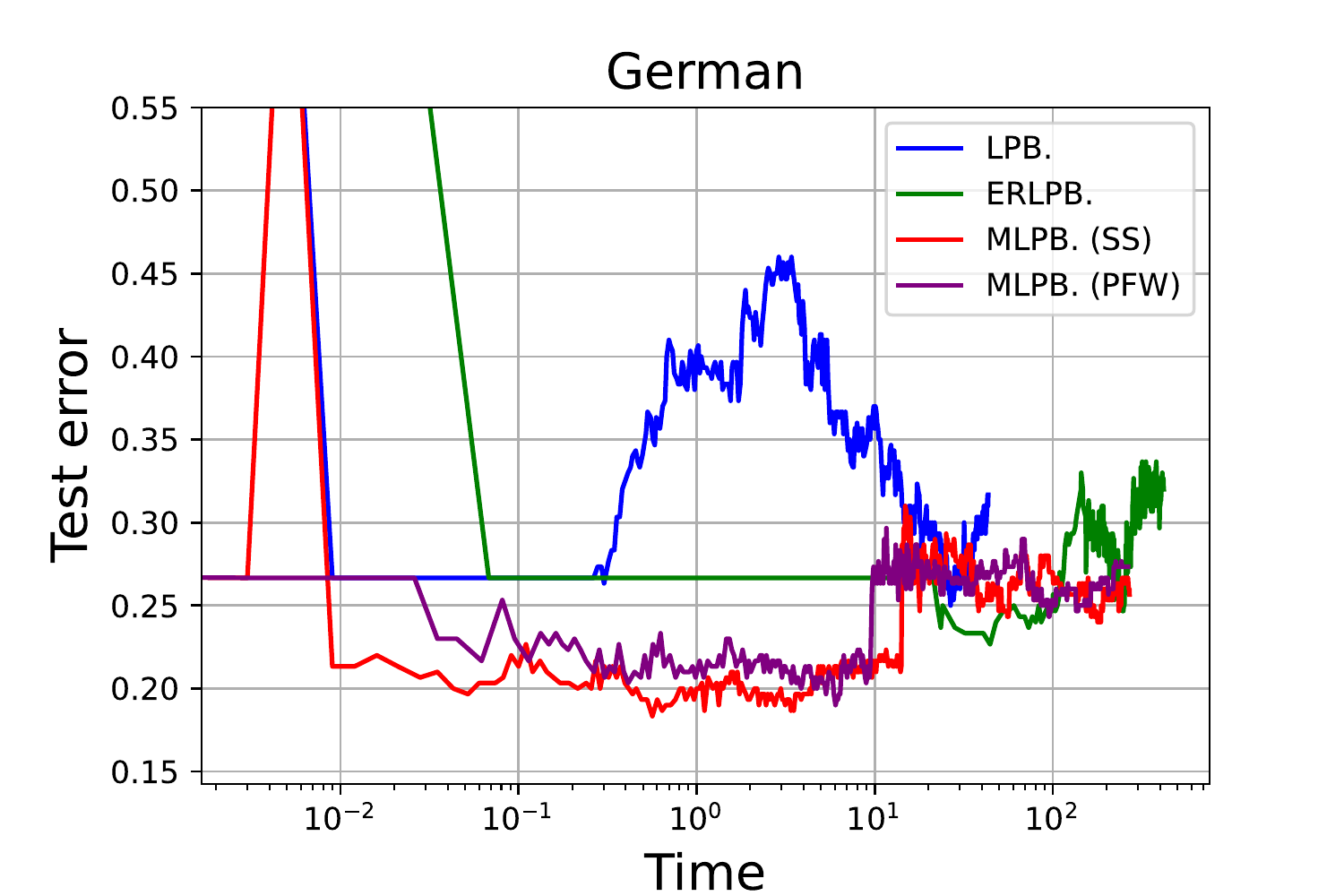}
        \end{minipage}
        &
        \begin{minipage}[t]{0.31\hsize}
            \centering
            \includegraphics[keepaspectratio, scale=0.30]
            {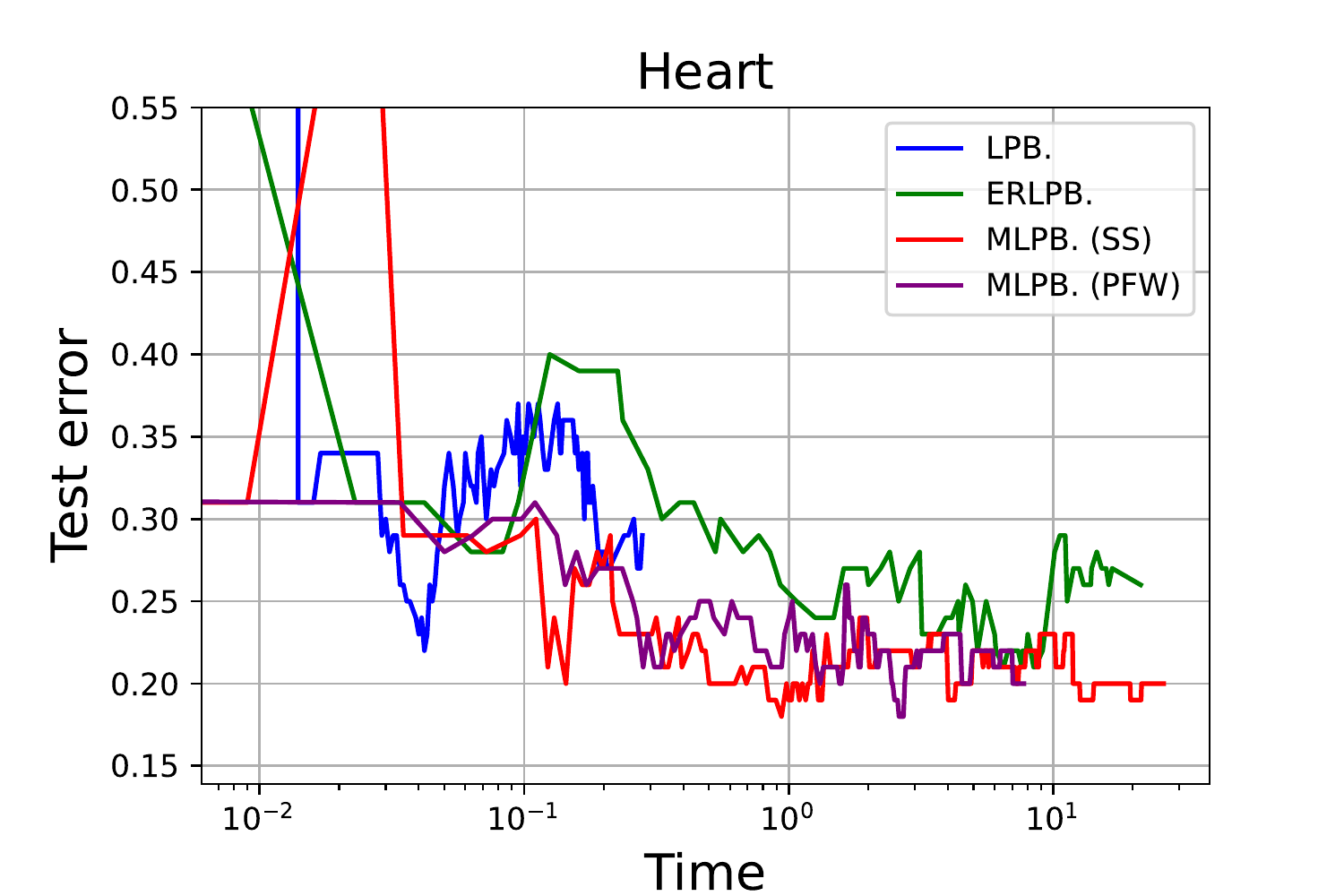}
        \end{minipage}
        \\
        \begin{minipage}[t]{0.31\hsize}
            \centering
            \includegraphics[keepaspectratio, scale=0.30]
            {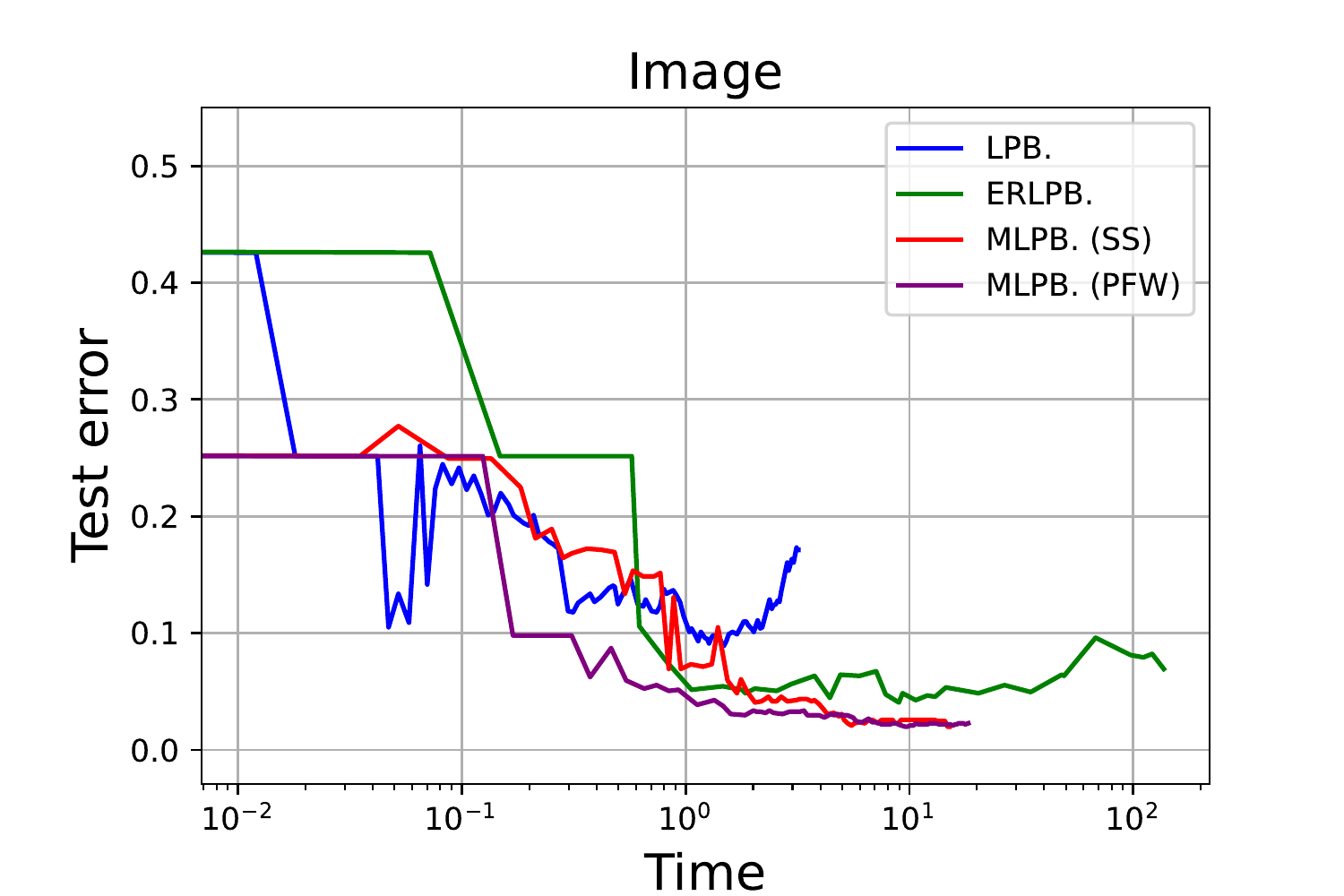}
        \end{minipage}
        &
        \begin{minipage}[t]{0.31\hsize}
            \centering
            \includegraphics[keepaspectratio, scale=0.30]
            {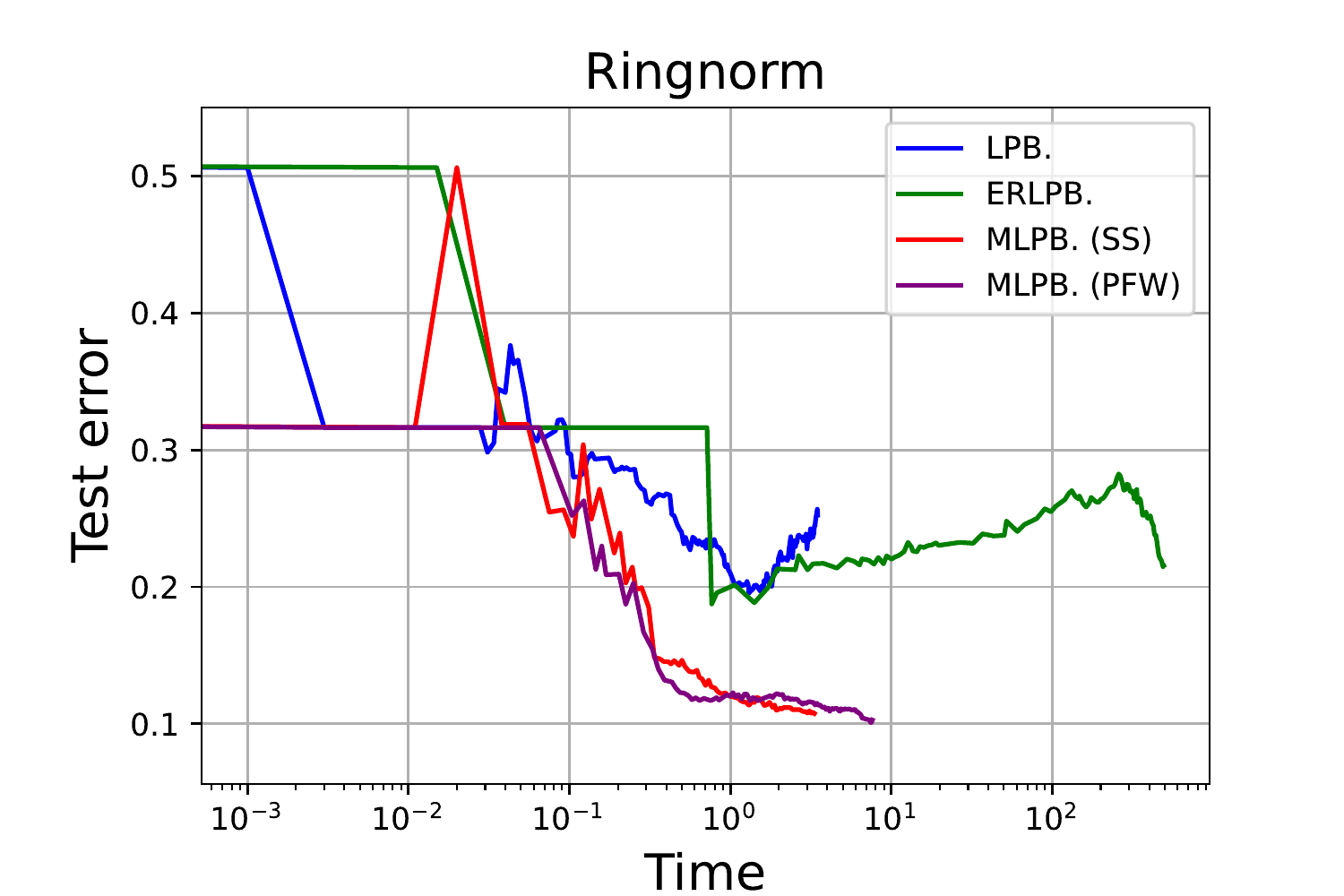}
        \end{minipage}
        &
        \begin{minipage}[t]{0.31\hsize}
            \centering
            \includegraphics[keepaspectratio, scale=0.30]
            {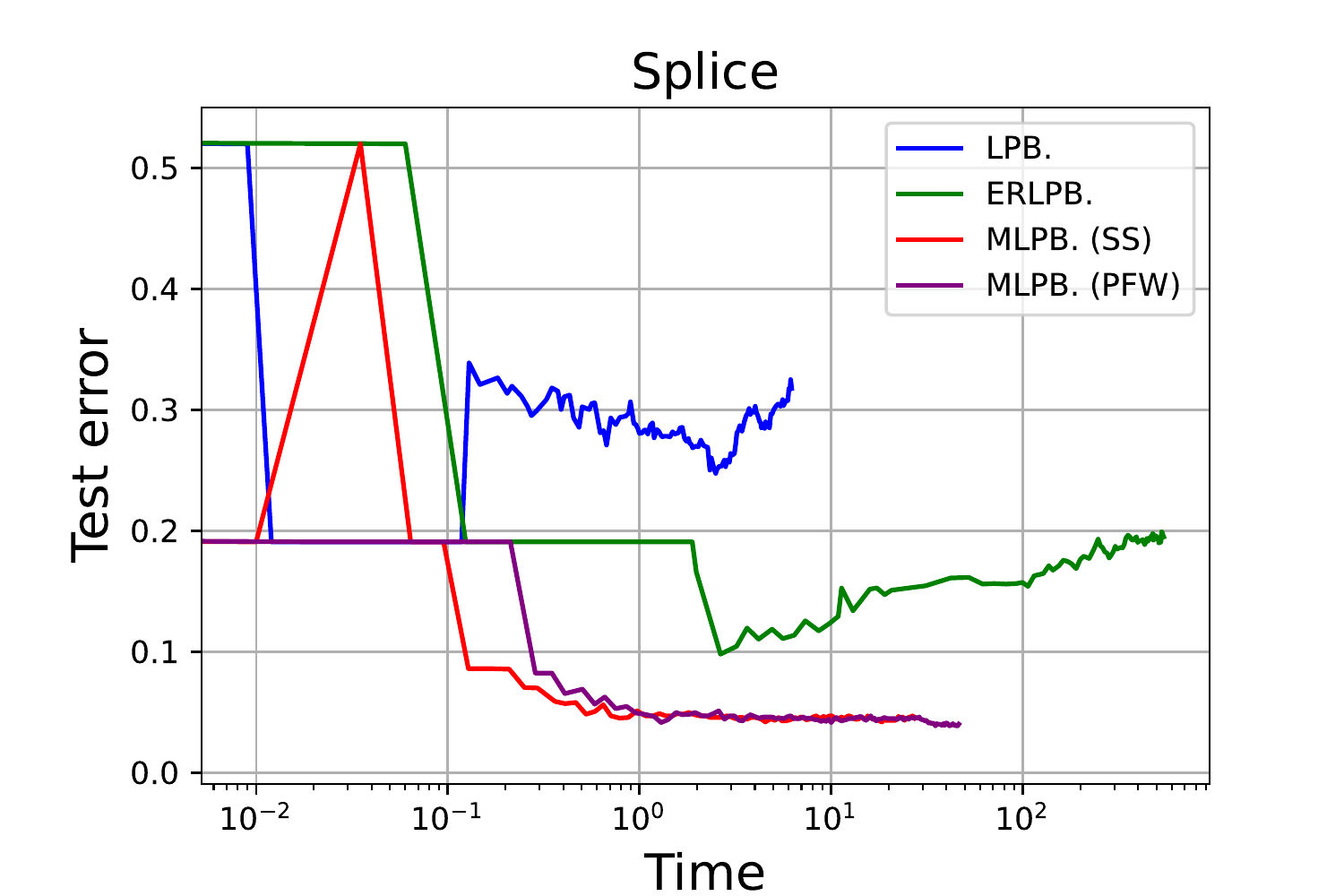}
        \end{minipage}
        \\
        \begin{minipage}[t]{0.31\hsize}
            \centering
            \includegraphics[keepaspectratio, scale=0.30]
            {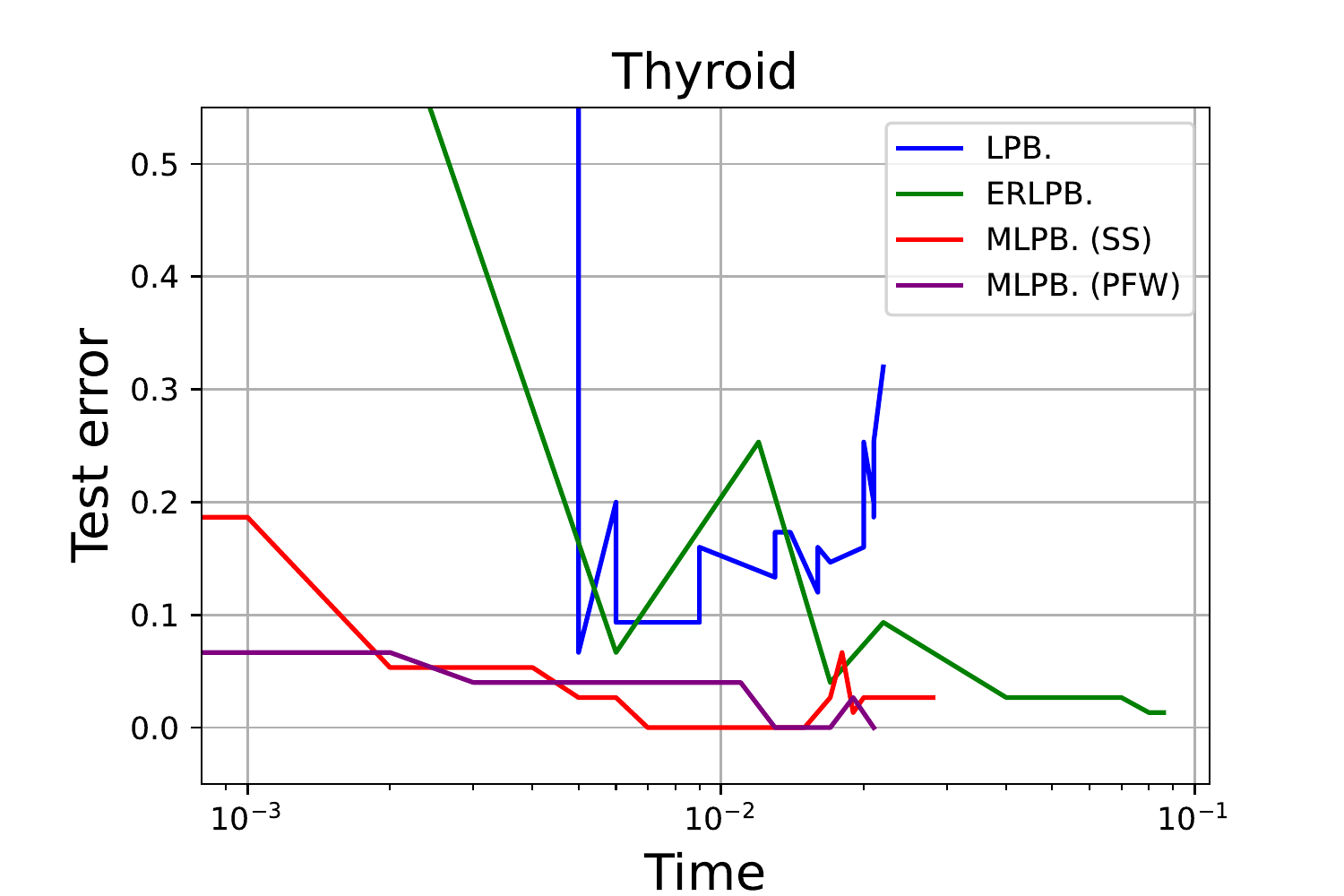}
        \end{minipage}
        &
        \begin{minipage}[t]{0.31\hsize}
            \centering
            \includegraphics[keepaspectratio, scale=0.30]
            {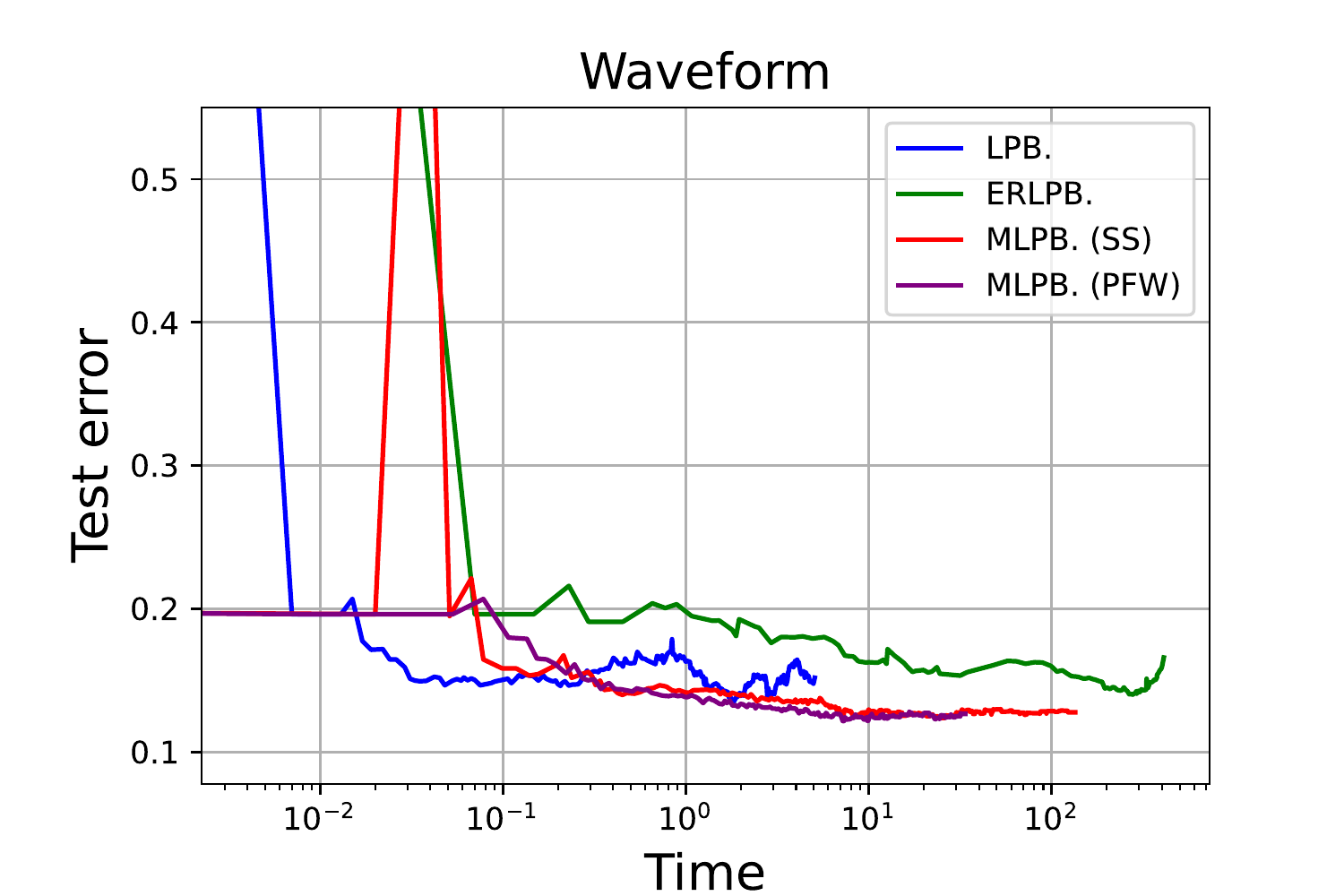}
        \end{minipage}
        &
        \begin{minipage}[t]{0.31\hsize}
            \centering
            \includegraphics[keepaspectratio, scale=0.30]
            {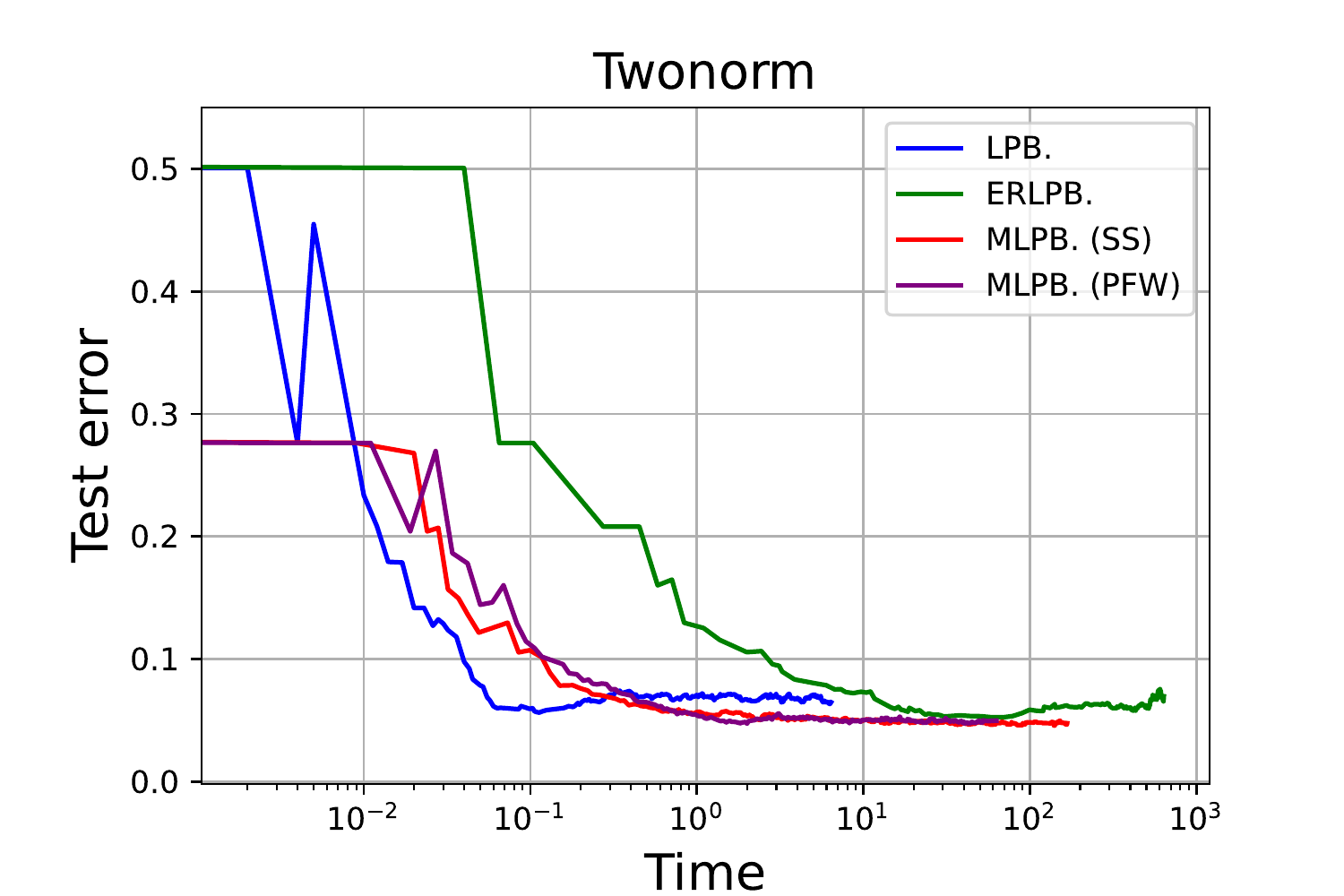}
        \end{minipage}
    \end{tabular}
    \caption{%
        Time vs. test errors for the 0th fold of each dataset %
        with parameters $\nu = 0.1m$ and $\epsilon = 0.01$. %
        Note that the time axis is log-scale. %
        For many datasets, MLPBoost tends to decrease %
        the test error. %
    }
    \label{fig:appendix_test_errors}
\end{figure}

Further, we compare the test error decrease. 
Figure~\ref{fig:appendix_test_errors} shows the test error curves. 
MLPB.~(SS) achieves low test errors in most datasets. 

\begin{figure}[p]
    \centering
    \begin{tabular}{ccc}
        \begin{minipage}[t]{0.31\hsize}
            \centering
            \includegraphics[keepaspectratio, scale=0.30]
            {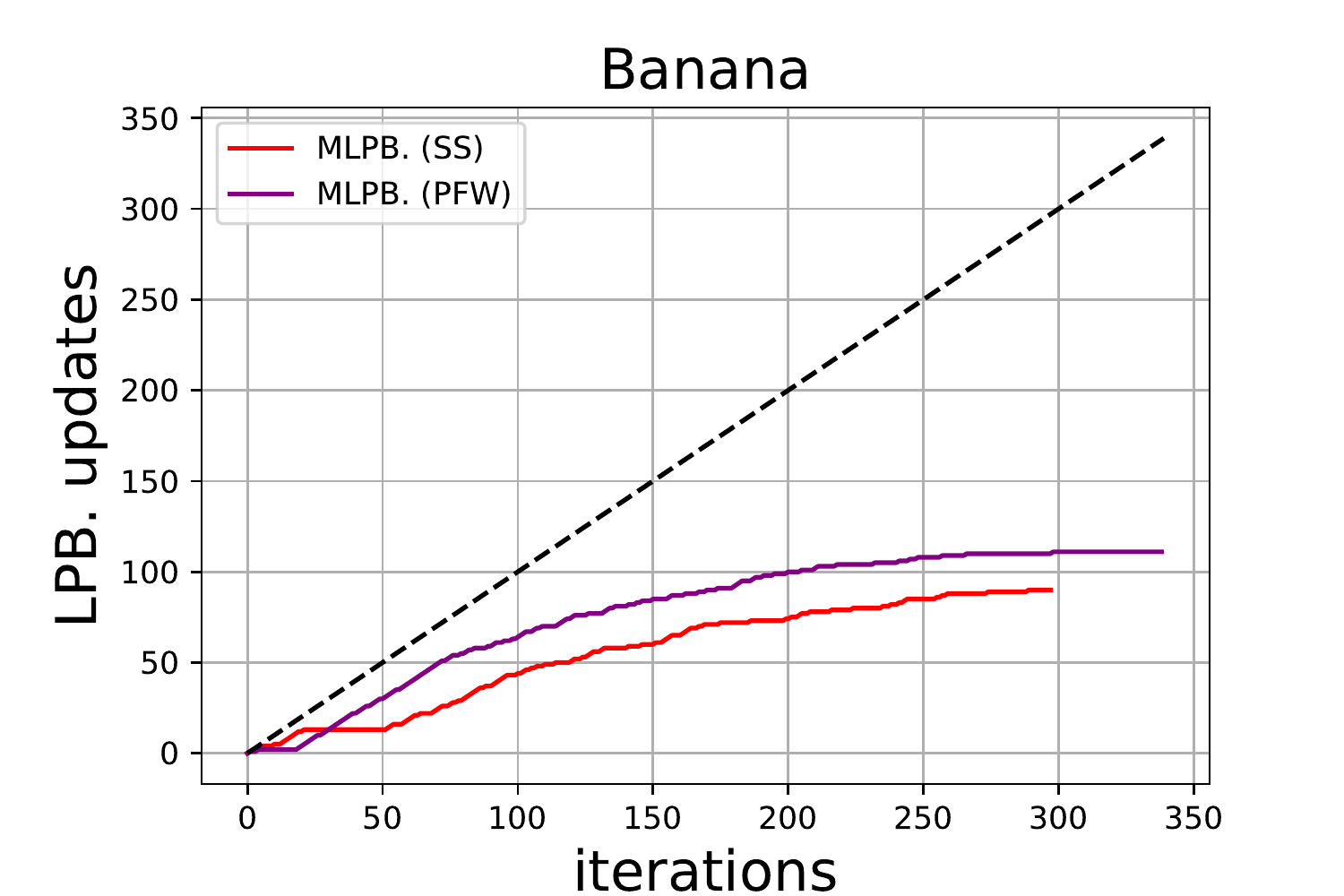}
        \end{minipage}
        &
        \begin{minipage}[t]{0.31\hsize}
            \centering
            \includegraphics[keepaspectratio, scale=0.30]
            {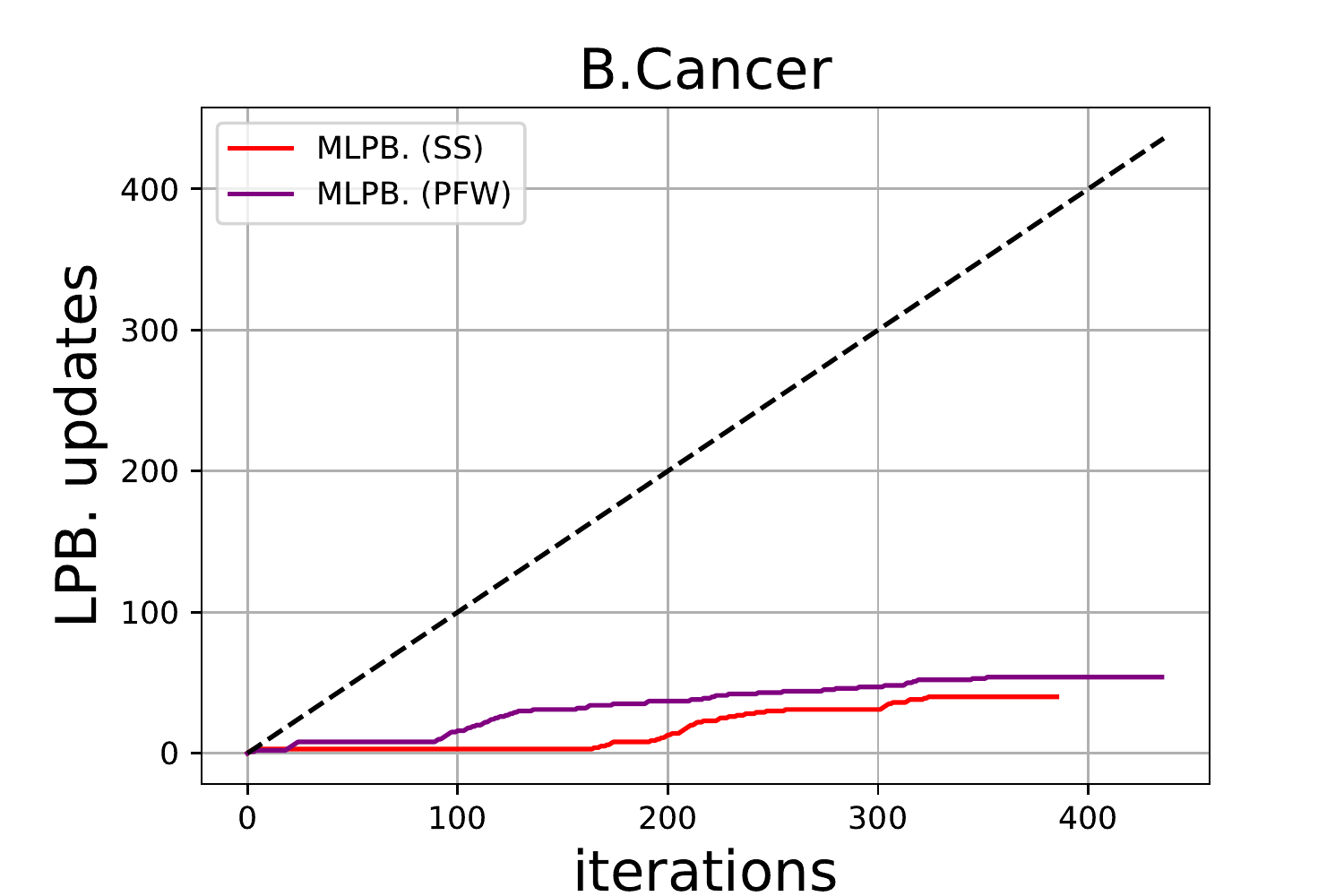}
        \end{minipage}
        &
        \begin{minipage}[t]{0.31\hsize}
            \centering
            \includegraphics[keepaspectratio, scale=0.30]
            {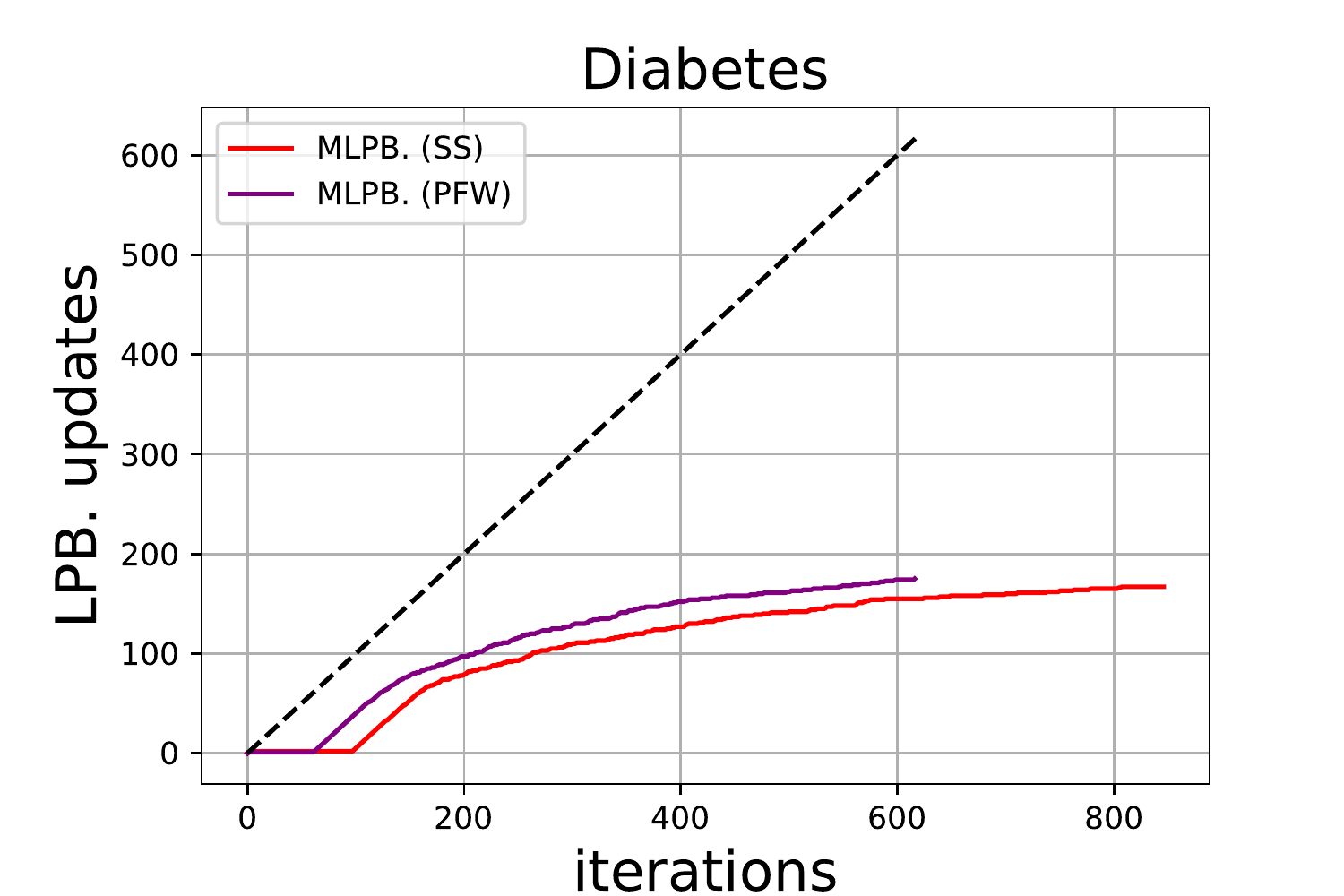}
        \end{minipage}
        \\
        \begin{minipage}[t]{0.31\hsize}
            \centering
            \includegraphics[keepaspectratio, scale=0.30]
            {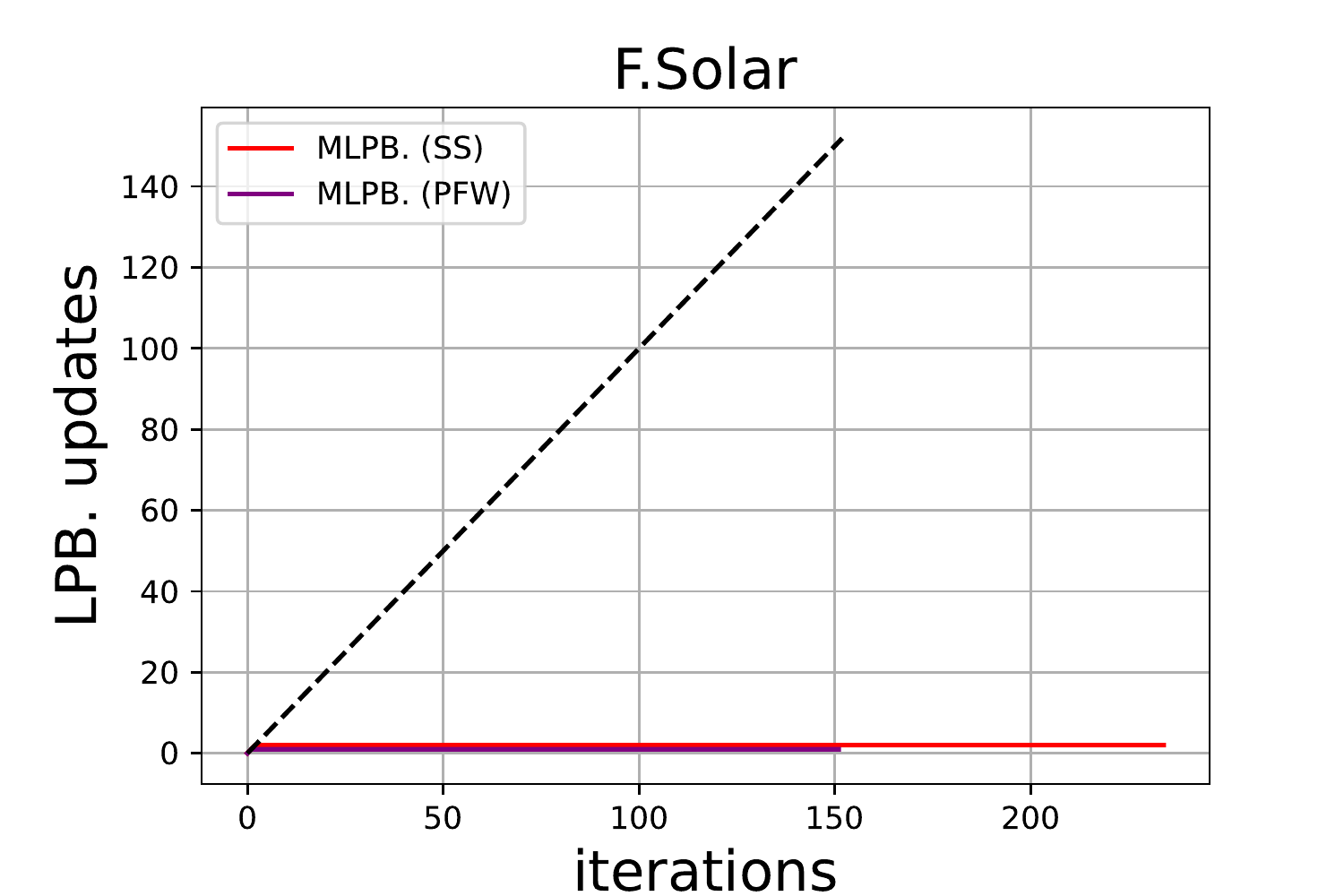}
        \end{minipage}
        &
        \begin{minipage}[t]{0.31\hsize}
            \centering
            \includegraphics[keepaspectratio, scale=0.30]
            {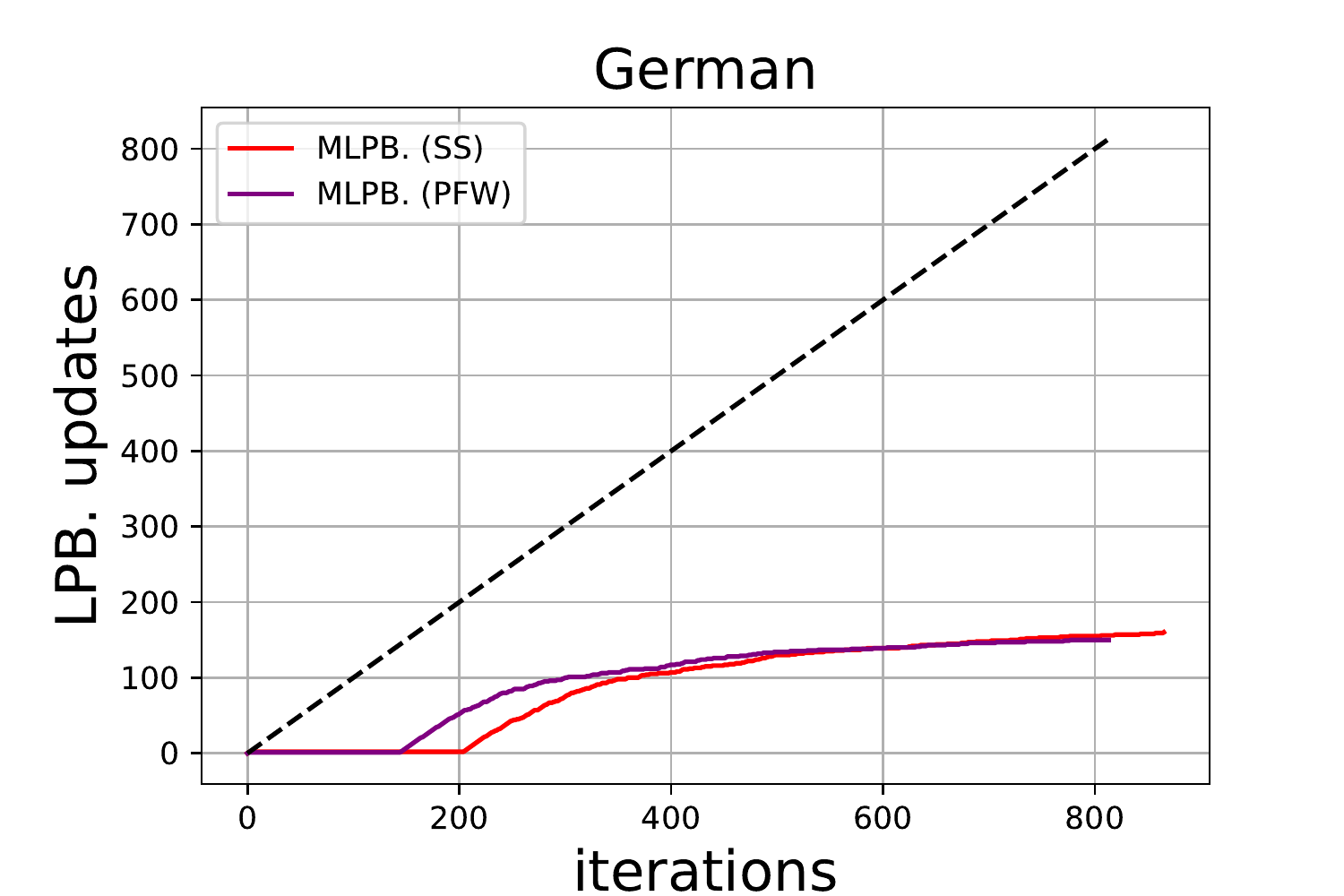}
        \end{minipage}
        &
        \begin{minipage}[t]{0.31\hsize}
            \centering
            \includegraphics[keepaspectratio, scale=0.30]
            {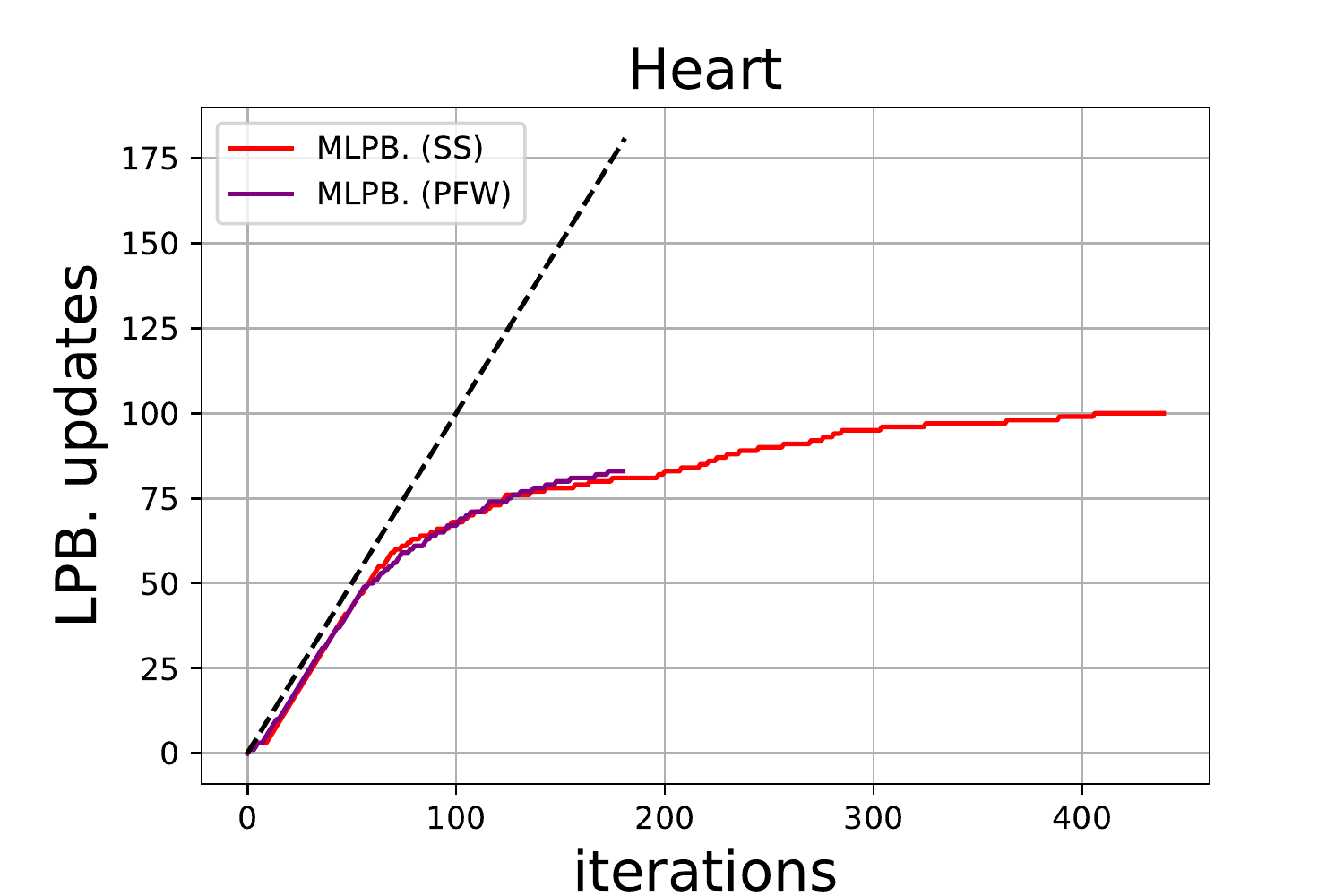}
        \end{minipage}
        \\
        \begin{minipage}[t]{0.31\hsize}
            \centering
            \includegraphics[keepaspectratio, scale=0.30]
            {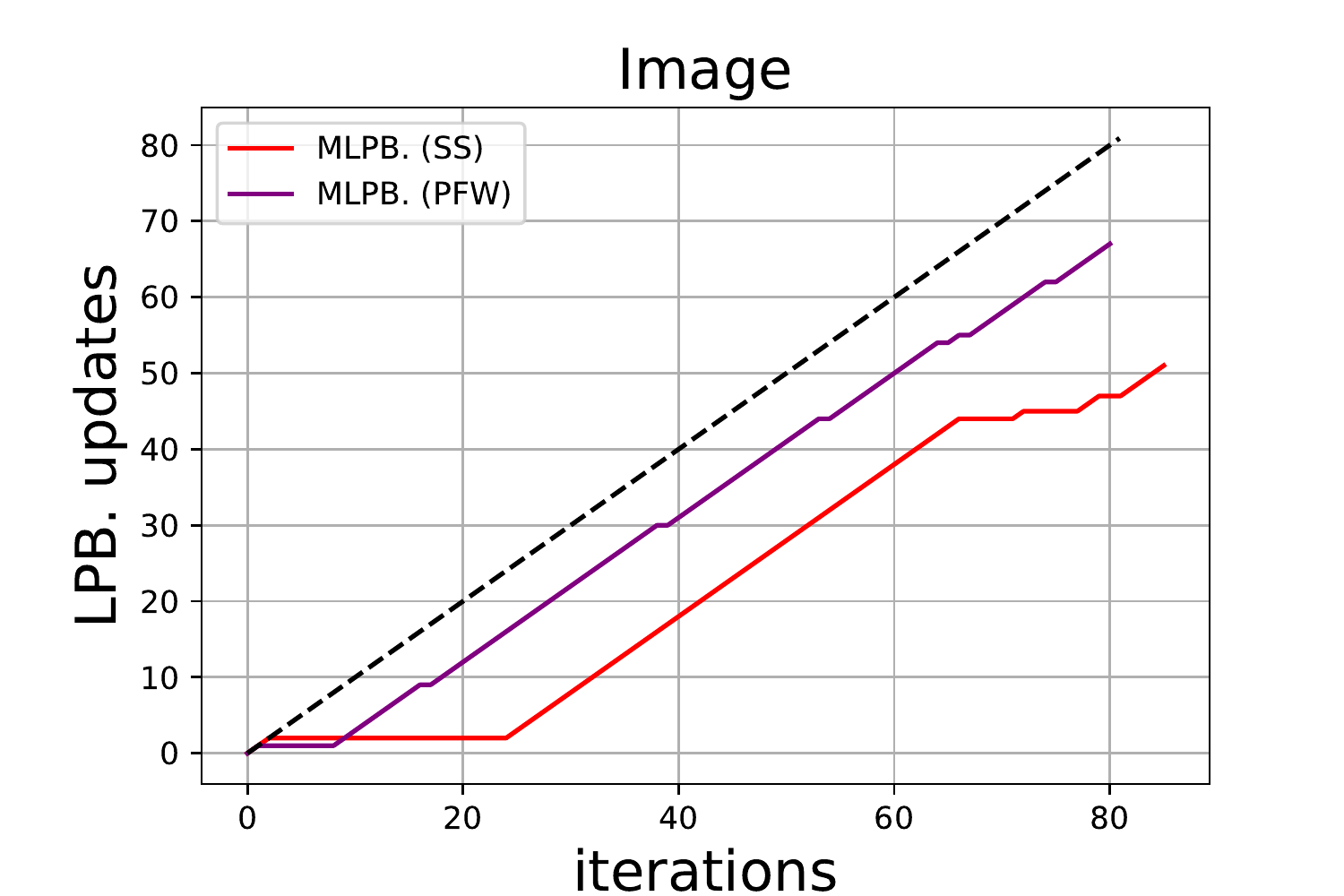}
        \end{minipage}
        &
        \begin{minipage}[t]{0.31\hsize}
            \centering
            \includegraphics[keepaspectratio, scale=0.30]
            {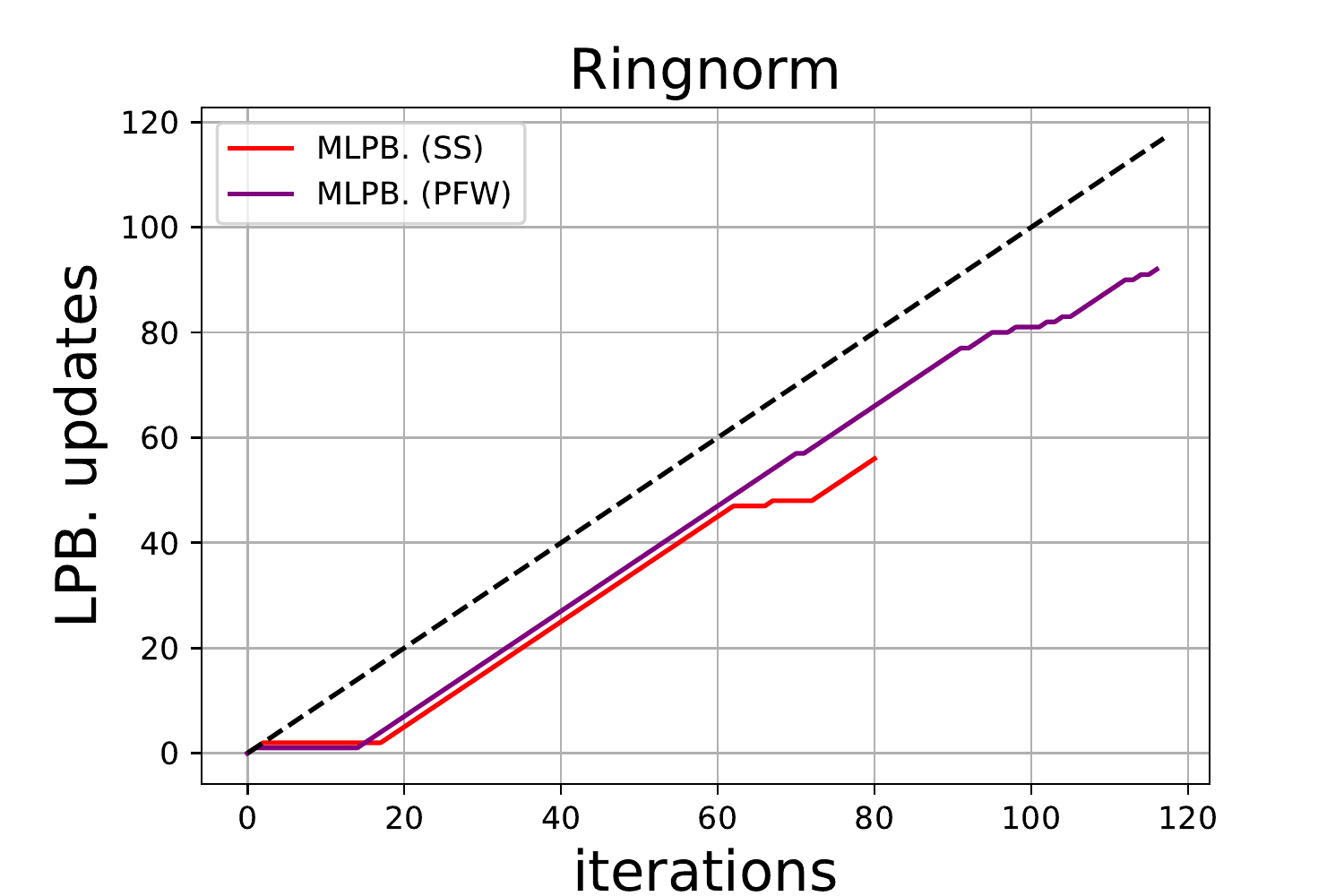}
        \end{minipage}
        &
        \begin{minipage}[t]{0.31\hsize}
            \centering
            \includegraphics[keepaspectratio, scale=0.30]
            {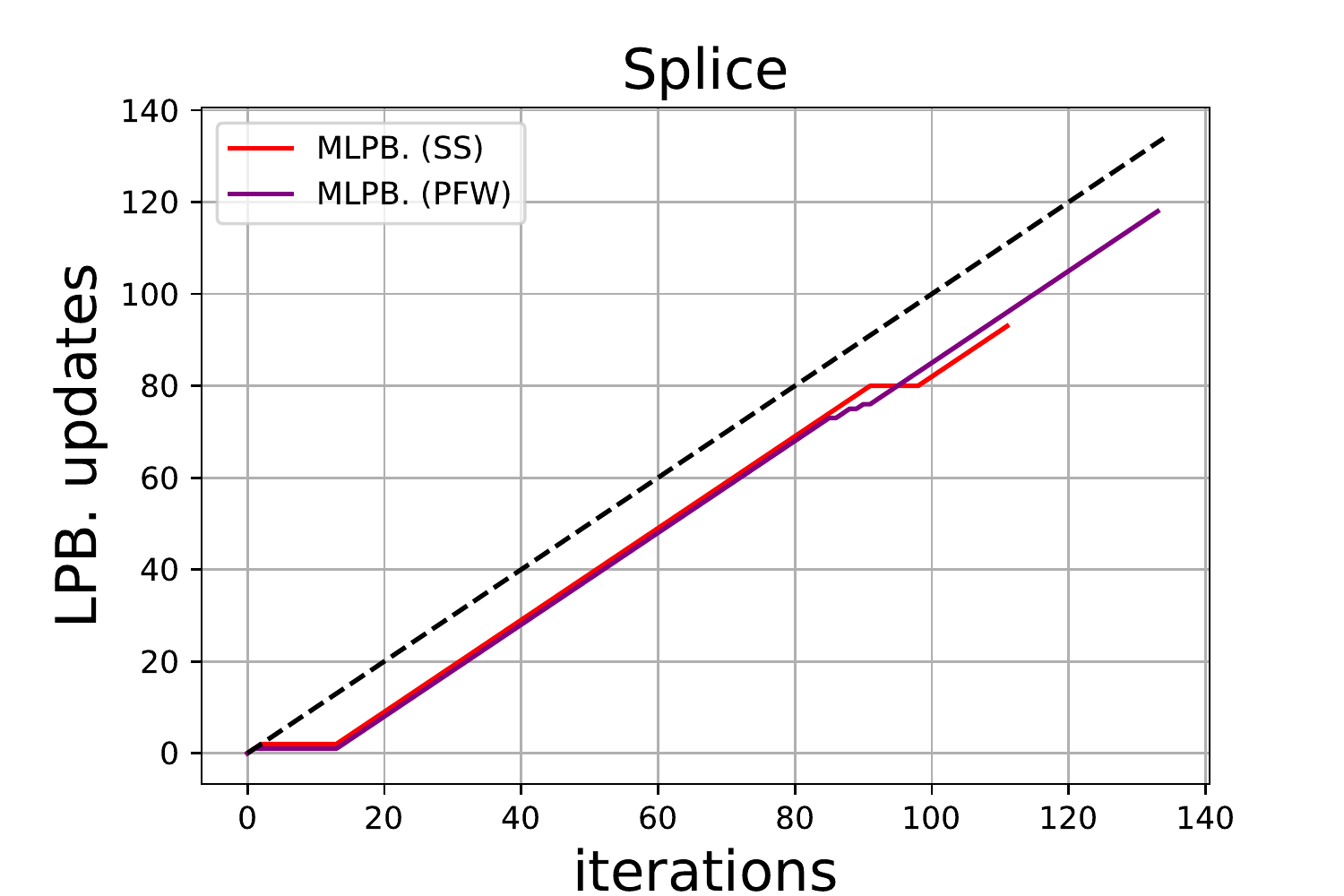}
        \end{minipage}
        \\
        \begin{minipage}[t]{0.31\hsize}
            \centering
            \includegraphics[keepaspectratio, scale=0.30]
            {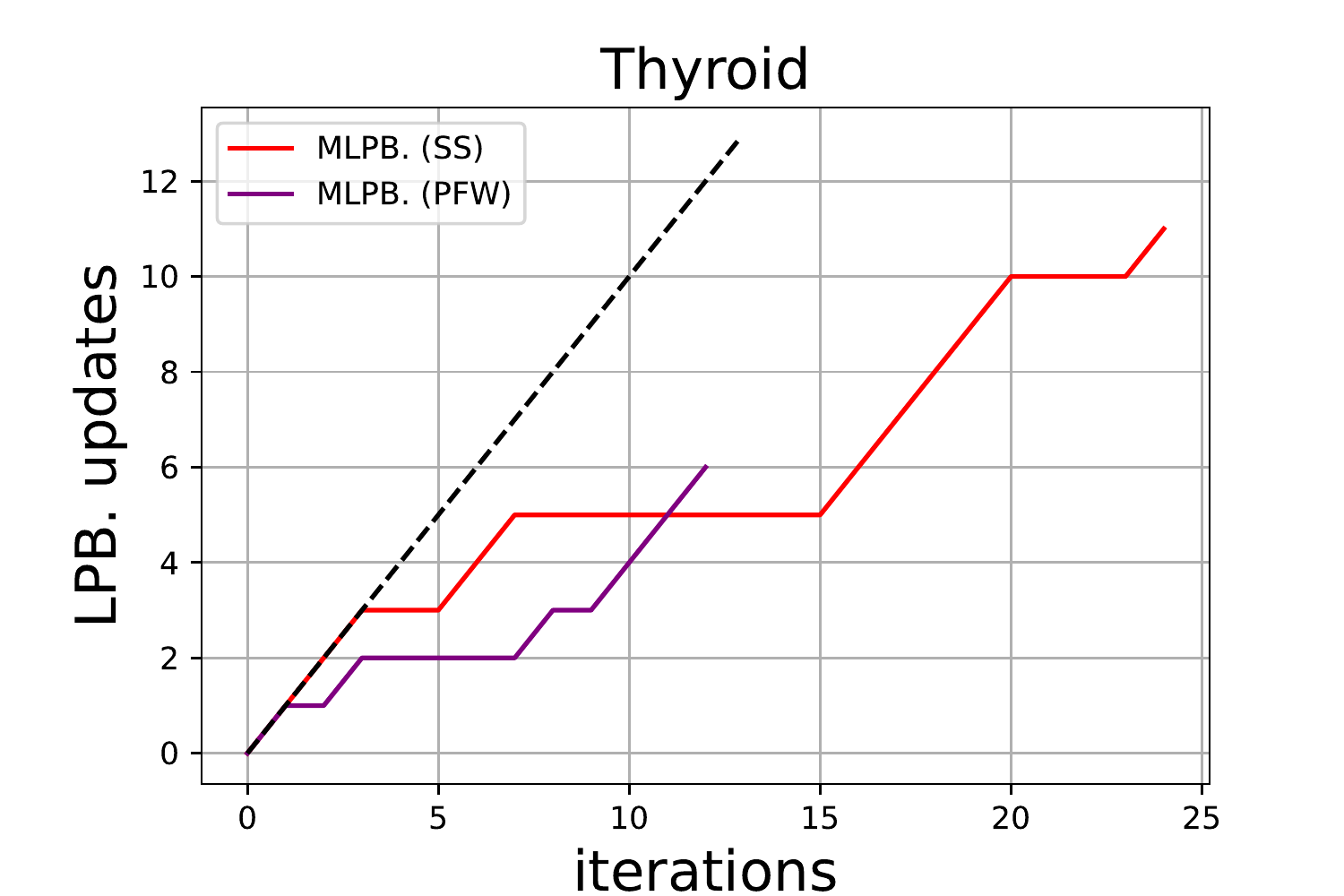}
        \end{minipage}
        &
        \begin{minipage}[t]{0.31\hsize}
            \centering
            \includegraphics[keepaspectratio, scale=0.30]
            {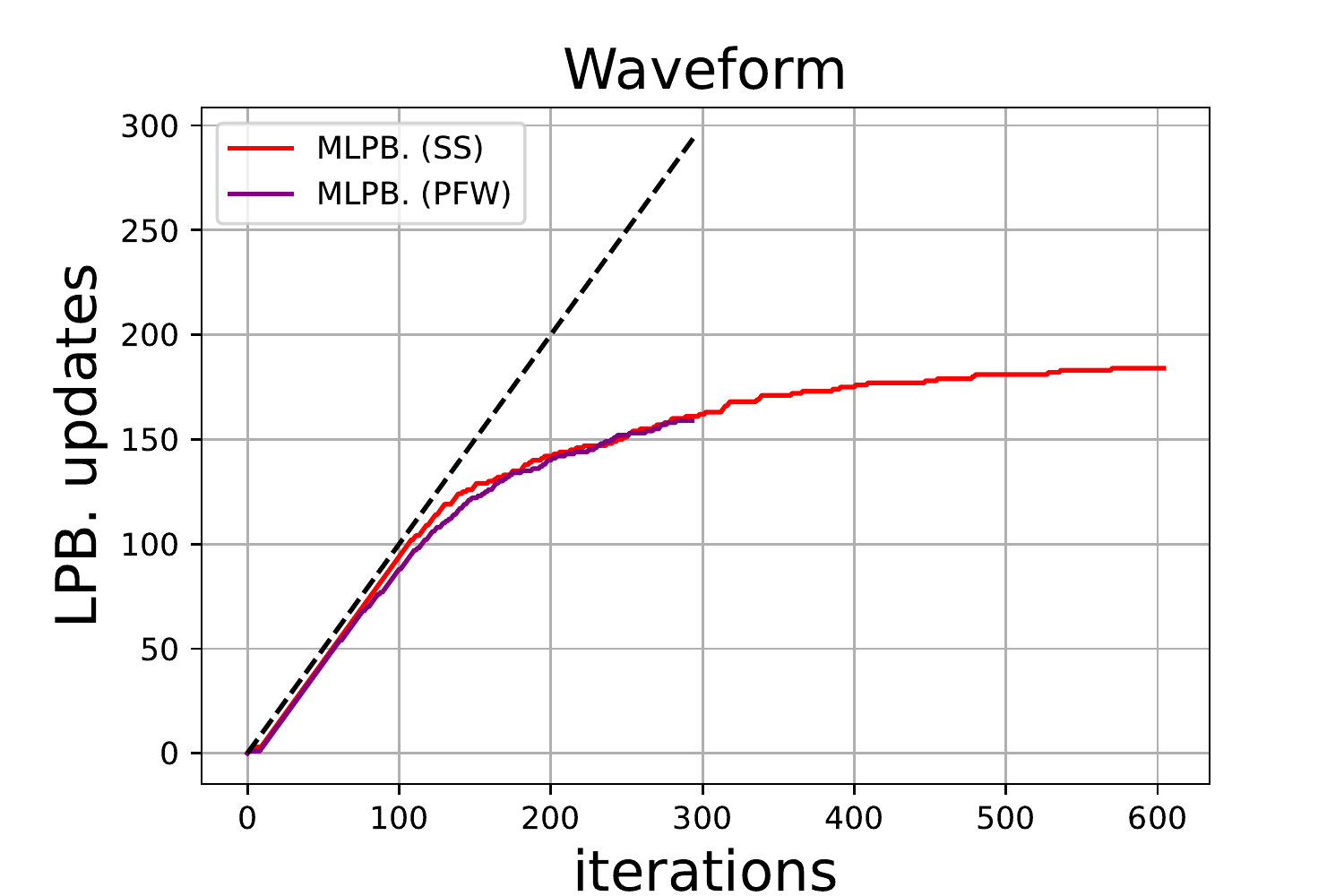}
        \end{minipage}
        &
        \begin{minipage}[t]{0.31\hsize}
            \centering
            \includegraphics[keepaspectratio, scale=0.30]
            {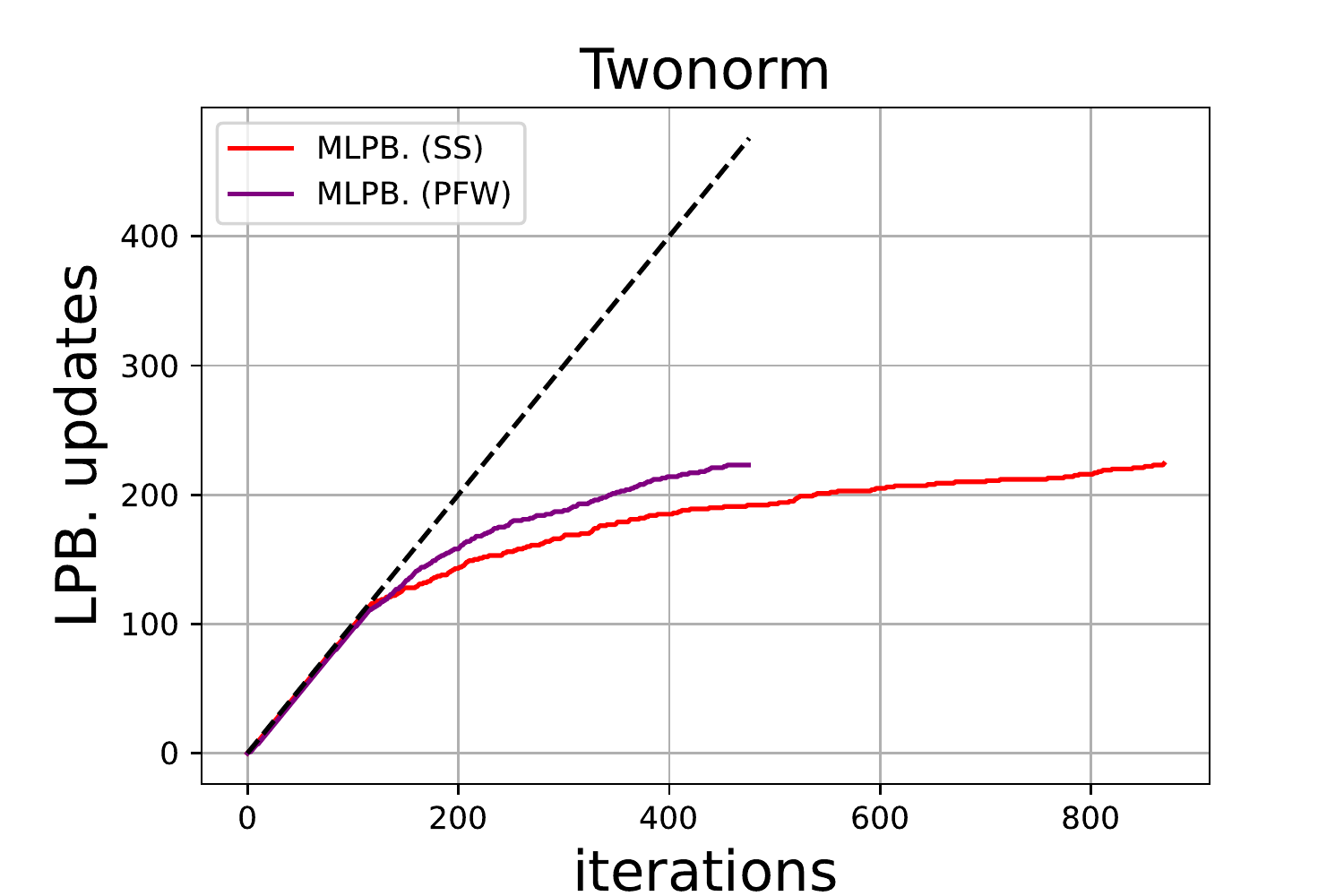}
        \end{minipage}
    \end{tabular}
    \caption{%
        The number of $\secalg$ updates %
        for each benchmark dataset %
        with parameters $\nu = 0.1m$ and $\epsilon = 0.01$. %
        The dotted line indicates the linear function for comparison. %
        Since the shape of the titanic dataset is %
        the same as the F. Solar dataset, %
        we omit it. %
    }
    \label{fig:appendix_lpbcall}
\end{figure}

Now, we compare MLPBoosts to the Frank-Wolfe algorithms, FW and PFW. 
Figure~\ref{fig:appendix_lpbcall} shows 
the number of $\secalg$ updates. 
This figure shows that 
the secondary update $\secalg$ yields better progress 
in early iterations. In the latter half, 
$\fwalg$ yields better progress. 

\begin{figure}[p]
    \centering
    \begin{tabular}{ccc}
        \begin{minipage}[t]{0.31\hsize}
            \centering
            \includegraphics[keepaspectratio, scale=0.30]
            {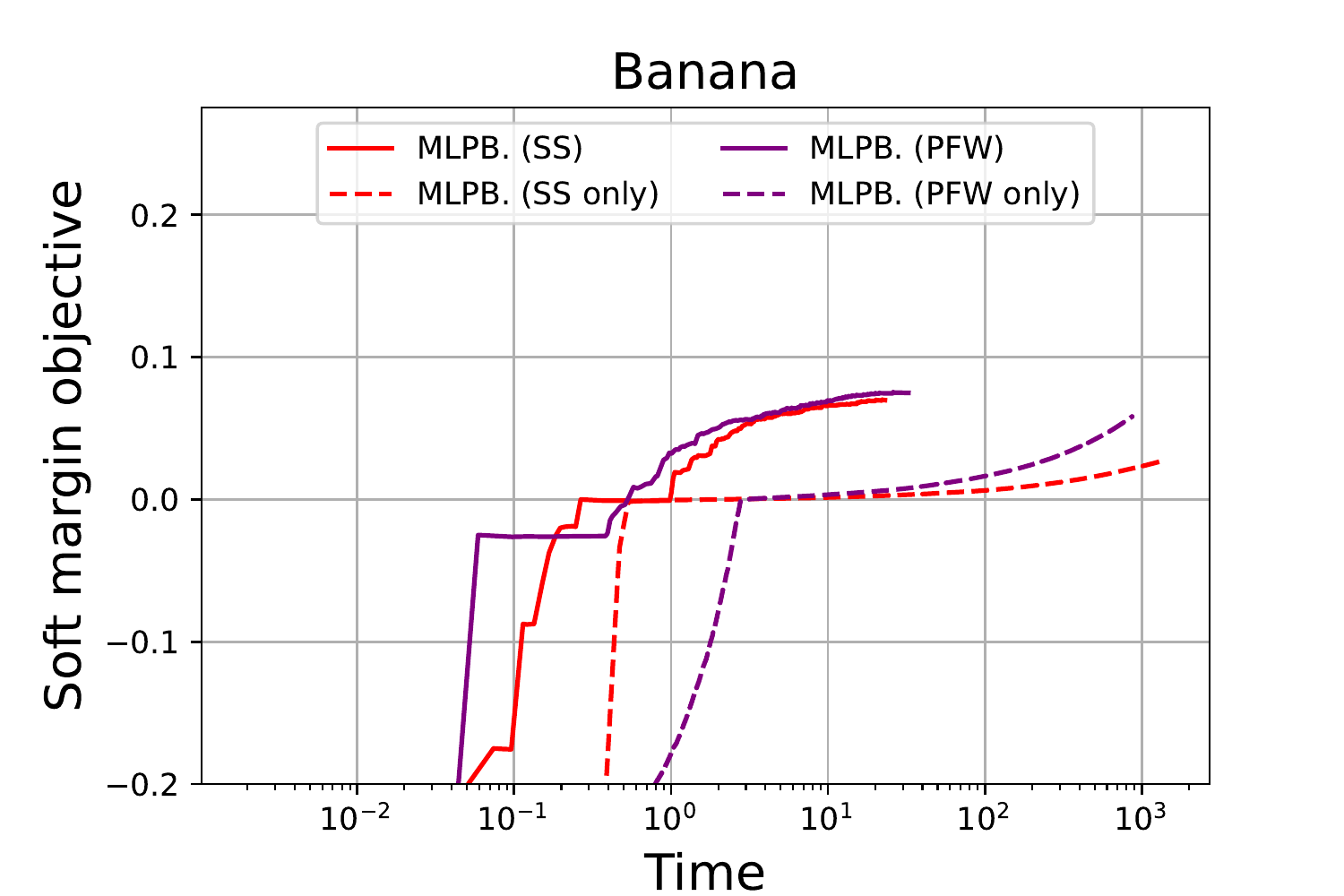}
        \end{minipage}
        &
        \begin{minipage}[t]{0.31\hsize}
            \centering
            \includegraphics[keepaspectratio, scale=0.30]
            {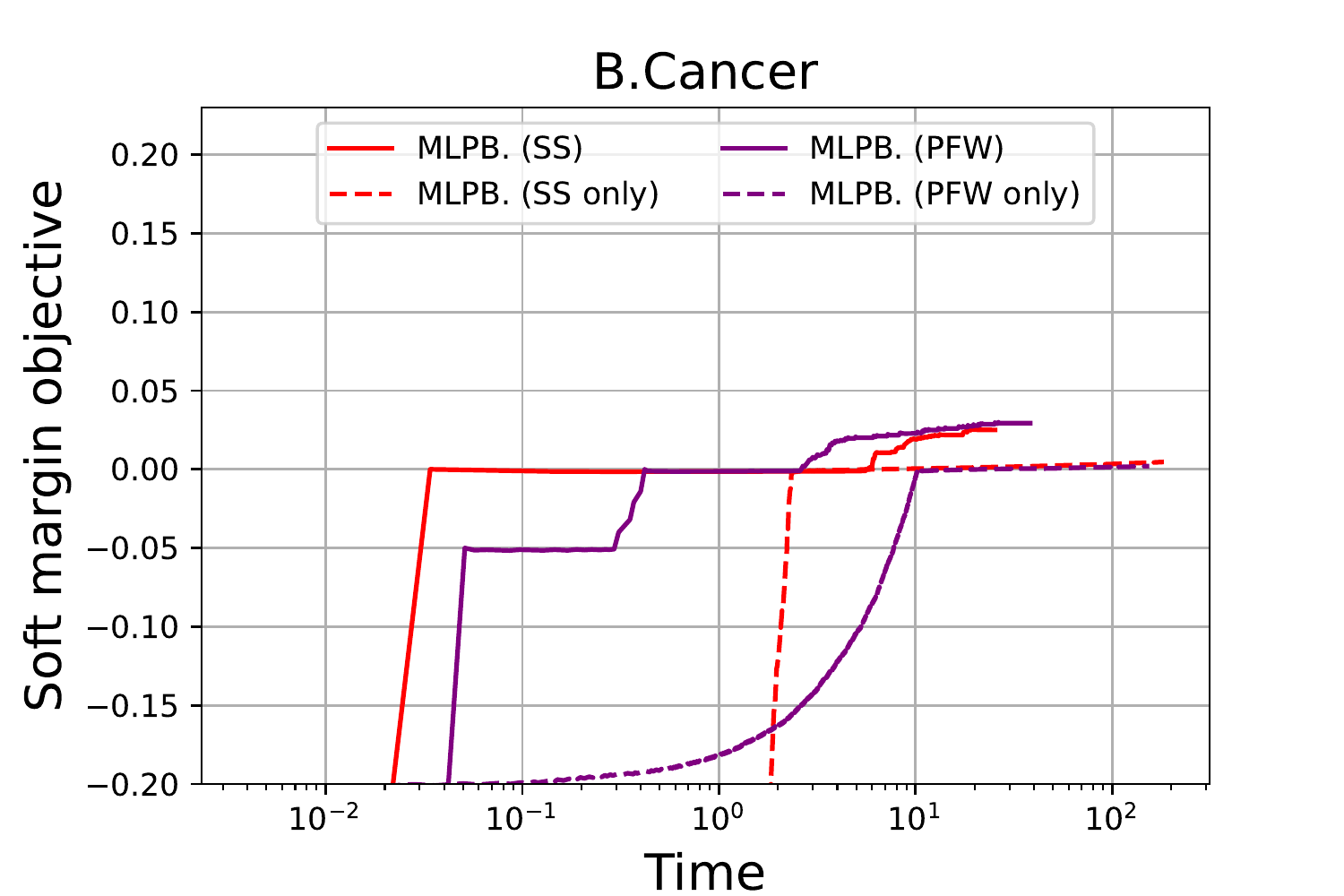}
        \end{minipage}
        &
        \begin{minipage}[t]{0.31\hsize}
            \centering
            \includegraphics[keepaspectratio, scale=0.30]
            {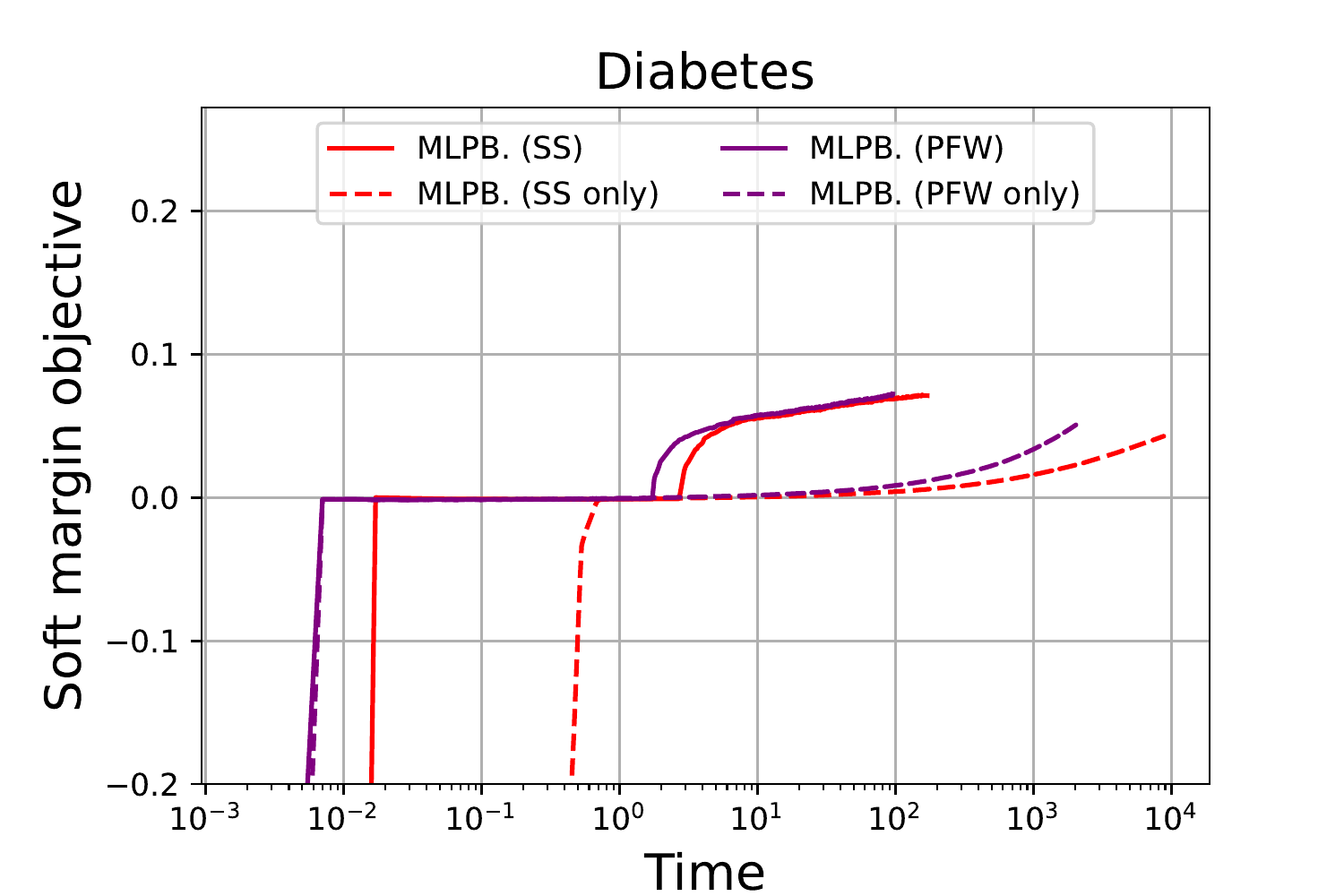}
        \end{minipage}
        \\
        \begin{minipage}[t]{0.31\hsize}
            \centering
            \includegraphics[keepaspectratio, scale=0.30]
            {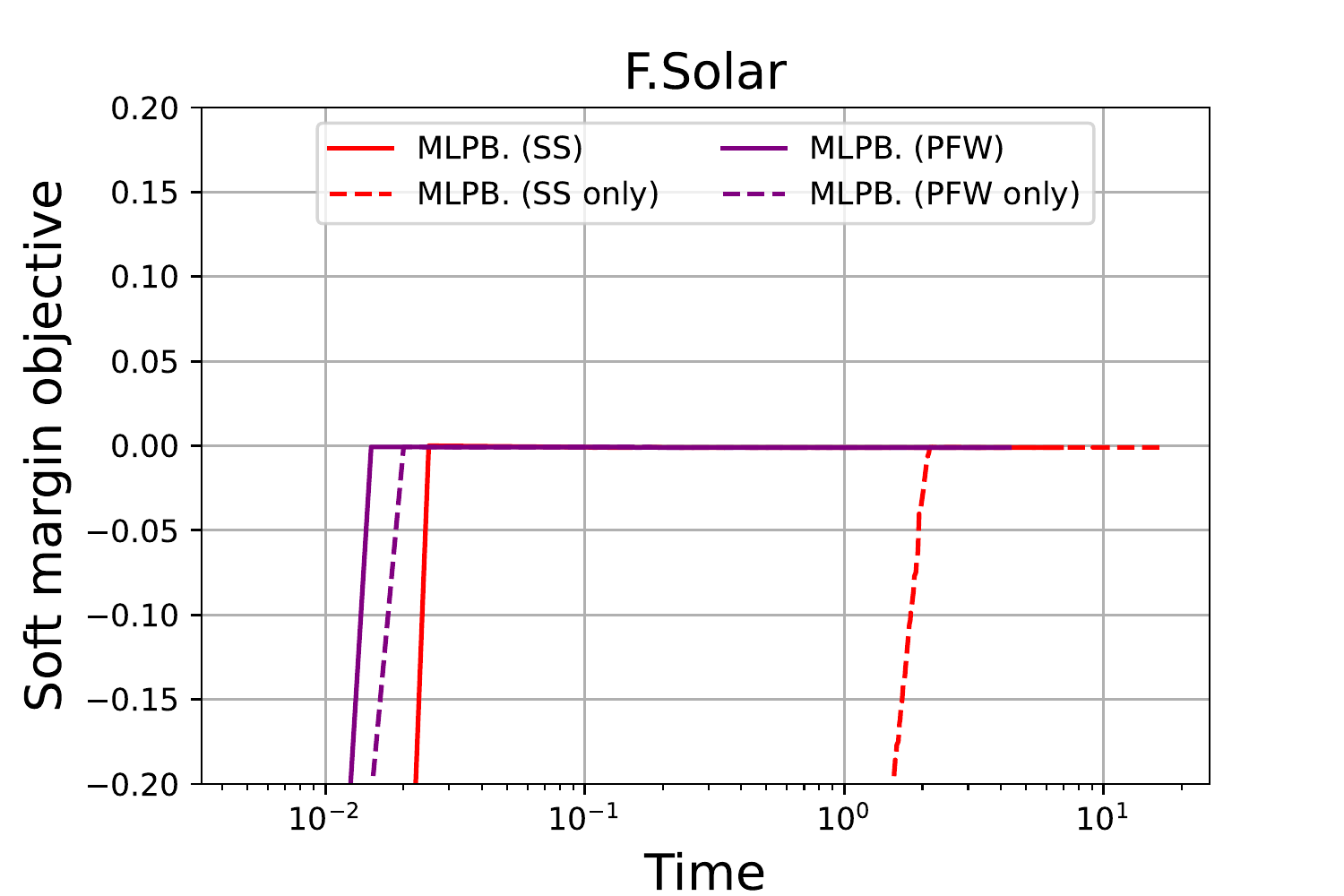}
        \end{minipage}
        &
        \begin{minipage}[t]{0.31\hsize}
            \centering
            \includegraphics[keepaspectratio, scale=0.30]
            {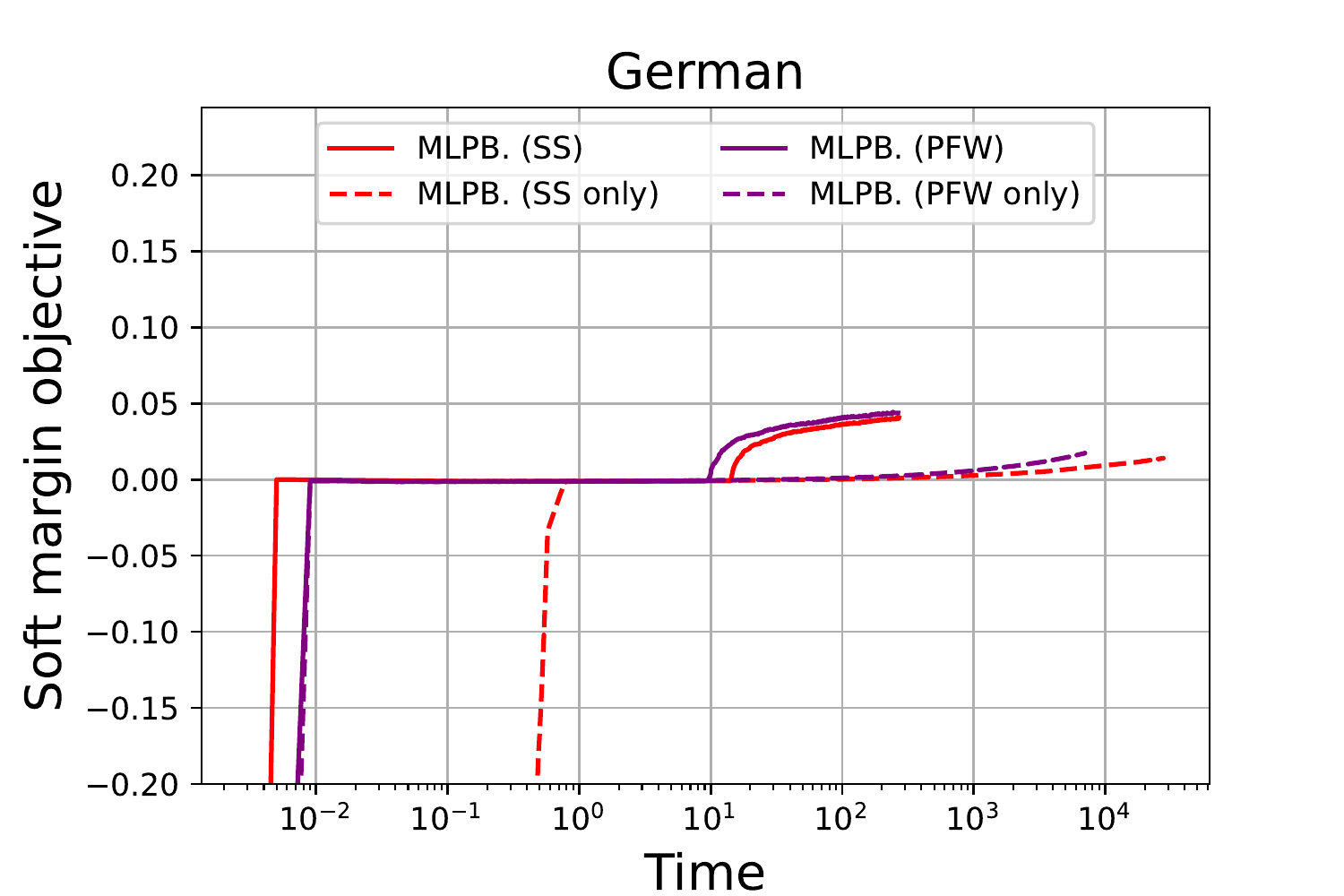}
        \end{minipage}
        &
        \begin{minipage}[t]{0.31\hsize}
            \centering
            \includegraphics[keepaspectratio, scale=0.30]
            {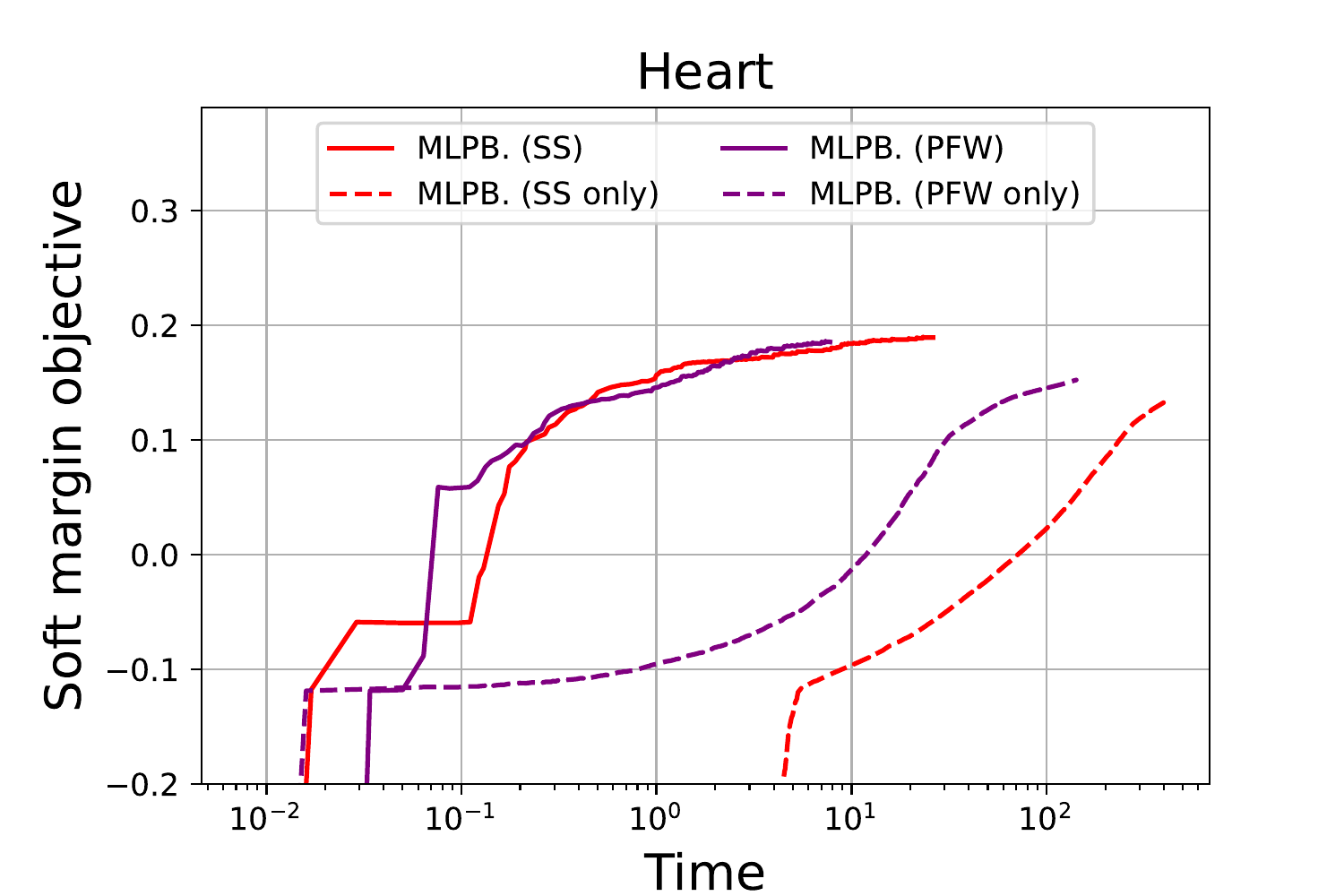}
        \end{minipage}
        \\
        \begin{minipage}[t]{0.31\hsize}
            \centering
            \includegraphics[keepaspectratio, scale=0.30]
            {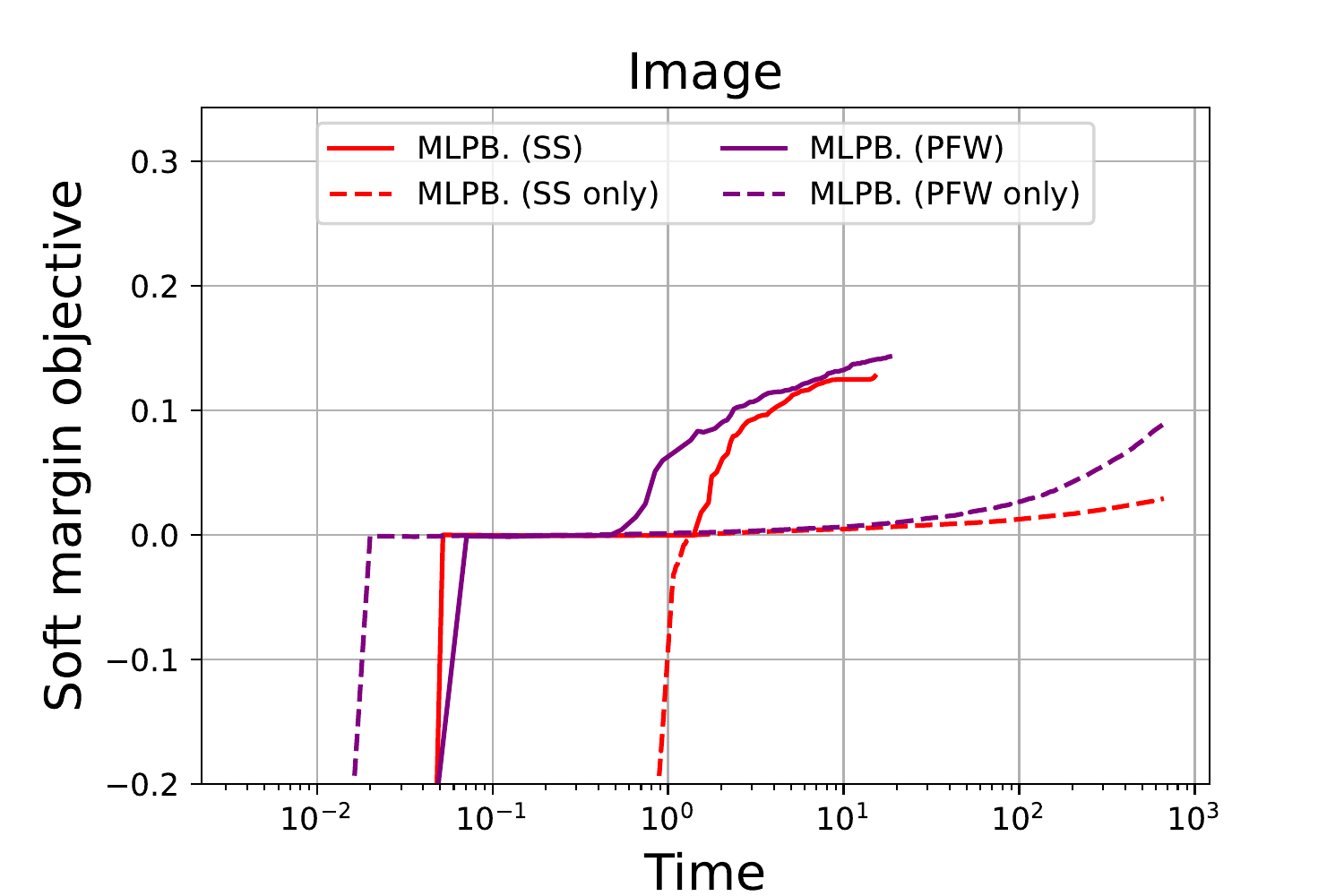}
        \end{minipage}
        &
        \begin{minipage}[t]{0.31\hsize}
            \centering
            \includegraphics[keepaspectratio, scale=0.30]
            {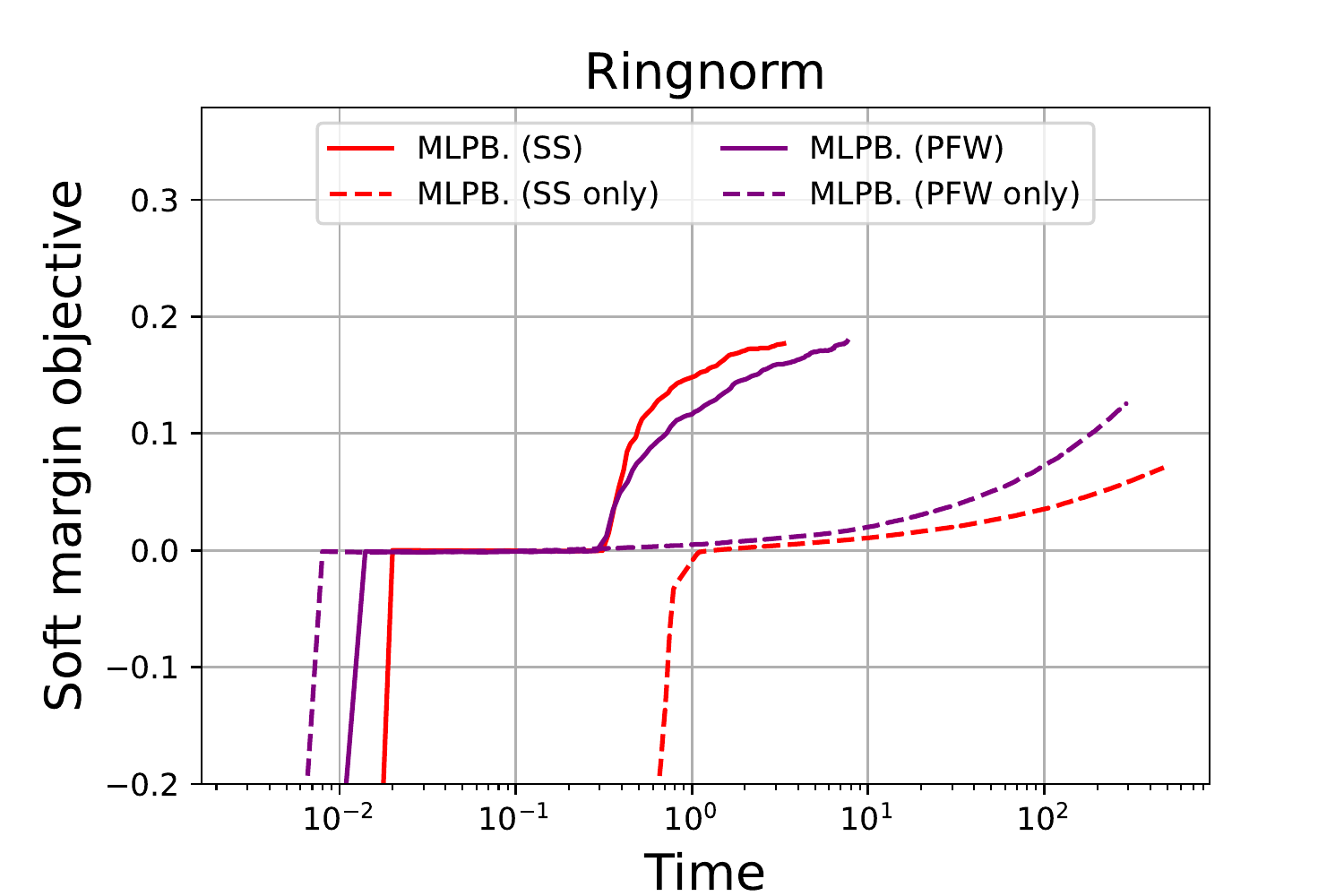}
        \end{minipage}
        &
        \begin{minipage}[t]{0.31\hsize}
            \centering
            \includegraphics[keepaspectratio, scale=0.30]
            {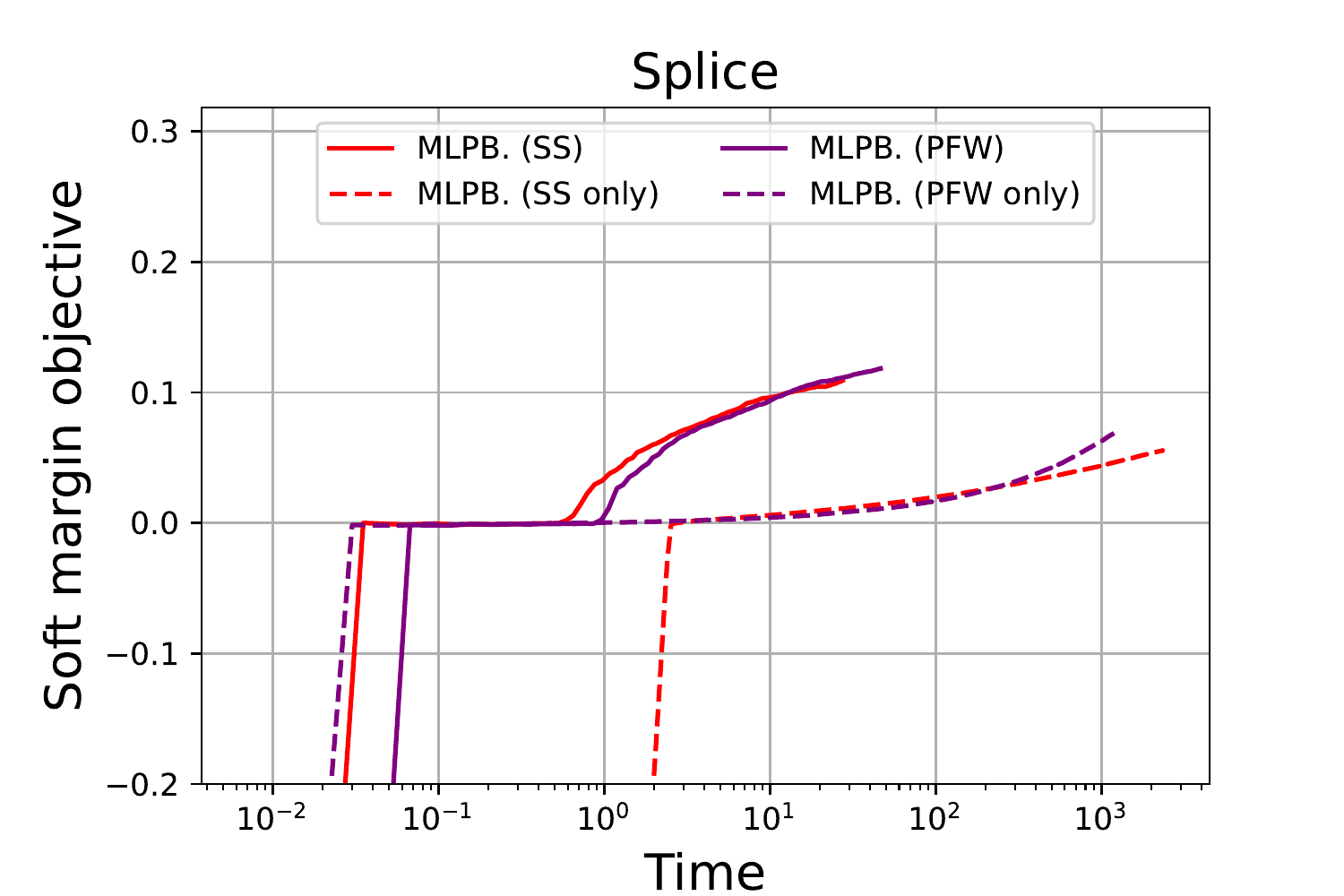}
        \end{minipage}
        \\
        \begin{minipage}[t]{0.31\hsize}
            \centering
            \includegraphics[keepaspectratio, scale=0.30]
            {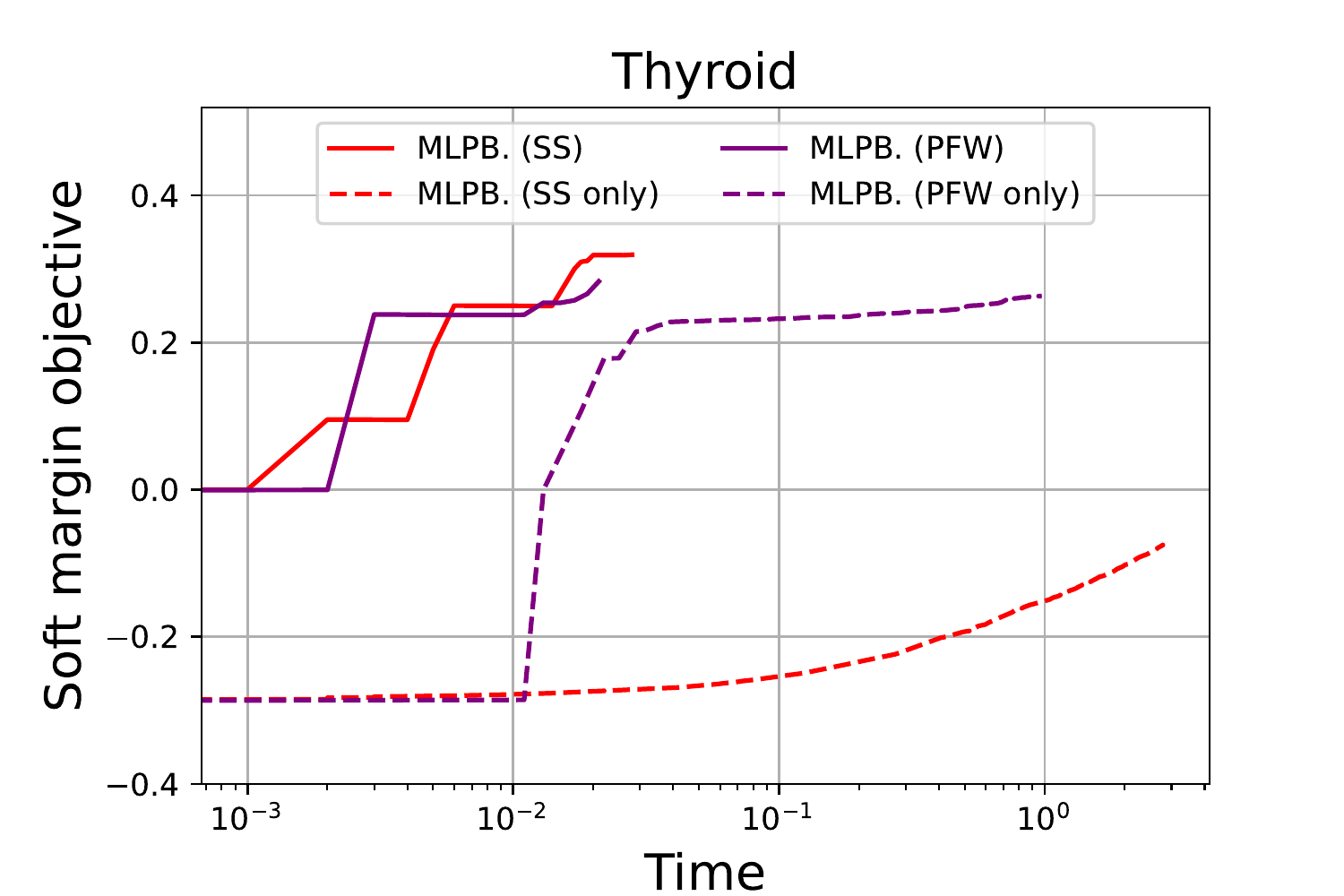}
        \end{minipage}
        &
        \begin{minipage}[t]{0.31\hsize}
            \centering
            \includegraphics[keepaspectratio, scale=0.30]
            {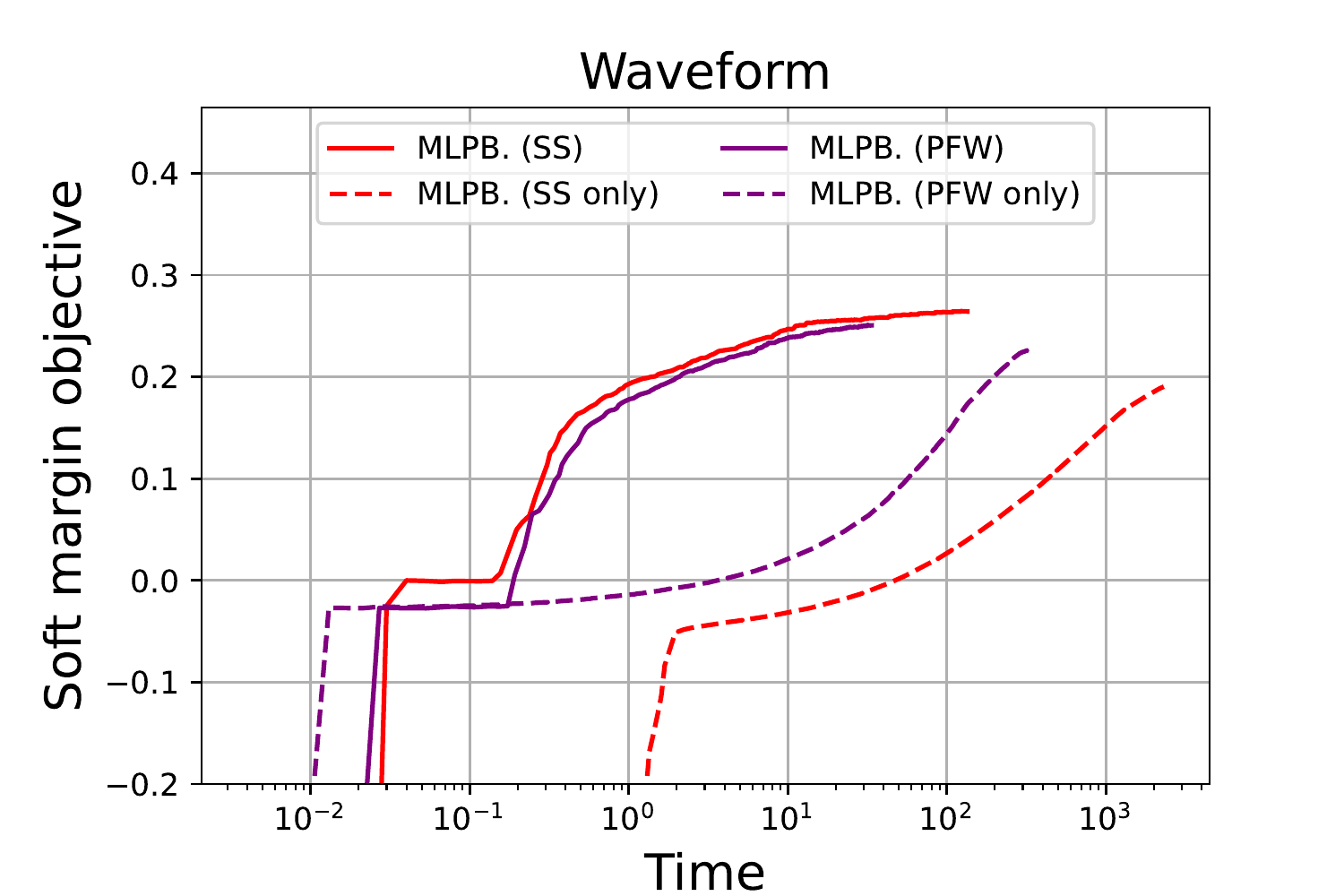}
        \end{minipage}
        &
        \begin{minipage}[t]{0.31\hsize}
            \centering
            \includegraphics[keepaspectratio, scale=0.30]
            {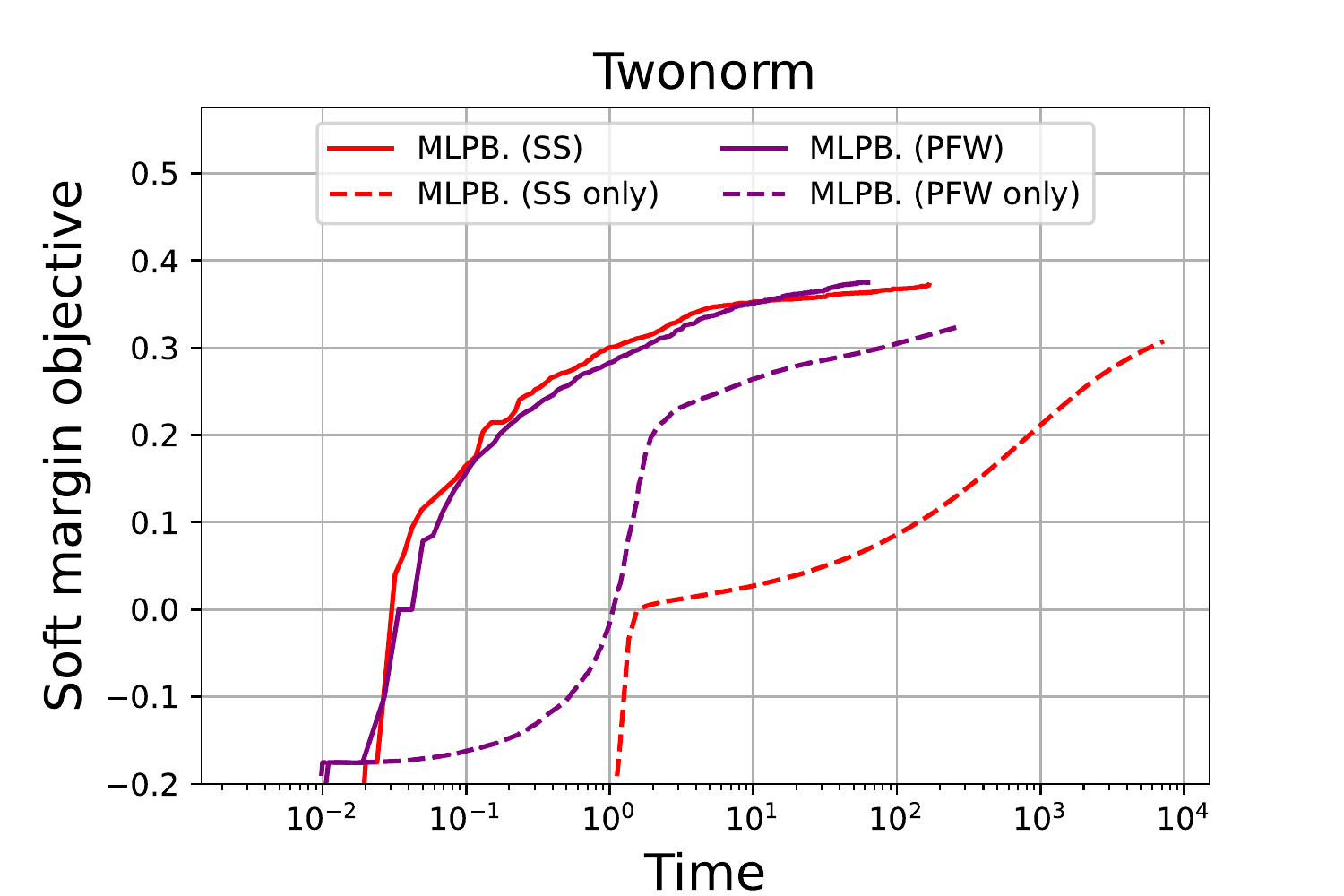}
        \end{minipage}
    \end{tabular}
    \caption{%
        Comparison of the FW algorithms and MLPBoosts. %
        As this figure shows, the secondary update $\secalg$ yields %
        huge progress. %
    }
    \label{fig:appendix_mlpb_compare}
\end{figure}

Finally, we verify the effectiveness of the secondary update $\secalg$, 
shown in algorithm~\ref{alg:lpb_subroutine}. 
For comparison, we measured the soft margin objective and time 
for FW, PFW, MLPB.~(SS), and MLPB.~(PFW). 
FW is the FW algorithm with short-steps, 
and PFW is the Pairwise FW algorithm. 
MLPB.~(SS) and MLPB.~(PFW) are MLPBoosts 
with algorithms~\ref{alg:ss_rule} and~\ref{alg:pairwise_rule}, 
respectively. 
Figure~\ref{fig:appendix_mlpb_compare} shows the results. 
As this figure shows, 
the secondary update $\secalg$ 
improves the objective value significantly. 

\end{document}